\documentclass{article}
\usepackage{arxiv}

\usepackage[utf8]{inputenc}
\usepackage{times}
\usepackage{url}
\usepackage{graphicx}

\usepackage{graphics}
\usepackage{pgf}
\usepackage[hidelinks]{hyperref}

\usepackage{amsmath}

\usepackage{amsthm}
\usepackage{comment}

\usepackage{rotating}
\usepackage{afterpage}
\usepackage{arydshln}
\usepackage[T1]{fontenc}

\usepackage{color}
\usepackage{multirow}

\usepackage{tikz}
\usepackage{verbatim}
\usepackage{enumitem}
\usepackage[titlenumbered,linesnumbered,ruled,vlined]{algorithm2e}
\usepackage{makecell}

\newcommand\Tau{\mathcal{T}}
\newcommand\toprule{\Xhline{1pt}}
\newcommand\midrule{\Xhline{0.5pt}}
\newcommand\bottomrule{\Xhline{1pt}}

\theoremstyle{definition}
\newtheorem{definition}{Definition}

\theoremstyle{definition}
\newtheorem{proposition}{Proposition}

\theoremstyle{definition}
\newtheorem{rmarkN}{Remark}

\theoremstyle{definition}

\theoremstyle{plain}
\newtheorem{xampl}{Example}

\newcommand{\footlabel}[2]{%
    \addtocounter{footnote}{1}%
    \footnotetext[\thefootnote]{%
        \addtocounter{footnote}{-1}%
        \refstepcounter{footnote}\label{#1}%
        #2%
    }%
    $^{\ref{#1}}$%
}

\newcommand{\footref}[1]{%
    $^{\ref{#1}}$%
}

\usetikzlibrary{arrows,shapes}

\date{}

\begin{document}
\title{Incomplete MaxSAT Approaches for Combinatorial Testing}

\author{Carlos Ansótegui \\
    Logic \& Optimization Group (LOG) \\
    University of Lleida, Lleida, Spain \\
    \texttt{carlos.ansotegui@udl.cat}
    \And
    Felip Manyà \\
    Artificial Intelligence Research Institute (IIIA, CSIC) \\
    Campus UAB, 08193 Bellaterra, Spain \\
    \texttt{felip@iiia.csic.es}
    \And
    Jesus Ojeda \\
    Logic \& Optimization Group (LOG) \\
    University of Lleida, Lleida, Spain \\
    \texttt{jesus.ojedacontreras@udl.cat}
    \And
    Josep M. Salvia \\
    Logic \& Optimization Group (LOG) \\
    University of Lleida, Lleida, Spain \\
    \texttt{josh.salvia@gmail.com}
    \And
    Eduard Torres \\
    Logic \& Optimization Group (LOG) \\
    University of Lleida, Lleida, Spain \\
    \texttt{eduard.torres@udl.cat}
    }


\maketitle

\begin{abstract}

We present a Satisfiability (SAT)-based approach for building Mixed Covering Arrays with Constraints of minimum length, referred to as the Covering Array Number problem. This problem is central in Combinatorial Testing for the detection of system failures. In particular, we show how to apply Maximum Satisfiability (MaxSAT) technology by describing efficient encodings for different classes of complete and incomplete MaxSAT solvers to compute optimal and suboptimal solutions, respectively. Similarly, we show how to solve through MaxSAT technology a closely related problem, the Tuple Number problem, which we extend to incorporate constraints. For this problem, we additionally provide a new MaxSAT-based incomplete algorithm. The extensive experimental evaluation we carry out on the available Mixed Covering Arrays with Constraints benchmarks and the comparison with state-of-the-art tools confirm the good performance of our approaches.

\end{abstract}

\section{Introduction}

The Combinatorial Testing (CT) problem \cite{nie2011survey} addresses the question of how to efficiently verify the proper operation of a system, where a system can be a program, a circuit, a package that integrates several pieces of software, a GUI interface, a cloud application, etc. This problem requires to explore the parameter space of the system by iteratively testing different settings of the parameters to detect errors, bugs or faults. If we consider the system parameters as variables, a setting can be described as a \emph{full} assignment to these parameters.

Exploring all the parameter space exhaustively, i.e., the set of all possible full assignments, is, in general, out of reach. Notice that if a system has a set of parameters $P$, the number of different full assignments is 
$\prod_{p \in P} g_p = \mathcal{O}\left(g^{|P|}\right)$, 
where $g_p$ is the cardinality of the domain of parameter $p$ and $g$ is the cardinality of the greatest domain.  %

The good news is that, in practice, there is no need to explore all the parameter space to detect errors, bugs or faults. We just need to \emph{cover} a portion of the possible parameter combinations \cite{KuhnWG04}. For example, most software errors (75\%-80\%) are caused by certain individual parameters or by the interaction of just two of them.

To cover that portion of parameter combinations exhaustively, Covering Arrays (CAs) play an important role in CT. Given a set of parameters $P$ and a strength $t$, a Covering Array $CA(N;t,P)$ is a test suite of $N$ tests that guarantee to cover all the possible interactions of $t$ parameters (referred as $t$-tuples). Since executing a test in the system has a cost, we are interested in working with relatively small covering arrays. We refer to the minimum $N$ for which a $CA(N;t,P)$ exists as the Covering Array Number, denoted by $CAN(t,P)$. In particular, we are interested in building an optimal CA, i.e., a covering array of length $CAN(t,P)$. 
Notice that it is guaranteed that the number of tests required to cover all $t$-way parameter combinations, for fixed $t$, grows logarithmically in the number of parameters \cite{Colbourn_2004}, 
which indicates 
that optimal or near-optimal covering arrays can be used in practical terms. The computational challenge is to build optimal CAs in a reasonable time frame.

In this paper, we focus on \emph{Mixed} Covering Arrays with \emph{Constraints} (MCACs). The term \emph{Mixed} refers to the possibility of having parameter domains of different sizes. The term \emph{Constraints} refers to the existence of some parameter interactions that are not allowed in the system. These forbidden interactions are usually implicitly described by a set of constraints. The problem
of computing an MCAC of minimum length, to which we refer in this paper as the Covering Array Number problem, is NP-hard \cite{MaltaisM10}.

There exist several greedy approaches that tackle the problem of building minimum MCACs, such as PICT \cite{Czerwonka06},  based on the OTAT framework \cite{BryceCC05}, and ACTS \cite{BorazjanyYLKK12}, based on the IPOG algorithm \cite{DuanLYKK17}. One downside of these approaches is that they become more inefficient as the hardness of the set of forbidden interactions increases. Therefore, we are more interested in constraint programming approaches, which are better suited for handling constraints. For example, CALOT \cite{YamadaKACOB15} is a tool for building MCACs based on Satisfiability (SAT) technology \cite{handbooksat} that can handle constraints efficiently.%

Within constraint programming techniques \cite{RossiVB06}, SAT technology  provides a highly competitive generic problem approach for solving decision problems. In particular, the decision problem to be solved is translated into a SAT instance (a propositional formula) and a SAT solver is used to determine whether there is a solution. In this paper, we will \emph{review} in detail the CALOT tool, which essentially solves a sequence of SAT instances to compute an optimal MCAC. Each SAT instance in the sequence encodes the decision query of whether there exists an MCAC of a certain length. By iteratively bounding the length, the optimum can be determined.

Since the problem of computing minimum MCACs is, in essence, an optimization problem, we also consider its reformulation into the Maximum Satisfiability (MaxSAT) problem \cite{handbooksat}, which is an optimization version of the SAT problem.

We show empirically that MaxSAT approaches outperform ACTS and CALOT (the state-of-the-art) once the suitable MaxSAT encodings are used. We evaluate both complete or exact MaxSAT solvers (certify optimality) and incomplete MaxSAT solvers (provide suboptimal solutions). In particular, we show that while complete MaxSAT solvers perform similar to CALOT (substantially in contrast to previously reported experiments with MaxSAT solvers \cite{YamadaKACOB15}), incomplete MaxSAT solvers obtain better suboptimal solutions and faster than ACTS and CALOT on many instances. This confirms the practical interest of incomplete MaxSAT approaches because, in real environments, we are mainly concerned with obtaining the best possible solution within a given budget.

Having confirmed the good performance of MaxSAT approaches for computing minimum MCACs, we explore another related problem, the Tuple Number (TN) Problem. Informally, the TN problem is to determine the minimum set of missing $t$-tuples in a test suite of $N$ tests, or the maximum set of $t$-tuples that these $N$ tests cover. In \cite{Carrizales-TurrubiatesRT11}, this problem is studied in the context of Covering arrays with uniform domains and without constraints. In this paper, we explore (for the first time) the Mixed and with Constraints variants of the TN problem, assessing the performance of complete and incomplete MaxSAT approaches. Obviously, this problem is of interest when $N < CAN(t,P)$\footnote{For $N \geq  CAN(t,P)$, the Tuple Number problem essentially corresponds to determine the number of allowed tuples in the corresponding MCAC problem.}. We additionally present another incomplete approach based on MaxSAT technology to which we refer as MaxSAT Incremental Test Suite (Maxsat ITS), that \emph{incrementally} builds the test suite with the help of a MaxSAT query that aims to maximize the coverage of allowed tuples at every step.

The Covering Array Number problem is concerned with reporting solutions with the least number of tests. From a practical point of view, whether we are satisfied with suboptimal solutions will depend on the cost of the tests. This cost basically includes the cost of generating the tests (computational resources) and the cost of testing the system. In particular, when the cost is too prohibitive in terms of our budget, and we are satisfied with covering a statistically significant portion of the tuples, we aim to solve (even suboptimally) the Tuple Number problem. Therefore, there exist real-world scenarios where all the approaches described in this paper are of practical interest.

The rest of the paper is structured as follows: section \ref{sec:prelim} introduces definitions on CAs, SAT/MaxSAT instances, constraints and SAT solvers. For computing MCACs of a given length, section \ref{sec:mcacSAT} defines different SAT encodings and sections \ref{sec:preproc} and \ref{sec:symmetry} describe techniques to make the SAT encodings more efficient. Section \ref{sec:calot} introduces the incremental SAT algorithm CALOT for computing minimum MCACs. Subsequently, section \ref{sec:mcacMaxSAT} defines MaxSAT encodings and section \ref{sec:MaxSATbased} describes how to efficiently apply MaxSAT solvers.  For the Tuple Number problem, section \ref{sec:shortening} defines a MaxSAT encoding and section \ref{sec:incompleteTN} presents a new incomplete approach using MaxSAT solvers. To assess the impact of the presented approaches, section \ref{sec:experimental} reports on an extensive experimental investigation on the available MCAC benchmarks. Finally, section \ref{sec:conclusions} concludes the paper.

\section{Preliminaries}\label{sec:prelim}

\begin{definition}
A System Under Test (SUT) model is a tuple $\langle P,\varphi \rangle$, where 
$P$ is a finite set of variables $p$ of finite domain, called SUT parameters, and $\varphi$ is a set of constraints on $P$, called SUT constraints, that implicitly represents the parameterizations that the system accepts. 
We denote by $d(p)$ and $g_p$, respectively, the domain and the domain cardinality of $p$.
For the sake of clarity, we will assume that the system accepts at least one parameterization.    
\end{definition}

In the following, we assume $S=\langle P,\varphi \rangle$ to be a SUT model. We will refer to $P$ as $S_P$, and to $\varphi$ as $S_{\varphi}$.

\begin{definition} An assignment is a set of pairs $(p,v)$ where $p$ is a variable and $v$ is a value of the domain of $p$. A test case for $S$ is a full assignment $A$ to the variables in $S_P$ such that $A$ entails $S_\varphi$  (i.e. $A \models S_\varphi$) . A parameter tuple of $S$ is a subset $\pi \subseteq S_P$. A value tuple of $S$ is a partial assignment to $S_P$; in particular, we refer to a value tuple of length $t$ as a $t$-tuple. 
\end{definition}

\begin{definition}
A $t$-tuple $\tau$ is forbidden if $\tau$ does not entail $S_\varphi$ (i.e. $\tau \models \neg S_\varphi$). Otherwise, it is allowed. We refer to the set of allowed $t$-tuples as $\Tau_a^{t,S} = \{\tau\ |\ \tau \not\models \neg S_\varphi\}$, to the set of forbidden $t$-tuples as $\Tau_f^{t,S} = \{\tau\ |\ \tau \models \neg S_\varphi\}$, and to the whole set of $t$-tuples in the SUT model $S$ as $\Tau^{t,S} = \Tau_a \cup \Tau_f$.\\[2mm]
When there is no ambiguity, we refer to $\Tau_a^{t,S},\Tau_f^{t,S},\Tau^{t,S}$ as $\Tau_a,\Tau_f,\Tau$, respectively.
\end{definition}

\begin{definition} A test case $\upsilon$ \emph{covers} a value tuple $\tau$ if both assign the same domain value to the variables in the value tuple, i.e., $\upsilon \models \tau$. 
\end{definition}

\begin{definition}
A Mixed Covering Array with Constraints (MCAC), denoted by $CA(N;t,S)$, is a set of $N$ test cases for a SUT model $S$ such that all $t$-tuples are at least covered by one test case. The term \emph{Mixed} reflects that the domains of the parameters in $S_P$ are allowed to have different cardinalities. The term \emph{Constraints} reflects that $S_\varphi$ is not empty.%
\end{definition}

\begin{definition}
The Covering Array Number, $CAN(t,S)$, is the minimum $N$ for which there exists an MCAC $CA(N;t,S)$.
An upper bound $ub^{CAN(t,S)}$ for $CAN(t,S)$ is an integer such that $ub^{CAN(t,S)} \geq CAN(t,S)$, and a lower bound $lb^{CAN(t,S)}$ is an integer such that $CAN(t,S) > lb^{CAN(t,S)}$.\\[2mm]
When there is no ambiguity, we refer to $ub^{CAN(t,S)}$ ($lb^{CAN(t,S)}$) as $ub$ ($lb$).

\end{definition}

\begin{definition}
The Tuple Number, $T(N;t,S)$, is the maximum number of $t$-tuples that can be covered by a set of $N$ tests for a SUT model $S$.
An upper bound $ub^{T(N;t,S)}$ for $T(N;t,S)$ is an integer such that $ub^{T(N;t,S)} \geq T(N;t,S)$, and a lower bound $lb^{T(N;t,S)}$ is an integer such that $T(N;t,S) > lb^{T(N;t,S)}$.\\[2mm]
When there is no ambiguity, we refer to $ub^{T(N;t,S)}$ ($lb^{T(N;t,S)}$) as $ub$ ($lb$).
\end{definition}

\begin{definition} The MCAC problem is to find an MCAC of size $N$.\\
The Covering Array Number problem is to find an MCAC of size $CAN(t,S)$.\\
The Tuple Number problem is to find a test suite of size $N$  that covers $T(N;t,S)$ $t$-tuples.\\[2mm]
The MCAC problem is a decision problem.
The Covering Array Number and the Tuple Number problems, to which we refer in short as the $CAN(t,S)$ and $T(N;t,S)$ problems, respectively, are optimization problems.
\end{definition}

\begin{definition} A literal is a propositional variable~$x$ or a negated propositional variable~$\neg x$. A clause is a disjunction of literals. A formula in Conjunctive Normal Form (CNF) is a conjunction of clauses.
\end{definition}

\begin{definition} A weighted clause is a pair $(c,w)$, where $c$ is a clause and $w$, its weight, is a natural number or infinity. A clause is hard if its weight is infinity (or no weight is given); otherwise, it is soft. A Weighted Partial MaxSAT instance is a multiset of weighted clauses.
\end{definition}

\begin{definition} A truth assignment for an instance $\phi$ is a mapping that assigns to each propositional variable in $\phi$ either 0 (False) or 1 (True). A truth assignment is \emph{partial} if the mapping is not defined for all the propositional variables in $\phi$.
\end{definition}

\begin{definition}
A truth assignment $I$ satisfies a literal $x$ $(\neg x)$ if $I$ maps $x$ to 1 (0); otherwise, it is falsified. A truth assignment $I$ satisfies a clause if $I$ satisfies at least one of its literals; otherwise, it is violated or falsified. The cost of a clause $(c,w)$ under $I$ is 0 if $I$ satisfies the clause; otherwise, it is $w$.
Given a partial truth assignment $I$, a literal or a clause is undefined if it is neither satisfied nor falsified.  A clause $c$ is a unit clause under $I$ if $c$ is not satisfied by $I$ and contains exactly one undefined literal.  
\end{definition}

\begin{definition}
The cost of a formula $\phi$ under a truth assignment $I$, denoted by $cost(I, \phi)$, is the aggregated cost of all its clauses under $I$.
\end{definition}

\begin{definition}
The Weighted Partial MaxSAT (WPMaxSAT) problem for an instance $\phi$ is to find an assignment in which the sum of weights of the falsified soft clauses is minimal (referred to as the optimal cost of $\varphi$) and all the hard clauses are satisfied. The Partial MaxSAT problem is the WPMaxSAT problem when all the soft clauses have the same weight. The MaxSAT problem is the Partial MaxSAT problem when there are no hard clauses. The SAT problem is the Partial MaxSAT problem when there are no soft clauses. 
\end{definition}

\begin{definition}
An instance of Weighted Partial MaxSAT, or any of its variants, is unsatisfiable if its optimal cost is $\infty$.
A SAT instance $\phi$ is satisfiable if there is a truth assignment $I$, called model, such that $cost(I, \phi) = 0$.
\end{definition}

\begin{definition}
An unsatisfiable core is a subset of clauses of a SAT instance that is unsatisfiable.  
\end{definition}

\begin{definition}
Given a SAT instance $\phi$ and a partial truth assignment $I$, we refer as Unit Propagation, denoted by $UP(I,\phi)$, to the Boolean inference mechanism (propagator) defined as follows:  Find a unit clause in $\phi$ under $I$, where $l$ is the undefined literal. Then, propagate the unit clause, i.e. extend $I$ with $x=1$ ($x=0$) if $l \equiv x$ ($l \equiv \neg x$) and repeat the process until a fixpoint is reached or a conflict is derived (i.e. a clause in $\phi$ is falsified by $I$).\\[2mm]
We refer to $UP(I,\phi)$ simply as $UP(\phi)$ when $I$ is empty.
\end{definition}

\begin{definition}
Let $A$ and $B$ be SAT instances.\\
 $A \models B$ denotes that $A$ entails $B$, i.e. all assignments satisfying $A$ also satisfy $B$. \\
It holds that $A \models B$ iff $A \wedge \neg B$ is unsatisfiable.\\
$A \vdash_{UP} B$ denotes that, for every clause $c \in B$, $UP(A \wedge \neg c)$ derives a conflict. \\
If $A \vdash_{UP} B$ then $A \models B$.
\end{definition}

\begin{definition} 
  A \emph{pseudo-Boolean} (PB) constraint is a Boolean function of the  form $ \sum_{i=1}^{n} q_il_i \diamond k $, where $k$ and the $q_i$ are  integer constants, $l_i$ are literals, and
  $\diamond \in \{<, \leq, =, \geq, >\}$.
\end{definition}

\begin{definition} A Cardinality (Card) constraint is a PB constraint where all $q_i$ are equal to 1.

\end{definition}

\begin{definition} An \emph{At-Most-One} (AMO) constraint is a cardinality constraint of the form $\sum_{i=1}^{n} l_i \leq 1 $.
\end{definition}

\begin{definition} An \emph{At-Least-One} (ALO) constraint is a cardinality constraint of the
  form $\sum_{i=1}^{n} l_i \geq 1 $.
\end{definition}

\begin{definition} An \emph{Exactly-One} (EO) constraint is a cardinality constraint of the form {$\sum_{i=1}^{n} l_i=1$}.
\end{definition}

\newcommand{\satcomm}[1]{\textsf{\scriptsize #1}}
\begin{algorithm}[h!t]%
    \SetAlgorithmName{Code}{SATSolver}{}
    \SetAlgoRefName{SATSolver}
    \caption{Members and functions interface}
    \label{alg:extra_funcs}
    \SetKw{KwIfT}{?}
    \SetKw{KwElseT}{:}
    \SetKw{KwInput}{Input:}
    \SetKw{KwAnd}{and}
    \SetKwProg{KwFunc}{function}{}{}
    \SetKwComment{KwComm}{\#}{}
    
    \SetKw{KwIff}{if}
    \SetKw{KwThenn}{then}
    \SetKw{KwElsee}{else}
    \DontPrintSemicolon
    
    \KwComm{\textbf{Attributes}}
    $n\_vars$ \KwComm{\satcomm{number of variables of the formula loaded}}
    $core$ \KwComm{\satcomm{last core found}}
    $model$ \KwComm{\satcomm{last model found}}
    
    \KwComm{\textbf{Methods}}
    
    \KwFunc{$assume(x : literal)$}{
       \KwComm{\satcomm{Sets the literal $x$ in the solver trail}}
    }
    \KwFunc{$add\_clause(c : clause)$}{
       \KwComm{\satcomm{Adds the clause $c$ to the solver}}
    }
    \KwFunc{$add\_retractable(c : clause)$}{
       \KwComm{\satcomm{Adds the clause $c$ to the retractable list of clauses of the solver}}
    }
    \KwFunc{$retract\_clause(c : clause)$}{
       \KwComm{\satcomm{Retracts the clause $c$ from the solver's list of retractable clauses}}
    }
    
    \KwFunc{$solve()$}{
        \KwComm{\satcomm{If formula is satisfiable, $status \leftarrow SAT, sat.model$ is updated}}
        \KwComm{\satcomm{If formula is unsatisfiable, $status \leftarrow UNSAT, sat.core$ is updated}}
        \KwRet{status}
    }
    
    \KwFunc {$add( \phi : \mbox{SAT formula} )$} {
        \lForEach{$c_i \in \phi$}{$sat.add\_clause(c_i)$}
    }
    \KwFunc {$retract(\phi  : \mbox{SAT formula} )$} {
        \lForEach{$c_i \in \phi$}{ $sat.retract\_clause(c_i)$}
    }
    
    \KwComm{\textbf{Overloaded functions for SAT-based MaxSAT algorithms}}
    
    \KwFunc {$add( \phi : \mbox{Weighted Partial MaxSAT formula} ) $} {
        \ForEach{$(c_i, w_i) \in \phi$}{
        \leIf{$w_i = \infty$}{$sat.add\_clause(c_i)$}{$sat.add\_retractable(c_i)$}
        }
    }
    \KwFunc {$retract(\phi  : \mbox{Weighted Partial MaxSAT formula})$} {
        \lForEach{$(c_i, w_i) \in \phi$}{
            \KwIff $w_i \neq \infty$ \KwThenn $sat.retract\_clause(c_i)$
        }
    }
\end{algorithm}
\let\satcomm\undefined

The interface of a modern SAT solver is presented in code fragment \ref{alg:extra_funcs}. The input instance is added to the solver with functions $add\_clause$ and $add\_retractable$ (in case the clause can be retracted) (lines 5-7), which operate on a single clause, while functions $add$ and $retract$ operate on a set of clauses. The last two functions are overloaded to ease the usage of SAT solvers within MaxSAT solvers (lines 10-13 and 14-18).  Variable $n\_vars$ indicates the number of variables of the input formula (line 1).

Function $solve$ (lines 8-9) returns UNSAT (SAT) if the input formula is unsatisfiable (satisfiable) and sets variable $core$ ($model$) to the corresponding unsatisfiable core (model). Function $assume$ (line 4) allows to place an \emph{assumption} on the truth value of a literal before function $solve$ is called. Finally, modern SAT solvers also support an incremental solving mode, which allows to keep the learnt clauses across calls to the function $solve$. 

\section{The MCAC problem as SAT}
\label{sec:mcacSAT}

In this section, we present the SAT encoding described in \cite{YamadaKACOB15} to decide whether there exists a $CA(N;t,S)$ for a given SUT model $S=\langle P,\varphi \rangle$. It is similar to previous encodings described in \cite{hnich05,hnich_constraint_2006,banbara10,Toru_NANBA2012,Ansotegui2013}.

In the following, we list the set of constraints that define the SAT encoding and describe the semantics of the propositional variables they refer to. To encode each constraint, we assume that AMO and EO cardinality constraints are translated into CNF through the regular encoding \cite{AnsoteguiM04,gent2004new} and the typical transformations \cite{Tseitin1983}  of propositional formulas into CNF are implicitly applied.

First, we define variables $x_{i,p,v}$ to be true iff test case $i$ assigns value $v$ to parameter~$p$, and state that each parameter in each test case takes exactly one value as follows (where $[N] = \{1,\ldots,N\}$):

\begin{equation}\label{EqEOXs}\tag{$X$}
\bigwedge_{i \in [N]} \bigwedge_{p \in P} \sum_{v \in d(p)} x_{i,p,v} = 1
\end{equation}

Second, as described in \cite{NanbaTK12}, in order to enforce the SUT constraints~$\varphi$, for each test case $i$, we add the CNF formula that encodes the constraints of~$\varphi$ into SAT and substitute each appearance of the pair $(p,v)$ in $\varphi$ by the corresponding literal on propositional variable $x_{i,p,v}$ for each test case $i$.

\begin{equation}\label{SUTX}\tag{$SUTX$}
\bigwedge_{i \in [N]} CNF\left(\varphi\left\{\frac{\neg x_{i,p,v}}{p \neq v},\frac{x_{i,p,v}}{p=v}\right\}\right)
\end{equation}

Third, we introduce propositional variables $c_{\tau}^{i}$ and state that if they are true, then tuple $\tau$ must be covered at test $i$, by forcing the variables $p$ in the test case to be assigned to the value specified in $\tau$, as follows:

\begin{equation}\label{EqCImpliesX}\tag{$CX$}
\bigwedge_{i \in [N]} \bigwedge_{\tau \in \Tau_a} \bigwedge_{(p,v) \in \tau} (c^{i}_{\tau} \rightarrow x_{i,p,v})
\end{equation}

Notice that only $t$-tuples that can be covered by a test case are encoded, i.e., $\tau \in \mathcal{T}_a$. In section \ref{sec:preproc}, we discuss how to detect the $t$-tuples forbidden by the SUT constraints.\\[1mm] %

Finally, we state that every $t$-tuple $\tau \in \Tau_a$, must be covered at least by one test case, as follows:

\begin{equation}\label{EqCovers}\tag{$C$}
\bigwedge_{\tau \in \Tau_a} \bigvee_{i \in [N]} c^{i}_{\tau}
\end{equation}

\begin{proposition}
Let $Sat_{CX}^{N,t,S}$ be $X \wedge C \wedge CX \wedge SUTX$. $Sat_{CX}^{N,t,S}$ is satisfiable iff  a $CA(N;t,S)$ exists.
\end{proposition}

Inspired by the incremental SAT approach in \cite{YamadaKACOB15} (see section~\ref{sec:calot}), we present another encoding where $C$ and $CX$ are replaced by $CCX$:

\begin{align}\label{CXAlg2}
\bigwedge_{i \in [N]} \bigwedge_{\tau \in \Tau_a} \bigwedge_{(p,v) \in \tau} & (c^{i}_{\tau} \rightarrow c^{i-1}_{\tau} \vee x_{i,p,v}) \tag*{(a) ($CCX$)} \\
\bigwedge_{\tau \in \Tau_a} & c^{N}_{\tau} \tag*{(b)\phantom{ ($CCX$)}} \\
\bigwedge_{\tau \in \Tau_a} & (c^{N}_{\tau} \rightarrow \neg \tag*{(c)\phantom{ ($CCX$)}} c^{0}_{\tau}) \nonumber
\end{align}

Variables $c^i_{\tau}$ have now a different semantics, i.e., if they are true, $\tau$ is covered by test case $i$ or by any lower  test case $j$, where $1 \leq j \leq i$ (equation a). In order to guarantee that $\tau$ will be covered by some test, notice that we just need to force  $c^{N}_{\tau}$ to be true and $c^{0}_{\tau}$ to be false (variables $c^{0}_{\tau}$ are additionally included in the encoding).  This can be achieved by adding the unit clauses $c^{N}_{\tau}$ (equation b) and the implication $c^{N}_{\tau} \rightarrow \neg c^{0}_{\tau}$ (equation c) for every allowed tuple $\tau$. 

The seasoned reader may wonder why we do not simply replace equation (c) by $\bigwedge_{\tau \in \Tau_a}  \neg c^{0}_{\tau} $.
Indeed, this is possible. First, notice that UP on the conjunction of equations (b) and (c) will derive exactly the same.
Second, for encoding some problems where it is not mandatory to cover all the tuples (see section \ref{sec:shortening}), we have to erase equation (b) from $CCX$ and also guarantee that if a tuple $\tau$ is not covered in an optimal solution, i.e., $c^{N}_{\tau}$ has to be False, then the related clauses in $CCX$ have to be satisfied (these are hard clauses) and, if possible, to be trivially satisfied, i.e., without requiring search. 
Equation (c) eases this case for all the scenarios in section \ref{sec:shortening}.
Notice that, once $c^{N}_{\tau}$ is False, clauses in equation (c) are trivially satisfied and, by setting the remaining $c^{i}_\tau$ vars to True, clauses in equation (a) are also trivially satisfied.

\begin{proposition}
Let $Sat_{CCX}^{N,t,S}$ be $X \wedge CCX \wedge SUTX$. $Sat_{CCX}^{N,t,S}$ is satisfiable iff a $CA(N;t,S)$ exists.
\end{proposition}

\begin{rmarkN}
\label{rmrk:ccx-alternative}
There are some variations of equation (a) in $CCX$ that can be beneficial when using some SAT solvers, as we will see in section \ref{sec:exp-inc}. For example, we can use full implication instead of half implication in equation (a), i.e., $(c^{i}_{\tau} \leftrightarrow c^{i-1}_{\tau} \vee x_{i,p,v})$, or we can even use $(c^{i}_{\tau} \rightarrow c^{i-1}_{\tau} \vee x_{i,p,v}) \wedge (c^{i}_{\tau} \leftarrow c^{i-1}_{\tau})$. Also, we can consider full implication in equation (c) and, for some of the problems analyzed in section \ref{sec:exp-inc}, we can even replace equation (c) by $\bigwedge_{\tau \in \Tau_a}  \neg c^{0}_{\tau} $.
\end{rmarkN}

\begin{xampl}
\label{ex:autodriving-sat}
As an example of SUT problem, we focus on the domain of 
autonomous driving. Table \ref{tab:sut} shows the parameters and values, $S_P$, and the SUT constraints, $S_\varphi$:

\begin{table}[ht]
    \centering
    \begin{tabular}{cc||c}
        \toprule
        $P \in S_P$ & Abbrv. & Values \\
        \midrule
        Luminosity & L & Day (dy), Night (ni) \\
        Environment & E & Highway (hw), Urban (ur), Country (co) \\
        Motor & M & Combustion (cb), Electric (el) \\
        Sensor & S & Camera (ca), Radar (ra), Lidar (li) \\
        \midrule
        \multicolumn{3}{c}{$S_\varphi$} \\ 
        \midrule
        \multicolumn{3}{c}{$((L = ni) \land (E = co)) \rightarrow (S \neq ca)$}\\
        \multicolumn{3}{c}{$((E = hw) \lor (E = co)) \rightarrow (S \neq li)$}\\
        \multicolumn{3}{c}{$(M = el) \rightarrow (E = ur)$} \\
        \bottomrule
    \end{tabular}
    \caption{Example of Autonomous Driving System Under Test.}
    \label{tab:sut}
\end{table}

We show how to build $Sat_{CX}^{N=10,t=2,S}$, where $N=10$ is an upper bound $ub$ for this SUT (see section \ref{sec:preproc} and Example \ref{ex:autonomous-forbidden-tupls}).

To encode the $X$ constraint, we add:

\begin{align}
   x_{1,L,dy} + x_{1,L,ni} =   &1,  & \mspace{-54.0mu} x_{1,E,hw} + x_{1,E,ur} + x_{1,E,co} =    &1, \notag\\
   x_{1,M,cb} + x_{1,M,el} =   &1,  & \mspace{-54.0mu} x_{1,S,ca} + x_{1,S,ra} + x_{1,S,li} =    &1{} \notag \\
                               &    & \mspace{-54.0mu}\vdots \hspace{3.8cm}                      &  \tag{Ex. X} \\
   x_{10,L,dy} + x_{10,L,ni} = &1,  & \mspace{-54.0mu} x_{10,E,hw} + x_{10,E,ur} + x_{10,E,co} = &1, \notag\\
   x_{10,M,cb} + x_{10,M,el} = &1,  & \mspace{-54.0mu} x_{10,S,ca} + x_{10,S,ra} + x_{10,S,li} = &1{} \notag 
\end{align}

Next, for each test $(1,...,10)$, we encode the SUT constraints $SUTX$:

\begin{gather}
(x_{1,L,ni} \land x_{1,E,co}) \rightarrow \neg x_{1,S,ca} \notag \\
(x_{1,E,hw} \lor x_{1,E,co}) \rightarrow \neg x_{1,S,li} \notag \\
x_{1,M,el} \rightarrow x_{1,E,ur} \notag \\
\vdots \tag{Ex. SUTX} \\
(x_{10,L,ni} \land x_{10,E,co}) \rightarrow \neg x_{10,S,ca} \notag \\
(x_{10,E,hw} \lor x_{10,E,co}) \rightarrow \neg x_{10,S,li} \notag \\
x_{10,M,el} \rightarrow x_{10,E,ur} \notag
\end{gather}

Finally, the encoding of the $CX$ and $C$ constraints is shown below. We identify the set of allowed tuples, ($\Tau_a$), as described in section \ref{sec:preproc}. In particular, there are 
$|\Tau_a|=33$ allowed tuples.

\begin{gather}
    c_{\tau_1}^1 \rightarrow x_{1,L,dy}, \quad \ldots, \quad c_{\tau_{33}}^1 \rightarrow x_{1,M,el} \notag \\
    c_{\tau_1}^1 \rightarrow x_{1,E,hw}, \quad \ldots, \quad c_{\tau_{33}}^1 \rightarrow x_{1,S,li} \notag \\
    \vdots \qquad\qquad \ddots \qquad\qquad \vdots \tag{Ex. CX} \\
    c_{\tau_1}^{10} \rightarrow x_{10,L,dy}, \quad \ldots, \quad c_{\tau_{33}}^{10} \rightarrow x_{10,M,el} \notag \\
    c_{\tau_1}^{10} \rightarrow x_{10,E,hw}, \quad \ldots, \quad c_{\tau_{33}}^{10} \rightarrow x_{10,S,li} \notag
\end{gather}

\begin{equation}
    (c_{\tau_1}^1 \lor c_{\tau_1}^2 \lor \cdots \lor c_{\tau_1}^{10}), \quad \ldots, \quad (c_{\tau_{33}}^1 \lor c_{\tau_{33}}^2 \lor \cdots \lor c_{\tau_{33}}^{10}) \tag{Ex. C}
\end{equation}

To build $Sat_{CCX}^{N=10,t=2,S}$, we encode the $CCX$ constraint instead of the $C$ and $CX$ constraints:

\begin{gather}
    c_{\tau_1}^1 \rightarrow c_{\tau_1}^0 \lor x_{1,L,dy}, \quad \ldots, \quad c_{\tau_{33}}^1 \rightarrow c_{\tau_{33}}^0 \lor x_{1,M,el} \notag \\
    c_{\tau_1}^1 \rightarrow c_{\tau_1}^0 \lor x_{1,E,hw}, \quad \ldots, \quad c_{\tau_{33}}^1 \rightarrow c_{\tau_{33}}^0 \lor x_{1,S,li} \notag \\
    \vdots \qquad\qquad \ddots \qquad\qquad \vdots \tag{Ex. CCX a} \\
    c_{\tau_1}^{10} \rightarrow c_{\tau_1}^{9} \lor x_{10,L,dy}, \quad \ldots, \quad c_{\tau_{33}}^{10} \rightarrow c_{\tau_{33}}^9 \lor x_{10,M,el} \notag \\
    c_{\tau_1}^{10} \rightarrow c_{\tau_1}^{9} \lor x_{10,E,hw}, \quad \ldots, \quad c_{\tau_{33}}^{10} \rightarrow c_{\tau_{33}}^9 \lor x_{10,S,li} \notag \\
    c_{\tau_1}^{10}, \quad \ldots, \quad c_{\tau_{33}}^{10} \tag{Ex. CCX b} \\
    c_{\tau_1}^{10} \rightarrow \neg c_{\tau_1}^0, \quad \ldots, \quad c_{\tau_{33}}^{10} \rightarrow \neg c_{\tau_{33}}^0 \tag{Ex. CCX c}
\end{gather}

Once we run a SAT solver on any of the previous SAT instances, if there exists a $CA(10;2,S)$, it will return a satisfying truth assignment.
To recover the particular $CA(10;2,S)$ implicitly found by the solver, we just need to check the assignment to the $x_{i,p,v}$ variables. For example, if $x_{1,L,dy}$ is True then parameter $Luminosity$ takes value $day$ at test $1$. Table \ref{tab:ex_ca_res_sat} shows the result of this conversion.

\begin{table}[ht]
    \centering
    \begin{tabular}{l|@{\hskip 2mm}c@{\hskip 2mm}|@{\hskip 2mm}c@{\hskip 2mm}|@{\hskip 2mm}c@{\hskip 2mm}|@{\hskip 2mm}c@{\hskip 2mm}}
        \toprule
         & L & E & S & M \\
        \midrule
        $t_1$ & dy & hw & ca & cb \\
        $t_2$ & ni & hw & ra & cb \\
        $t_3$ & ni & ur & ca & el \\
        $t_4$ & dy & ur & ra & cb \\
        $t_5$ & dy & ur & li & cb \\
        $t_6$ & dy & co & ca & cb \\
        $t_7$ & ni & co & ra & cb \\
        $t_8$ & dy & ur & ra & el \\
        $t_9$ & ni & ur & li & cb \\
        $t_{10}$ & ni & ur & li & el \\
        \bottomrule
    \end{tabular}
    \caption{$CA(10;2,S)$ for the autonomous driving SUT.}
    \label{tab:ex_ca_res_sat}
\end{table}

\end{xampl}

\section{Preprocessing for the MCAC problem}\label{sec:preproc}

In the context of the Covering Array Number problem, we define an upper bound $ub$ and a lower bound $lb$ to be integers such that $ub \geq CAN(t,S) > lb$. When $ub = lb +1$, we can stop the search and report $ub$ as the minimum covering array number $CAN(t,S)$. 

To get an initial value for $ub$, we can execute a greedy approach to obtain a suboptimal $CA(N;t,S)$ and set $ub$ to $N$.  For example, in the experiments, we use the tool ACTS \cite{BorazjanyYLKK12} that supports Mixed Covering Arrays with Constraints. Moreover, a lower $ub$ also implies a smaller initial encoding.

Additionally, by inspecting the solution, i.e., the test cases that certify the suboptimal $CA(N;t,S)$, we can compute which tuples are not covered, the set of \emph{forbidden} tuples, since the suboptimal $CA(N;t,S)$ guarantees to cover all allowed $t$-tuples.

Furthermore, let $r$ be the maximum number of allowed $t$-tuples associated to any parameter tuple of length $t$. Then, we can set $lb = r-1$, since these $r$ value tuples (mutually exclusive) need to be covered by different test cases.\\

\begin{algorithm}[ht]
    \SetAlgoRefName{ForbiddenTuples}
    \caption{Detection of forbidden tuples. \label{ForbiddenAlg}}
    \SetKw{KwIfT}{?}
    \SetKw{KwElseT}{:}
    \SetKw{KwInput}{Input:}
    \DontPrintSemicolon
    
    \KwInput{SUT model $S$, SAT solver $sat$}
    
    $sat.add(Sat_{CX}^{N=1,t,S}[\mbox{\ref{EqEOXs}}, \mbox{\ref{SUTX}}])$\;
    
    $\Tau_f = \emptyset$\;
    
    \For{$\tau \in \Tau$}{
        \For{$(p,v) \in \tau$}{
            $sat.assume(x_{1,p,v})$\;
        }
        \lIf{$sat.solve()=\mbox{UNSAT}$} {$\Tau_f \leftarrow \Tau_f \cup \tau$}
    }
    \Return{$\Tau_f$}
\end{algorithm}

Using an approach like ACTS, not based on constraint programming techniques, has a drawback. It may not be efficient enough if testing the satisfiability of $\phi$ (the set of SUT constraints) is computationally hard. In this case, to detect the forbidden tuples, we can simply apply algorithm \ref{ForbiddenAlg}. This algorithm tests, for every tuple $\tau$ (lines 4-7), if it is compatible with the SUT constraints (line 2) through a SAT query; if the solver results in unsatisfiability (line 7), the tuple is added to the set of forbidden tuples $\Tau_f$, which is ultimately returned by the algorithm (line 8).

For $t=2$, which is already of practical importance \cite{KuhnWG04}, the experiments carried out in this paper show that this detection process is negligible runtime-wise. 

\begin{xampl}
\label{ex:autonomous-forbidden-tupls}
We show the result of algorithm \ref{ForbiddenAlg} applied to the autonomous driving SUT presented in Example \ref{ex:autodriving-sat}. It yields the following forbidden tuples $\Tau_f$ for $t=2$:
$$\Tau_f = \{(E=co, M=el),(E=hw,S=li),(E=hw,M=el),(E=co,S=li)\}$$

Then, the first parameter tuple with more allowed tuples, according to \ref{ForbiddenAlg}, would be $(E, S)$. It has 7 allowed tuples, implying that $lb = 6$.

\end{xampl}

\section{Symmetry Breaking for the MCAC problem}\label{sec:symmetry}

As \cite{YamadaKACOB15}, we fix the $r$ $t$-tuples that conducted us to set the initial $lb$ (see section \ref{sec:preproc}) to test cases $\{1,\ldots,r\}$. This helps us break row symmetries for the first~$r$ test cases. We will refer to this as fixed-tuple symmetry breaking.

There are other alternatives. We can impose row symmetry breaking constraints as \cite{FlenerFHKMPW02}; since each row (test) represents a number in base 2, we can add constraints to order the tests in monotonic increasing order, from test $0$ to test $N-1$. 
We can also apply, as explained above, fixed-tuple symmetry breaking to the first~$r$ tuples (first partition) and apply row symmetry breaking constraints to the remaining $ub-lb+1$ test cases (second partition). Furthermore, we can impose an order among the tuples in the first partition and the second partition, so that if two sets share the same value for the fixed tuple, then the one representing the lower number must be in the first partition.

Our experimental analysis shows that fixed-tuple symmetry breaking is superior to any other of the mentioned alternatives. For lack of space, we restricted all the experiments to this symmetry breaking approach.

\begin{xampl}
\label{ex:autonomous-sym}
We show how to apply symmetry breaking to the SUT in Example \ref{ex:autodriving-sat}.

Recall that $(E, S)$ was the parameter tuple with the largest number of allowed tuples we selected. The set of allowed value tuples is:
$\{\tau_1 = (E=hw,S=ca), \tau_2 = (E=hw,S=ra), \tau_3 = (E=ur,S=ca),\tau_4 = (E=ur,S=ra), \tau_5 = (E=ur,S=li), \tau_6 = (E=co,S=ca), \tau_7 = (E=co,S=ra)\}$.

\filbreak
To apply the fixed-tuple symmetry breaking variant, we just need to fix each allowed value tuple in a different test as shown below:

\begin{gather}
    x_{1,E,hw} \land x_{1,S,ca} \notag \\
    x_{2,E,hw} \land x_{2,S,ra} \notag \\
    x_{3,E,ur} \land x_{3,S,ca} \notag \\
    x_{4,E,ur} \land x_{4,S,ra} \tag{Ex. SYM X} \\
    x_{5,E,ur} \land x_{5,S,li} \notag \\
    x_{6,E,co} \land x_{6,S,ca} \notag \\
    x_{7,E,co} \land x_{7,S,ra} \notag
\end{gather}

\end{xampl}

\section{Solving the $CAN(t,S)$ problem with Incremental SAT}\label{sec:calot}

In this section, we present the \ref{alg:calot} algorithm, which is an incremental SAT approach for computing optimal covering arrays with SUT constraints described by \cite{YamadaKACOB15}. The input to the algorithm is an upper bound $ub$ (computed as in section \ref{sec:preproc}), the strength $t$ and the SUT model $S$.  In line 2, the incremental SAT solver is initialized with the SAT instance $Sat_{CCX}^{N=ub,t,S}$. Additionally, breaking symmetries for the first $lb +1$ tuples, as described in section \ref{sec:symmetry}, are added to the SAT solver. The output is the covering array number and an optimal model. 

\begin{algorithm}[ht]
    \SetAlgoRefName{CALOT}
    \caption{Algorithm 2 in \cite{YamadaKACOB15}}
    \label{alg:calot}
    \SetKw{KwIfT}{?}
    \SetKw{KwElseT}{:}
    \SetKw{KwInput}{Input:}
    \DontPrintSemicolon
    
    \KwInput{upper bound $ub$, strength $t$, SUT model $S$}
    
    $sat.add(Sat_{CCX}^{N=ub,t,S})$\;
    
    Fix $lb+1$ value tuples to break symmetries (see Section \ref{sec:symmetry})\;

    $b_{model} \leftarrow \emptyset$
    
    \For{$i=N,...,lb+1$} {
    
        \lIf{$sat.solve()=\mbox{UNSAT}$} {\Return{$(i, b_{model})$}}
        
        $sat.add(\bigwedge_{\tau \in \Tau} c^{i-1}_{\tau})$\;
        
        $b_{model} \leftarrow sat.model$\;
        
        \For{$\tau \in \Tau_a$}{
            \lFor{$(p,v) \in \tau$}{
                {$b_{model}[x_{i,p,v}]$
                \KwIfT $sat.add(\{x_{i,p,v}\})$ %
                \KwElseT $sat.add(\{\neg x_{i,p,v}\})$} %
            }
        }
    }
    \Return{$(lb+1, b_{model})$}\;
\end{algorithm}

The algorithm works in a top-down search manner by iteratively decreasing the $ub$ till it reaches $lb+1$ (line 5) or the current SAT instance is unsatisfiable (line 6). To decrease the $ub$ by one, the algorithm adds the set of unit clauses $\bigwedge_{\tau \in \Tau_a} c^{i-1}_{\tau}$ (line 7), which state that every $t$-tuple is covered by a test case with an index smaller
than~$i$.

There is a subtle detail in lines 9 and 10. Whenever the algorithm finds a new upper bound, variables $x_{i,p,v}$ related to the previous upper bound are fixed to the value in the last model found ($b_{model}$ in line 8), so that 
these variables do not need to be decided in the next iterations. As \cite{YamadaKACOB15} report, %
not fixing these variables 
can have some negative impact on the performance.\\[1mm]

\begin{rmarkN}
The original \cite{YamadaKACOB15}'s algorithm pseudocode is slightly different. First, it assigns the $i$-th test at iteration $i$ to the value it had in the previous model found instead of assigning the $i+1$-th test. This does not correspond to the description given in the text of the paper and may lead to an incomplete algorithm.
Second, the set of constraints \ref{CXAlg2}, described in \cite{YamadaKACOB15}, does not set $c^{N}_{\tau}$ to True as we do in this paper, which makes the pseudocode perform a dummy first step that can cause to report a wrong optimum. We think that these are merely errors in the description, and we have fixed them. Since the tool CALOT is not available from the authors for reproducibility, we have tried to do our best to reproduce (or extend) the idea behind their work.
\end{rmarkN}

In section \ref{sec:MaxSATbased}, we will see that this SAT incremental approach resembles how SAT-based MaxSAT algorithms behave \cite{AnsoteguiBL13,MorgadoHLPM13}. Actually, in contrast to \cite{YamadaKACOB15}, we show that MaxSAT technology can be effectively applied to solve Covering Arrays.

\section{The $CAN(t,S)$ problem as Partial MaxSAT}\label{sec:mcacMaxSAT}

\cite{AnsoteguiIMJ13} proposes an encoding into Partial MaxSAT to build covering arrays without constraints of minimum size.  The main idea is to use an indicator variable $u_i$ that is True iff  test case $i$ is used to build the covering array. The objective function of the optimization problem, which aims to minimize the number of variables $u_i$ set to True, is encoded into Partial MaxSAT by adding the following set of soft clauses: 

\begin{equation}\label{SOFTU}\tag{$SoftU$}
\bigwedge_{i \in [lb+2 \ldots N]}(\neg u_i, 1)
\end{equation}
Notice that we only need to use $N-(lb+1)$ indicator variables since we know that the covering array will have at least $lb+1$ tests (see section \ref{sec:preproc}).

To avoid symmetries, it is also enforced that if test case $i+1$ belongs to the minimum covering array, so does the previous test case $i$: 

\begin{equation}\label{BSU}\tag{$BSU$}
\bigwedge_{i \in [lb+2 \ldots N-1]} (u_{i+1} \rightarrow u_i)
\end{equation}

Then, variables $u_i$ are connected to variables $c^i_{\tau}$, expressing that if we want test $i$ to be the proof that $\tau$ is covered, then test $i$ must be in the optimal solution \footnote{Notice that $\tau$ could be covered by other tests but the respective $c^{i}_{\tau}$ vars be False.}:

\begin{equation}\label{EqCImpliesU}\tag{$CU$}
\bigwedge_{i \in [lb+2 \ldots N]} \bigwedge_{\tau \in \Tau_a}  (c^{i}_{\tau} \rightarrow u_{i})
\end{equation}
\begin{proposition}
Let $PMSat_{CX}^{N,t,S,lb}$ be $SoftU \wedge BSU \wedge CU \wedge Sat_{CX}^{N,t,S}$. If $N \geq CAN(t,S)$, the optimal cost of the Partial MaxSAT instance $PMSat_{CX}^{N,t,S,lb}$ is $CAN(t,S)-(lb+1)$, otherwise it is $\infty$.
\end{proposition}

In order to build the Partial MaxSAT version of $Sat_{CCX}^{N,t,S}$, we just need to change how variables $u_i$ are related to variables $c^{i}_{\tau}$. This constraint reflects that if $u_i$ is False (i.e., test $i$ is not in the solution and, therefore, due to constraint $BSU$, none of the tests $>i$ cannot be in the solution either), then the tuple $\tau$ has to be covered at some test below $i$:

\begin{equation}\label{EqCCXImpliesU}\tag{$CCU$}
\bigwedge_{i \in [lb+2 \ldots N]} \bigwedge_{\tau \in \Tau_a}  (\neg c^{i-1}_{\tau} \rightarrow u_{i})
\end{equation}

\begin{proposition}
Let $PMSat_{CCX}^{N,t,S,lb}$ be $SoftU \wedge BSU \wedge CCU \wedge Sat_{CCX}^{N,t,S}$. If $N \geq CAN(t,S)$, the optimal cost of the Partial MaxSAT instance $PMSat_{CCX}^{N,t,S,lb}$ is $CAN(t,S)-(lb+1)$, otherwise it is $\infty$.
\end{proposition}

\begin{rmarkN}
In \cite{AnsoteguiIMJ13}, variables $u_i$ are instead connected to variables $x_{i,p,v}$ in the following way:
\begin{equation}\label{EqXImpliesU}\tag{$XU$}
\bigwedge_{i \in [N]} \bigwedge_{p \in P} (u_i \leftrightarrow \bigvee_{v \in d(p)} x_{i,p,v}) 
\end{equation}
This is a more compact encoding but it requires equation~\ref{EqEOXs} to use an AMO constraint instead of an EO constraint. 
\end{rmarkN}

Finally, we can convert these Partial MaxSAT instances into Weighted Partial MaxSAT modifying $SoftU$ as follows:

\begin{equation}\label{WSOFTU}\tag{$WSoftU$}
\bigwedge_{i \in [lb+2 \ldots N]}(\neg u_i, w_i) 
\end{equation}

If we use $w_i = 2^{i-(lb+2)}$ we naturally introduce a lexicographical preference in the soft constraints. This is a key detail to alter the behaviour of SAT-based MaxSAT algorithms when solving Covering Arrays. If the MaxSAT solver applies the stratified approach \cite{ansotegui_improving_2012} (see for more details section \ref{sec:MaxSATbased}), it suffices to use $w_i = i-(lb+2)+1$, i.e., to increase the weights linearly. This is of interest since a high number of tests in $WSoftU$ can result into too large  weights for some MaxSAT solvers.

\begin{proposition}
Let $WPMSat_{CCX}^{N,t,S,lb}$ be $WSoftU \wedge BSU \wedge CCU \wedge Sat_{CCX}^{N,t,S}$. 

If $N \geq CAN(t,S)$ and $w_i = 2^{i-(lb+2)}$  the optimal cost of the Partial MaxSAT instance $PMSat_{CCX}^{N,t,S,lb}$ is $2^{CAN(t,S) - (lb + 1)} -1$, otherwise it is $\infty$.

If $N \geq CAN(t,S)$ and $w_i = i-(lb+2)+1$ the optimal cost of the Partial MaxSAT instance $PMSat_{CCX}^{N,t,S,lb}$ is $(1+n) \cdot n/2$ where $n=CAN(t,S) - (lb + 1)$, otherwise it is $\infty$.

\end{proposition}

\begin{xampl}
\label{ex:autonomous-pmsat}

We extend our working example to obtain the Partial MaxSAT and Weighted Partial MaxSAT encodings described in this section. We first describe how we encode  $SoftU$ (left) and $BSU$ (right) constraints:

\begin{align}
\begin{gathered}
    (\neg u_{10}, 1) \notag \\
    (\neg u_9, 1) \tag{Ex. SoftU and BSU}  \\
    (\neg u_8, 1) \notag
\end{gathered}
&&
\begin{gathered}
    u_{10} \rightarrow u_9 \notag \\
    u_9 \rightarrow u_8 \notag
\end{gathered}
\end{align}

Recall that in our example $ub=10$ and $lb=6$. Therefore, we will have $N - (lb + 1) = 10 - (6 + 1) = 3$ $u_i$ indicator variables.

To build the $PMSat_{CX}^{N=10,t=2,S,lb=6}$ instance we add to $Sat_{CX}^{N=10,t=2,S}$ the $CU$ constraint:

\begin{gather}
    c_{\tau_1}^{10} \rightarrow u_{10}, \quad \ldots, \quad c_{\tau_{33}}^{10} \rightarrow u_{10} \notag \\
    c_{\tau_1}^9 \rightarrow u_9, \quad \ldots, \quad c_{\tau_{33}}^9 \rightarrow u_9 \tag{Ex. CU} \\
    c_{\tau_1}^8 \rightarrow u_8, \quad \ldots, \quad c_{\tau_{33}}^8 \rightarrow u_8 \notag
\end{gather}

To build $PMSat_{CCX}^{N=10,t=2,S,lb=6}$ we add to $Sat_{CCX}^{N=10,t=2,S}$ the $CCU$ constraint:

\begin{gather}
    \neg c_{\tau_1}^{9} \rightarrow u_{10}, \quad \ldots, \quad \neg c_{\tau_{33}}^{9} \rightarrow u_{10} \notag \\
    \neg c_{\tau_1}^8 \rightarrow u_9, \quad \ldots, \quad \neg c_{\tau_{33}}^8 \rightarrow u_9 \tag{Ex. CCU} \\
    \neg c_{\tau_1}^7 \rightarrow u_8, \quad \ldots, \quad \neg c_{\tau_{33}}^7 \rightarrow u_8 \notag
\end{gather}

The weighted counterparts, $WPMSat_{CX}^{N=10,t=2,S,lb=6}$ and $WPMSat_{CCX}^{N=10,t=2,S,lb=6}$, need only to replace $SoftU$ by $WSoftU$ (using $w_i = i-(lb+2)+1$), as follows:

\begin{gather}
    (\neg u_{10}, 3) \notag \\
    (\neg u_9, 2) \tag{Ex. WSoftU} \\
    (\neg u_8, 1) \notag
\end{gather}

In order to build the resulting MCAC from the MaxSAT solver truth assignment, we will discard the $x_{i,p,v}$ vars whose corresponding $u_i$ is assigned to False (i.e. test $i$ does not belong to the solution), and proceed as in Example \ref{ex:autodriving-sat}.

\end{xampl}

\section{Solving the $CAN(t,S)$ problem with  MaxSAT}\label{sec:MaxSATbased}

In this section, we show that SAT-based MaxSAT approaches can simulate the \ref{alg:calot} algorithm, while the opposite is not true. This is an interesting insight since the MaxSAT approach additionally provides the option of applying a plethora of MaxSAT algorithms.

Let us first introduce a short description of SAT-based MaxSAT algorithms. For further details, please consult \cite{AnsoteguiBL13,MorgadoHLPM13}. Roughly speaking, SAT-based MaxSAT algorithms proceed by reformulating the MaxSAT optimization problem into a sequence of SAT decision problems. Each SAT instance of the sequence encodes whether there exists an assignment to the MaxSAT instance with a cost less than or equal to a certain $k$. SAT instances with a $k$ less than the optimal cost are unsatisfiable, while the others are satisfiable. The SAT solver is executed in incremental mode to keep the clauses learnt at each iteration over the sequence of SAT instances. Thus, SAT-based MaxSAT can also be viewed as a particular application of incremental SAT solving.

There are two main types of SAT-based MaxSAT solvers: (i)~model-guided and (ii)~core-guided. The first ones iteratively refine (decrease) the upper bound and guide the search with satisfying assignments (models) obtained from satisfiable SAT instances. The second ones iteratively refine (increase) the lower bound and guide the search with the unsatisfiable cores obtained from unsatisfiable SAT instances. Both have strengths and weaknesses, and hybrid approaches exist \cite{ansotegui_exploiting_2016,ansotegui_wpm3_2017}.\\[1mm]

\subsection{The Linear MaxSAT Algorithm}
The Linear algorithm  \cite{een_translating_2006,le_berre_sat4j_2010}, described in Algorithm \ref{alg:linear},  is a  model-guided algorithm for WPMaxSAT. Let $\phi = \phi_s \cup \phi_h$ (line 1) be the input WPMaxSAT instance, where $\phi_s$ ($\phi_h$) is the set of soft (hard) clauses in $\phi$. 

\begin{algorithm}[ht]
    \SetAlgoRefName{Linear}
    \caption{Linear SAT-based algorithm}
    \label{alg:linear}
    \SetKw{KwIfT}{?}
    \SetKw{KwElseT}{:}
    \SetKw{KwInput}{Input:}
    \DontPrintSemicolon

    \KwInput{Weighted Partial MaxSAT formula $\phi \equiv \phi_s \cup \phi_h$, SAT solver $sat$}
    
    $sat.add(\phi_h)$
    
    $sat.add(\{c_i \lor b_i | (c_i, w_i) \in \phi_s\})$
    
    $ub \leftarrow \sum_{(c_i,w_i) \in \phi_s} w_i + 1$
    
    $pb \leftarrow \sum_{(c_i,w_i) \in \phi_s} w_i \cdot b_i \leq ub-1$
    
    $sat.add(pb.to\_cnf)$
    
    \While{True} {
        \lIf{$sat.solve()=\mbox{UNSAT}$} {\Return{$(ub, sat.model)$}}
        
        $ub \leftarrow \sum_{(c_i,w_i) \in \phi_s} w_i \cdot sat.model[b_i]$
        
        $sat.add(pb.update(ub - 1))$
    }
    
\end{algorithm}

At each iteration of the Linear algorithm, the SAT instance solved by the incremental SAT solver is composed of: (i)~the hard clauses  $\phi_h$ (line 2), which guarantee that any possible solution is a \emph{feasible} solution; (ii)~the reification of each soft clause $(c_i,w_i) \in \phi_s$ into clause $(c_i \vee b_i)$, where $b_i$ is a fresh auxiliary variable which acts as a collector of the truth value of the soft clause (line 3); 
and (iii)~the CNF translation of the PB constraint $\sum_{(c_i,w_i) \in \phi_s} w_i \cdot b_i \leq k$, where $k = ub-1$ bounds the aggregated cost of the falsified soft clauses, i.e., the value of the objective function.

Initially, $ub$ is set to $(\sum_{(c_i,w_i) \in \phi_s} w_i + 1)$ (line 4), that is semantically equivalent to $\infty$. Then, iteratively, if the incremental SAT solver returns satisfiable, $ub$ is updated to $(\sum_{(c_i,w_i) \in \phi_s} w \cdot sat.model[b_i])$ (line 9)~\footnote{$sat.model[b_i]$ is 1 if $b_i$ is assigned to True in the model, otherwise it is 0.}; otherwise, $ub$ is the optimal cost (line 8).
If the input instance is unsatisfiable the algorithm returns $\sum_{(c_i,w_i) \in \phi_s} w_i + 1$ (i.e., $\infty$).

A technical point to mention is that the PB constraint is translated into SAT thanks to an incremental PB encoding (line 5) so that whenever we tighten the upper bound, instead of retracting the original PB constraint and encode the new one, we just need to add some additional clauses (line 10). Additionally, if all the weights in the soft clauses are equal, instead of using an incremental PB encoding, we can use an incremental cardinality encoding for which more efficient encodings do exist.

\begin{proposition}\label{prop:linear-wpm-calot}
The Linear algorithm with Weighted Partial MaxSAT instance $WPMSat_{CCX}^{N,t,S,lb}$ as input can simulate the \ref{alg:calot} algorithm  (excluding lines 9 and 10). 
\end{proposition}

The key point establishing the connection of the Linear algorithm with the \ref{alg:calot} algorithm is to show that, given the same upper bound $k$ to both algorithms, the Linear algorithm can propagate the same set of  $c^{i-1}_{\tau}$ variables (line 7 in algorithm \ref{alg:calot}). 

Let us recall that the Linear algorithm, with input $\phi \equiv WPMSat_{CCX}^{N,t,S,lb}$, will generate a sequence of SAT instances composed of the original hard clauses $\phi_h$, the reification of the soft clauses $\bigwedge_{(c_i,w_i) \in \phi_s} (c_i \vee b_i)$, the translation to CNF of the PB constraint $\sum_{(c_i,w_i) \in \phi_s} w_i \cdot b_i \leq k$, where  $(c_i,w_i)$ represents the $i$-th soft clause in $WPMSat_{CCX}^{N,t,S,lb}$, i.e.,  $(\neg u_i, 2^{i-(lb+2)})$ when using the exponential increase, and the current upper bound $k$. 

\begin{proposition}\label{prop:units} If $\phi \equiv WPMSat_{CCX}^{N,t,S,lb}$, then 
\\
$CCU \wedge \bigwedge_{(\neg u_i, 2^{i-(lb+2)}) \in \phi_s} (\neg u_i \vee b_i) \wedge \sum_{(\neg u_i, 2^{i-(lb+2)}) \in \phi_s} 2^{i-(lb+2)} \cdot b_i \leq k  \vdash_{UP} \bigwedge_{k < i \leq N+1} \bigwedge_{\tau \in \Tau_a} c^{i-1}_{\tau}$. %

\end{proposition}

First of all, notice that the weight of a higher index test is strictly greater than the aggregated weights of the lower index tests. Given an upper bound $k$, an \emph{efficient} CNF translation of the PB constraint will allow Unit Propagation (UP) to derive that all $b$s associated with soft clauses with a weight greater than $k$ must be False. Then, from the set of clauses that reify the soft clauses (of the form $\neg u_i \vee b_i$), UP will also derive that the corresponding $u_i$ vars must be False and, from the set of hard clauses $CCU$, UP will derive that the corresponding $c^{i-1}_{\tau}$ must be true. 

If the input problem is a Partial MaxSAT instance, i.e., $PMSat_{CCX}^{N,t,S,lb}$ where the $i$-th soft clause is of the form $(\neg u_i,1)$, the Linear algorithm uses a cardinality constraint instead of a PB constraint to bound the aggregated cost of the falsified soft clauses. In this case, we can only guarantee that  $CCU \wedge \bigwedge_{(\neg u_i,1) \in \phi_s} (\neg u_i \vee b_i) \wedge \sum_{(\neg u_i,1) \in \phi_s}  b_i \leq k  \models \bigwedge_{k < i \leq N+1} \bigwedge_{\tau \in \Tau_a} c^{i-1}_{\tau}$. Notice that, given an upper bound  $k$, UP cannot derive on $\sum_{(\neg u_i,1) \in \phi_s}  b_i \leq k$  the set of $b_i$s that must be False, because all correspond to soft clauses of equal weight.\\[2mm]

{\bf CALOT algorithm cannot simulate the Linear Algorithm:} While the CALOT algorithm decreases the upper bound by one at each iteration, the Linear algorithm can decrease it more aggressively. This is the case when it finds a model with a lower cost than $k-1$ (line 9), which can significantly reduce the number of calls to the SAT solver. \\[1mm]

\subsection{The WPM1 MaxSAT algorithm}

The Fu\&Malik algorithm \cite{FuMalik06} is a core-guided SAT-based MaxSAT algorithm for Partial MaxSAT instances. In contrast to the Linear algorithm, which uses the models to iteratively refine the upper bound, the Fu\&Malik algorithm uses the unsatisfiable cores to refine the lower bound. In particular, the initial SAT instance $\varphi_0$ explored by the Fu\&Malik algorithm is composed of the hard clauses in the input MaxSAT instance $\phi_h$ plus the SAT clauses $c_i$ \emph{extracted} from the  soft clauses $(c_i,w_i)$. We refer to these $c_i$ clauses as soft-indicator clauses.

At each iteration, if $\varphi_k$ is satisfiable, the optimum is $k$. If $\varphi_k$ is unsatisfiable, the clauses in the unsatisfiable core retrieved by the SAT solver are analyzed. If none of the clauses is a soft-indicator clause, the Partial MaxSAT formula is declared unsatisfiable and the algorithm stops. Otherwise, the core tells us that we need to relax the soft-indicator clauses, i.e., we need to violate more clauses. To construct the next instance, $\varphi_{k+1}$, each soft-indicator clause in the core of $\varphi_{k}$ is relaxed with a fresh auxiliary variable $b$ and a hard EO cardinality constraint is added on these new variables, indicating that at least one clause must be violated (this is what the core told us) and at most one clause is violated (this prevents jumping over the optimum). \\[1mm]

{\bf The WPM1 algorithm} \cite{ansotegui_solving_2009,manquinho_algorithms_2009} is an extension of the Fu\&Malik algorithm that solves Weighted Partial MaxSAT instances by applying the split rule for weighted clauses. In particular, we are interested in using the \emph{Stratified} WPM1 algorithm (\ref{alg:wpm1}) \cite{ansotegui_improving_2012}, which clusters the input clauses according to their weights\footnote{Recall that hard clauses have weight $\infty$.}. These clusters were originally named as strata in \cite{ansotegui_improving_2012}. The algorithm incrementally merges the clusters solving the related subproblem until all clusters have been merged. In its simpler version, all the clauses in a cluster have the same weight (called the representative weight), and clusters are added in decreasing order respect to the representative weight, but other strategies can also be applied \cite{ansotegui_improving_2012}.

\newcommand{\graybox}[1]{%
  \begingroup
  \setlength{\fboxsep}{0pt}%
  \colorbox{gray!20}{\raisebox{0pt}[8pt][3pt]{#1}}%
  \endgroup
}

\begin{algorithm}[ht]
    \SetAlgoRefName{WPM1}
    \caption{Stratified WPM1}
    \label{alg:wpm1}
    \SetKw{KwIfT}{?}
    \SetKw{KwElseT}{:}
    \SetKw{KwInput}{Input:}
    \SetKw{KwAnd}{and}
    \SetKwProg{KwFunc}{function}{}{}
    \DontPrintSemicolon

    \KwInput{Weighted Partial MaxSAT formula $\phi$, SAT solver $sat$}

    $\phi_{wk}, \phi_{re}, status \leftarrow \emptyset, \phi, SAT$
    
    \While{True} {
        \If{status = SAT}{
            $sat.add(\phi_{st} \leftarrow$ next\_stratum$(\phi_{wk}, \phi_{re}))$
        
            $\phi_{wk}, \phi_{re} \leftarrow \phi_{wk} \cup \phi_{st}, \phi_{re} \setminus \phi_{st}$
        }
        
        \If{$(status \leftarrow sat.solve())=\mbox{SAT}$}{
            \lIf{$\phi_{re} = \emptyset$}{\Return{$(cost(sat.model, \phi), sat.model)$}}
        }
        
        \Else{
            
            \lIf{$(to\_relax \leftarrow \mbox{core\_analysis}(\phi_{wk}, sat.core))=\emptyset$} {\Return{$(\infty, \emptyset)$}}
        
            $relaxed, B\mbox{\graybox{, $residuals$}}  \leftarrow$ \graybox{split\_and\_}relax$(to\_relax, sat.n\_vars)$
            
            $sat.retract(to\_relax)$

            $sat.add(\phi_{rx} \leftarrow relaxed \cup (CNF(\sum_{b \in B} b = 1), \infty))$

            $\phi_{wk}, \phi_{re} \leftarrow (\phi_{wk} \setminus to\_relax) \cup \phi_{rx}, \phi_{re} \mbox{\graybox{$\cup \  residuals$}}$

        }
    }

\end{algorithm}

In the \ref{alg:wpm1} algorithm, variable $\phi_{wk}$ represents the formula that is MaxSAT equivalent to the merged clusters (strata) so far, while $\phi_{re}$ represents the remaining weighted clauses from the original input instance $\phi$. Whenever we solve to optimality the current instance $\phi_{wk}$, i.e., the SAT solver returned a SAT answer in the last call (line 4) but $\phi_{re} \neq \emptyset$, function $next\_stratum$ updates variable $\phi_{st}$ to the new stratum (cluster) to be merged with $\phi_{wk}$ \footnote{In \cite{ansotegui_improving_2012}, the first call to $next\_stratum$ returns the cluster of all hard clauses since their representative weight is $\infty$} (the working SAT instance (line 5) and variables $\phi_{wk}, \phi_{re}$ are updated accordingly (line 6)).  Otherwise, the SAT solver returned UNSAT in the previous call, meaning that we are still optimizing the current subproblem  $\phi_{wk}$ and need to call the SAT solver again (line 7). 

If the SAT solver returns a SAT answer and all the original clauses in $\phi$ have been considered, i.e. $\phi_{re} = \emptyset$, then we have optimized the input instance $\phi$ and return its cost and an optimal model (line 8).

If the SAT solver returns an UNSAT answer, first we analyze the unsatisfiable core returned by the SAT solver (line 10) and return the soft-indicator clauses to be relaxed in variable $to\_relax$, if any; otherwise, we have certified that the set of hard clauses is unsatisfiable, i.e., we return cost $\infty$ and an empty model.

Function $split\_and\_relax$ (line 11) first applies the split rule to the soft-indicator clauses in $to\_relax$ and generates two sets, one where all the clauses are normalized to have the minimum weight, and another with the residuals of each clause respect to the minimum weight in $to\_relax$. Second, the set of clauses with the minimum weight are extended, each with an additional fresh variable and stored in the set $relaxed$ as in the Fu\&Malik algorithm. The new fresh variables are returned in set $B$.

Finally, the original set of clauses $to\_relax$ is retracted from the SAT solver (line 12), and the new set $relaxed$ is added to the working SAT instance plus the cardinality constraint that increases the lower bound as in the Fu\&Malik algorithm (line 13)\footnote{Notice that $(CNF(\sum_{b \in b\_vars} b = 1), \infty)$ is a set of clauses that have $\infty$ weight.}. In line 14, $\phi_{wk}$ is updated to reflect the changes in the SAT working formula, and the remaining formula $\phi_{re}$ is extended with the residuals generated from the application of the split rule.

As a final remark, notice that if the statements in grey boxes of the \ref{alg:wpm1} algorithm are erased and function $next\_stratum$ is instructed to report sequentially, first the hard clauses and then the soft clauses, we get the original Fu\&Malik algorithm.

In the context of the Covering Array Number problem, the Fu\&Malik algorithm on the $PMSat_{CCX}^{N,t,S,lb}$ instance will perform a bottom-up search, i.e, the first query will correspond to the question of whether the covering array can be constructed with $k=0$ tests, then with $k=1$ tests, etc. This approach does not provide any intermediate upper bounds since the only query answered positively corresponds to the optimum. 

However, interestingly, by considering the weighted version of the Fu\&Malik algorithm, we can perform a top-down search on the Covering Array problem and provide intermediate upper bounds.

\begin{proposition}\label{prop:wpm1-wpm-calot}
The Stratified WPM1 algorithm with input $WPMSat_{CCX}^{N,t,S,lb}$ can simulate the \ref{alg:calot} algorithm (excluding lines 9 and 10). 
\end{proposition}

Back to the context of covering arrays, each cluster in $WPMSat_{CCX}^{N,t,S,lb}$ would be composed of a single soft clause $(\neg u_i, w_i)$, except the cluster containing all the hard clauses. The first subproblem seen by the Stratified WPM1 algorithm encodes the query of whether one can build a covering array using $N$ tests. The next subproblem incorporates the first soft clause $(\neg u_N, w_N)$ and encodes the query of whether one can construct the covering array using $N-1$ tests. Notice that each $\neg u_i$ will propagate, according to $CCU$, the corresponding $c^{i-1}_{\tau}$ vars as in the \ref{alg:calot} algorithm. Notice also that every solution of a subproblem is an upper bound for the covering array.%

The discussion of this section has provided insights into how to solve Covering Arrays through MaxSAT, but also into how to fix similar difficulties in other problems where MaxSAT is not yet effective enough.%

\subsection{Test-based Streamliners for the $CAN(t,S)$ problem} \label{sec:test-based-streamliners}

Notice that a solution for a $CAN(t,S)$ problem can be extended to multiple solutions in the previous MaxSAT translations. 
This happens when $CAN(t,S) < N$, since the assignment to the $x$ vars related to any test $i$ with $i>CAN(t,S)$ (useless from the point of view of the $CAN(t,S)$ problem) still needs to be consistent with the $X$ and $SUTX$ constraints. In general, notice that $SUTX$ can be NP-complete.

Lines 9 and 10 of the CALOT algorithm, as described in section \ref{sec:calot}, fix that problem but cannot 
directly be applied within MaxSAT algorithms since the solver is not aware of the $CAN(t,S)$ problem semantics. 

However, we can reproduce a similar effect. 
At the preprocessing step, we can build a \emph{dummy} test case $\upsilon$ by computing a solution to $S_\varphi$ (e.g. with a SAT solver) or select any of the test cases in the solution returned by the ACTS tool when computing the upper bound (see section \ref{sec:preproc}). Then, we can state in the MaxSAT encoding that if a given test $i$ is not part of the optimal solution (i.e., $u_i$ is False), then the corresponding $x$ vars are set to the value in the test case $\upsilon$. %

\begin{equation}\tag{$NUX$}\label{eq:NUX}
\bigwedge_{i \in [lb+2 \ldots N]} \left( \neg u_i  \rightarrow \bigwedge_{(p,v) \in \upsilon} x_{i,p,v} \right)     
\end{equation}

The \emph{dummy} test case $\upsilon$ exactly plays the role of the so-called streamliner constraints \cite{gomes_streamlined_2004}, which rule out some of the possible solutions but make the search of the remaining solutions more efficient.

There is yet another way to mitigate that potential bottleneck. We can indeed extend $SUTX$ clauses for test $i$ with literal $\neg u_i$. Therefore, whenever test $i$ is no longer in the optimal solution (i.e. $u_i$ is False), the corresponding $SUT$ constraints are trivially satisfied. However, in the experimental investigation, we confirmed that this option is less efficient than adding $NUX$ clauses.

\begin{xampl}
\label{ex:autonomous-nux}

For the SUT in Example \ref{ex:autodriving-sat}, let us assume that we use the following \emph{dummy} test:

   \begin{center}
    \begin{tabular}{@{\hskip 2mm}c@{\hskip 2mm}|@{\hskip 2mm}c@{\hskip 2mm}|@{\hskip 2mm}c@{\hskip 2mm}|@{\hskip 2mm}c@{\hskip 2mm}}
        \toprule
        L & E & S & M \\
        \midrule
        dy & ur & ra & cb \\
        \bottomrule
    \end{tabular} 
    \end{center}

Then, the $NUX$ encoding is:

\begin{gather}
    \neg u_{10} \rightarrow (x_{10,L,dy} \wedge x_{10,E,ur} \wedge x_{10,S,ra} \wedge x_{10,M,cb}) \notag \\
    \neg u_{9} \rightarrow (x_{9,L,dy} \wedge x_{9,E,ur} \wedge x_{9,S,ra} \wedge x_{9,M,cb}) \tag{Ex. NUX} \\
    \neg u_{8} \rightarrow (x_{8,L,dy} \wedge x_{8,E,ur} \wedge x_{8,S,ra} \wedge x_{8,M,cb})\notag
\end{gather}

\end{xampl}

\section{The $T(N;t,S)$ problem as Weighted Partial MaxSAT}\label{sec:shortening}

For some applications, we may not be able to use as many test cases as the covering array number (e.g. due to budget restrictions), but we may still be interested in solving the Tuple Number problem, i.e., to determine the maximum number of covered $t$-tuples we can get with a test suite of fixed size. This problem is also known as the Optimal Shortening Covering Arrays (OSCAR) problem. These \emph{shortened} covering arrays (to which we refer more precisely just as \emph{test suites} since they do not cover all $t$-tuples) have been used to improve the initialization of metaheuristic approaches for Covering Arrays (without SUT constraints) \cite{Carrizales-TurrubiatesRT11}. These metaheuristics obtain suboptimal Covering Arrays very quickly. 
Once again, MaxSAT technology can play an important role when SUT constraints are considered. Moreover, the size of the SAT/MaxSAT encodings for this problem are smaller than encodings for computing the Covering Array Number, since fewer tests are taken into consideration.

In the following, we show how we can modify the $Sat_{CX}^{N,t,S}$ and $Sat_{CCX}^{N,t,S}$ formulae to become Partial MaxSAT encodings of the Tuple Number problem.

The basic idea is that we need to soften the hard restriction that enforces all allowed $t$-tuples to be covered. To this end, we modify the SAT instance $Sat_{CX}^{N,t,S}$ as follows: First, we  soften all the clauses from equation \ref{EqCovers} which encode that every $t$-tuple $\tau$ must be covered by at least one test case, therefore allowing to violate (or relax) these constraints. For the sake of clarity, although not required for soundness, we introduce a new set of indicator variables $c_{\tau}$ that reify each ALO constraint in equation \ref{EqCovers} by introducing the following hard constraints:

\begin{equation}\label{EqReifyCovers}\tag{$RC$}
\bigwedge_{\tau \in \Tau_a} (c_{\tau} \leftrightarrow \bigvee_{i \in [N]} c^{i}_{\tau})
\end{equation}

Then, we add the following set of soft clauses: 
\begin{equation}\label{EqSoftCovers}\tag{$SoftC$}
\bigwedge_{\tau \in \Tau_a}(c_{\tau}, 1). 
\end{equation}

Finally, we we replace in $Sat_{CX}^{N,t,S}$ the set of constraints $C$ (the hard constraint that forced to cover  all the tuples) by the previous two sets of constraints.

\begin{proposition}
Let $S$ be a SUT model and let $TPMSat_{CX}^{N,t,S}$ be $Sat_{CX}^{N,t,S}\left\{\frac{SoftC \wedge RC}{C}\right\}$. The optimal cost of $TPMSat_{CX}^{N,t,S}$ is $|\Tau_a| - T(N;t,S)$.

\end{proposition}
\begin{rmarkN}
Even if $N>lb$, we cannot use fixed-tuple symmetry breaking since we do not know whether the $t$-tuples that we fix will lead to an optimal solution. Therefore, fixed-tuple symmetry is disabled for all the encodings in this section.
\end{rmarkN}

\begin{rmarkN}\label{remark:Tau_T}
When computing the tuple number, we can avoid the step of detecting all forbidden tuples since the encoding remains sound, i.e., we can interchange $\Tau_a$ by $\Tau$. Notice that those $c_{\tau}$ vars related to forbidden tuples will always be set to False. Moreover, notice that a core-guided algorithm may potentially detect easily as many unsatisfiable cores as forbidden tuples which include just the unit soft clause that represents the forbidden tuple.
\end{rmarkN}

In case we want to extend $Sat_{CCX}^{N,t,S}$ to compute the tuple number, we just need to notice that the previous defined role of $c_{\tau}$ corresponds exactly to variable $c^{N}_{\tau}$ in $Sat_{CCX}^{N,t,S}$, so we just need to soften the hard unit clauses $c^{N}_{\tau}$ (described in $CCX$) with weight 1.
\begin{proposition}
Let $S$ be a SUT model and let $TPMSat_{CCX}^{N,t,S}$ be $Sat_{CCX}^{N,t,S}\left\{\frac{(c^{N}_{\tau},1)}{(c^{N}_{\tau}) }\right\}$.  The optimal cost of $TPMSat_{CCX}^{N,t,S}$ is $|\Tau_a| - T(N;t,S)$.
\end{proposition}

\begin{xampl}
\label{ex:autonomous-tn}

We show how to build $TPMSat_{CX}^{N=10,t=2,S}$ for the SUT in Example \ref{ex:autodriving-sat}.

We must create a new variable $c_{\tau}$ for each value tuple in $\Tau_a$ and then replace constraint $C$ in $SAT_{CX}^{N=10,t=2,S}$ (see Example \ref{ex:autodriving-sat}) by $RC$ (left). Finally, we have to add the $SoftC$ soft clauses (right):

\begin{align}
\begin{gathered}
    c_{\tau_1} \leftrightarrow (c_{\tau_1}^1 \lor c_{\tau_1}^2 \lor \cdots \lor c_{\tau_1}^{10}) \notag \\
    \vdots \tag{Ex. RC and SoftC} \\
    c_{\tau_{33}} \leftrightarrow (c_{\tau_{33}}^1 \lor c_{\tau_{33}}^2 \lor \cdots \lor c_{\tau_{33}}^{10}) \notag
\end{gathered}
&&
\begin{gathered}
    (c_{\tau_1}, 1) \notag \\
    \vdots \notag \\
    (c_{\tau_{33}}, 1) \notag
\end{gathered}
\end{align}

For the $TPMSat_{CXX}^{N=10,t=2,S}$, we just have to soften, with weight 1, the set of clauses from $CCX$ $(b)$ in $SAT_{CCX}^{N=10,t=2,S}$ (see Example \ref{ex:autodriving-sat}).

\end{xampl}

In what follows, we present two extensions.

\subsection{Combining the $CAN(t,S)$ and $T(N;t,S)$  problems}\label{sec:comb_CA_TN}

The Covering Array and Tuple Number problems can lead to think about a more general formulation of the optimization problem where we want to maximize the number of covered $t$-tuples while minimizing the number of test cases. Notice that it will depend on the value of $N$ respect to the covering array number (not necessarily known a priori) whether we are, in essence, solving the covering array number or the tuple number problem.

To this end, we take the $PMSat_{CX}^{N,t,S,lb}$ encoding of the Covering Array Number problem for a SUT model $S$, $N$ tests and strength $t$. As earlier shown in this section, we first replace the set of hard constraints $C$ by $RC$ and $SoftCWU$.

\begin{equation}\label{EqSoftCoversWU}\tag{$SoftCWU$}
\bigwedge_{\tau \in \Tau_a}(c_{\tau}, |u_i|+1). 
\end{equation}

Notice that we prefer violating all soft  clauses $(\neg u_i,1)$ over violating a single soft clause $(c_{\tau}, |u_i|+1)$. This way, we guarantee that any solution to our new Weighted Partial MaxSAT instance maximises the number of covered $t$-tuples while minimises the number of needed test cases.

\begin{proposition}
If $N \geq CAN(t,S)$, the optimal cost of the Weighted Partial MaxSAT instance $PMSat_{CX}^{N,t,S,lb}\left\{\frac{SoftCWU \wedge RC }{C}\right\}$
is $CAN(t,S) - (lb + 1) + (|\Tau_a| - T(N;t,S)) \cdot (|u_i|+1)$, otherwise it is $N - (lb+1) + (|\Tau_a| - T(N;t,S)) \cdot (|u_i|+1)$. \footlabel{footnote_Ta_TN}{Notice that if $N \geq CAN(t,S)$, then $|\Tau_a| - T(N;t,S))$ is 0. However, we keep this expression in case we want to interchange $\mathcal{T}_a$ by $\mathcal{T}$
 (see Remark \ref{remark:Tau_T}).}
\end{proposition}

The same idea can be applied to $PMSat_{CCX}^{N,t,S,lb}$ by softening the unit hard clauses $(c^{N}_{\tau})$ in equation (b) from $CCX$ with weight $|u_i|+1$. Here, it is important to recall the discussion in section \ref{sec:mcacSAT} on the need of equation (c) in $CCX$. The other, perhaps more natural, alternative was to replace equation (c) in $CCX$ by $\bigwedge_{\tau \in \Tau_a}  \neg c^{0}_{\tau} $. The problem arises when, in an optimal solution, $\tau$ is not covered, what also implies that $(c^{N}_{\tau})$ is False.
Notice that we need to satisfy all clauses related to $\tau$ in $CCX$ but, in order to do that, we need to set all $c^{i}_{\tau}$ vars to False. This may not be compatible with equation $CCU$ (clauses of the form $\neg c^{i-1}_{\tau} \rightarrow u_i $) when some test $i$ is discarded to be in the solution and variable $u_i$ is set to False, since UP will derive in $CCU$ that $c^{i-1}_{\tau}$ is True. In this case, a contradiction is reached. On the other hand, as discussed in section \ref{sec:mcacSAT}, equation (c) allows to set all $c^{i}_{\tau}$ vars to True when $(c^{N}_{\tau})$ is False and trivially satisfy all clauses in $CCX$ related to $\tau$.

\begin{proposition}
If $N \geq CAN(t,S)$, the optimal cost of the Weighted Partial MaxSAT instance $PMSat_{CCX}^{N,t,S,lb}\left\{\frac{SoftCWU }{(c^{N}_{\tau})}\right\}$
is $CAN(t,S) - (lb + 1) + (|\Tau_a| - T(N;t,S)) \cdot (|u_i|+1)$, otherwise it is $N - (lb+1) + (|\Tau_a| - T(N;t,S)) \cdot (|u_i|+1)$. \footref{footnote_Ta_TN}
\end{proposition}

\subsection{The $CAN(t,S)$ problem with Relaxed Tuple Ratio Coverage as MaxSAT}\label{sec:pctg}

We can tackle other realistic settings where we still want to use the minimum number of tests, but there is no need to achieve a 100\% ratio of covered $t$-tuples (mandatory per definition in Covering Arrays). Notice that the last tests that shape the covering array number tend to cover very few not yet covered $t$-tuples. Therefore, if these tests are expensive enough in our setting, we may consider relaxing the ratio coverage and skip these tests.

The mentioned problem can be encoded by replacing the previously soft constraints on the $c_{\tau}$ vars with a hard cardinality constraint on the minimum number of $t$-tuples to be covered as follows:

\begin{equation}\label{EqCCard}\tag{$CCard$}
\sum_{\tau \in \Tau_a} c_{\tau} \geq \lceil|\Tau_a| \cdot rt\rceil
\end{equation}

\noindent where $rt$ is the ratio of allowed $t$-tuples that we want to cover. Notice that, for efficiency reasons, $CCard$ can be also described as $\sum_{\tau \in \Tau_a} \neg c_{\tau} \leq \lceil|\Tau_a| \cdot (1-rt)\rceil$.

\begin{rmarkN}
With this formulation, we cannot use the fixed-tuples symmetry breaking since we do not know whether we will require at least $lb$ tests to cover the specified ratio of allowed $t$-tuples.
\end{rmarkN}

\begin{proposition}
Let $RTPMSat_{CCX}^{N,t,S,rt}$ be $PMSat_{CCX}^{N,t,S,lb=0}\left\{\frac{CCard}{(c^{N}_{\tau})}\right\}$. The optimal cost of $RTPMSat_{CCX}^{N,t,S,rt}$ is the minimum $N'$ such that $T(N',t,S) \geq \lceil |\Tau_a| \cdot rt \rceil$.

\end{proposition}

\section{Incomplete MaxSAT Algorithms for the $T(N;t,S)$ problem}\label{sec:incompleteTN}

As argued earlier, if certifying optimality is not a requirement and we are just interested in obtaining a good suboptimal solution in a reasonable amount of time, we can apply incomplete MaxSAT algorithms on the encodings of the Tuple Number problem described in the previous section.
Additionally, in this section, we present a new incomplete algorithm to compute suboptimal solutions for the Tuple Number problem.

\subsection{MaxSAT based Incremental Test Suite Construction}\label{sec:hybrid}

A way to reduce the search space of any constraint problem is to add the so-called streamliner constraints \cite{gomes_streamlined_2004}. We recall that these constraints rule out some of the possible solutions but make the search of the remaining solutions more efficient. However, in practice, streamliners can rule out all the solutions. 

In our context, the streamliners constraints correspond to a set of tests that we think have the potential to be part of optimal solutions. By fixing these tests, we generate a new covering array problem, easier to solve, but whose Covering Array Number can be greater than or equal to that of the original covering array, because we may have missed all the optimal solutions. We iterate this process until all $t$-tuples get covered. To select the $k$ candidate test to be fixed at each iteration, we solve the Tuple Number problem restricted to length $k$.

In the context of the Tuple Number problem, this iterative process of fixing tests should not only finish when all $t$-tuples have been covered but also when the requested $N$ tests have been fixed.

To that end, here we combine a greedy iterative approach with the SAT-based MaxSAT approaches from section~\ref{sec:shortening} in the \ref{alg:hybridgreedy} algorithm.

\newcommand{\satcomm}[1]{\textsf{\scriptsize #1}}
\begin{algorithm}[h!t]
    \SetAlgoRefName{IncrementalCA}
    \caption{MaxSAT based Incremental Test Suite Construction \label{alg:hybridgreedy}}
    \SetKw{KwOr}{or}
    \SetKw{KwAnd}{and}
    \SetKw{KwIfT}{?}
    \SetKw{KwElseT}{:}
    \SetKw{KwGoTo}{go to}
    \SetKwProg{KwFunc}{function}{}{}
    \SetKw{KwInput}{Input:}
    \SetKwComment{KwComm}{\#}{}
    \DontPrintSemicolon
    
    \KwInput{SUT model $S$, Tests $N_i$ per iteration, SAT-based MaxSAT solver $msat$}

    $\Tau_r, \Upsilon \leftarrow \Tau_a, \emptyset$\;
    
    \While {$\Tau_r \neq \emptyset$ \KwAnd $|\Upsilon |<N$}{
        $N^\prime \leftarrow min(N_i, N-|\Upsilon |)$\;
    
        $msat.add\left ( TPMSat_{CCX}^{N^\prime,t,S} \right )$\;
    
        $msat.solve()$\; %
    
        $\upsilon \leftarrow \mbox{tests from } msat.model$\; 
        
        $\Upsilon \leftarrow \Upsilon \cup \upsilon$\;
        
        $\Tau_r \leftarrow \Tau_r \setminus \{\tau\ |\ \upsilon \models \tau\}$\;
    }
    \Return $\Upsilon $\;
\end{algorithm}
\let\satcomm\undefined

In this algorithm, we begin with the remaining tuples to cover $\Tau_r$, initially assigned to allowed tuples $\Tau_a$, as well as an empty test suite $\Upsilon$ (line 2). %
Then, we first check how many tests should be encoded; the minimum between the tests in iteration $N_i$ and the remaining number of tests left to complete the test suite, $N-|\Upsilon|$ (line 4), storing the result into $N^\prime$.
Next, we solve the Tuple Number problem for these $N^\prime$ tests, encoded as a $TPMSat_{CCX}^{N^\prime,t,S}$ formula (lines 5, 6) from section~\ref{sec:shortening}.
We extract the model from the MaxSAT solver, interpreting it into newly found test cases $\upsilon$ (line 7). Then, those new tests are added to test suite $\Upsilon$ (line 8).
Finally, the tuples covered by these new test cases are removed from $\Tau_r$ (line 9).
This iteration is repeated until no more tuples are left in $\Tau_r$ or we have reached the requested $N$ test cases (line 3), in which case we return the constructed test suite $\Upsilon$ (line 10).

\section{Experimental Evaluation}\label{sec:experimental}

In this section, we report on an extensive experimental investigation conducted to assess the  approaches proposed in the preceding sections. We start by defining the benchmarks, which include 28 industrial, real-world or real-life instances and 30 crafted instances, and the algorithms involved in the evaluation.

We contacted the authors of \cite{YamadaKACOB15} and \cite{Yamada16} 
to obtain the benchmarks used in their experiments. In particular, the available benchmarks are: (i) Cohen et al. \cite{cohen2008constructing}, with 5 real-world and 30 artificially generated (crafted) covering array problems; (ii) Segall et al. \cite{segall11}, with 20 industrial instances; (iii) Yu et al. \cite{Yu15}, with two real-life systems reported by ACTS users; and (iv) Yamada et al. \cite{Yamada16}, with an industrial instance named ``Company\_B''.

Table \ref{tab:inst-sizes} provides information about the System Under Test of each instance, where
$S_P$ is the number of parameters and their domain (e.g. the meaning of $2^{29}3^{1}$ in instance 7 is that the instance contains 29 parameters of domain 2 and 1 parameter of domain 3);
$S_{\varphi}$ is the number of SUT constraints and their sizes (e.g. the meaning of $2^{13}3^{2}$ in instance 7 is that the instance contains 13 constraints of size 2 and 2 constraints of size 3); and 
\emph{\# lits $CNF(S_{\varphi})$} is the number of literals of the CNF representation of $S_{\varphi}$ (i.e. the sum of the sizes of all clauses).

Table \ref{tab:inst-sizes}  also reports, for $t=2$, the following data:
$ub^{ACTS}$, which indicates the upper bound returned by the ACTS tool (see section \ref{sec:preproc}); $ub^{\simeq}$,  which is the best known upper bound (a star indicates that it is optimal, i.e., $CAN(2,S)$);
$lb$, which reports the lower bound (computed as in section \ref{sec:preproc}); and
$|\Tau_a|$ and $|\Tau_f|$, which report the number of allowed and forbidden tuples, respectively.

Finally, we also show, for the $PMSat_{CCX}^{N,t=2,S,lb}$ encoding of each instance, the following information:
\emph{\# vars}, which is the number of variables used by this encoding;
\emph{\# clauses}, which is the number of clauses;
\emph{\# lits}, which is the number of literals; and
\emph{size (MB)}, which is the file size of the WCNF formula in MB.

Notice that in this paper we focus on $t=2$ strength coverage.

\begin{table}[h!t]

\centering

\resizebox{\textwidth}{!}{

\begin{tabular}{l|llr|rcrrr|rrrr}
\toprule
\multicolumn{1}{c|}{} & \multicolumn{3}{c|}{System Under Test (SUT)} & \multicolumn{5}{c|}{Bounds for $t = 2$} & \multicolumn{4}{c}{$PMSat_{CCX}^{N,t=2,S,lb}$} \\

Instance & $S_P$ & $S_{\varphi}$ & \makecell[bc]{\# lits \\ $CNF(S_{\varphi})$} & $ub^{ACTS}$ & $ub^{\simeq}$ & lb & $|\Tau_a|$ & $|\Tau_f|$ & \# vars & \# clauses & \# lits & size (MB) \\
\midrule

Cohen et al. \cite{cohen2008constructing} &  &  &  &  &  &  &  &  &  &  &  & \\

\midrule
1 & $2^{86}3^{3}4^{1}5^{5}6^{2}$ & $2^{20}3^{3}4^{1}$ & 53 & 48 & 37 & 35 & 23876 & 474 & 1158588 & 2620675 & 7463282 & 60.01 \\
2 & $2^{86}3^{3}4^{3}5^{1}6^{1}$ & $2^{19}3^{3}$ & 47 & 32 & 30* & 29 & 20331 & 237 & 657890 & 1371738 & 3984183 & 29.91 \\
3 & $2^{27}4^{2}$ & $2^{9}3^{1}$ & 21 & 19 & 18* & 15 & 1838 & 14 & 36217 & 79008 & 222390 & 1.47 \\
4 & $2^{51}3^{4}4^{2}5^{1}$ & $2^{15}3^{2}$ & 36 & 22 & 20* & 19 & 7530 & 386 & 168852 & 358291 & 1025536 & 7.33 \\
5 & $2^{155}3^{7}4^{3}5^{5}6^{4}$ & $2^{32}3^{6}4^{1}$ & 86 & 54 & 45 & 35 & 76259 & 73 & 4142574 & 9720622 & 27451121 & 236.74 \\
6 & $2^{73}4^{3}6^{1}$ & $2^{26}3^{4}$ & 64 & 25 & 24* & 23 & 11382 & 1878 & 289001 & 597814 & 1730859 & 12.72 \\
7 & $2^{29}3^{1}$ & $2^{13}3^{2}$ & 32 & 12 & 9 & 5 & 1567 & 231 & 19566 & 49758 & 132435 & 0.85 \\
8 & $2^{109}3^{2}4^{2}5^{3}6^{3}$ & $2^{32}3^{4}4^{1}$ & 80 & 47 & 36* & 35 & 33680 & 1098 & 1597165 & 3590459 & 10247230 & 84.50 \\
9 & $2^{57}3^{1}4^{1}5^{1}6^{1}$ & $2^{30}3^{7}$ & 81 & 22 & 20* & 19 & 6835 & 1720 & 153584 & 325984 & 932515 & 6.63 \\
10 & $2^{130}3^{6}4^{5}5^{2}6^{4}$ & $2^{40}3^{7}$ & 101 & 47 & 41 & 35 & 52659 & 2029 & 2493173 & 5608369 & 16010703 & 135.34 \\
11 & $2^{84}3^{4}4^{2}5^{2}6^{4}$ & $2^{28}3^{4}$ & 68 & 47 & 39 & 35 & 23636 & 707 & 1123311 & 2523897 & 7200149 & 57.70 \\
12 & $2^{136}3^{4}4^{3}5^{1}6^{3}$ & $2^{23}3^{4}$ & 58 & 43 & 36* & 35 & 49522 & 978 & 2144718 & 4675267 & 13461992 & 108.23 \\
13 & $2^{124}3^{4}4^{1}5^{2}6^{2}$ & $2^{22}3^{4}$ & 56 & 40 & 36* & 35 & 38862 & 1701 & 1567084 & 3319517 & 9632256 & 75.77 \\
14 & $2^{81}3^{5}4^{3}6^{3}$ & $2^{13}3^{2}$ & 32 & 39 & 36* & 35 & 20544 & 618 & 810618 & 1697204 & 4936072 & 37.18 \\
15 & $2^{50}3^{4}4^{1}5^{2}6^{1}$ & $2^{20}3^{2}$ & 46 & 32 & 30* & 29 & 8388 & 155 & 273410 & 569181 & 1650514 & 12.10 \\
16 & $2^{81}3^{3}4^{2}6^{1}$ & $2^{30}3^{4}$ & 72 & 25 & 24* & 23 & 14600 & 2303 & 370051 & 765960 & 2218422 & 16.44 \\
17 & $2^{128}3^{3}4^{2}5^{1}6^{3}$ & $2^{25}3^{4}$ & 62 & 41 & 36* & 35 & 43390 & 66 & 1792402 & 3835891 & 11100545 & 88.14 \\
18 & $2^{127}3^{2}4^{4}5^{6}6^{2}$ & $2^{23}3^{4}4^{1}$ & 62 & 52 & 41 & 35 & 50128 & 28 & 2625808 & 6092947 & 17250882 & 146.38 \\
19 & $2^{172}3^{9}4^{9}5^{3}6^{4}$ & $2^{38}3^{5}$ & 91 & 51 & 43 & 35 & 98778 & 114 & 5064366 & 11694341 & 33170488 & 287.31 \\
20 & $2^{138}3^{4}4^{5}5^{4}6^{7}$ & $2^{42}3^{6}$ & 102 & 60 & 54 & 35 & 64620 & 3320 & 3903864 & 9411047 & 26386102 & 227.62 \\
21 & $2^{76}3^{3}4^{2}5^{1}6^{3}$ & $2^{40}3^{6}$ & 98 & 39 & 36* & 35 & 15442 & 2742 & 610938 & 1279170 & 3717471 & 27.90 \\
22 & $2^{72}3^{4}4^{1}6^{2}$ & $2^{20}3^{2}$ & 46 & 37 & 36* & 35 & 13405 & 1181 & 503127 & 1028139 & 3008516 & 22.48 \\
23 & $2^{25}3^{1}6^{1}$ & $2^{13}3^{2}$ & 32 & 14 & 12* & 11 & 1495 & 173 & 21856 & 47740 & 132915 & 0.85 \\
24 & $2^{110}3^{2}5^{3}6^{4}$ & $2^{25}3^{4}$ & 62 & 48 & 41 & 35 & 34204 & 570 & 1656252 & 3748659 & 10679658 & 88.30 \\
25 & $2^{118}3^{6}4^{2}5^{2}6^{6}$ & $2^{23}3^{3}4^{1}$ & 59 & 52 & 49 & 35 & 46968 & 52 & 2461280 & 5710735 & 16167454 & 136.81 \\
26 & $2^{87}3^{1}4^{3}5^{4}$ & $2^{28}3^{4}$ & 68 & 34 & 26 & 24 & 20921 & 667 & 719347 & 1643461 & 4647485 & 36.52 \\
27 & $2^{55}3^{2}4^{2}5^{1}6^{2}$ & $2^{17}3^{3}$ & 43 & 37 & 36* & 35 & 9714 & 43 & 365524 & 746797 & 2183919 & 16.18 \\
28 & $2^{167}3^{16}4^{2}5^{3}6^{6}$ & $2^{31}3^{6}$ & 80 & 57 & 50 & 35 & 96599 & 74 & 5535861 & 13181074 & 37087871 & 322.33 \\
29 & $2^{134}3^{7}5^{3}$ & $2^{19}3^{3}$ & 47 & 29 & 25* & 24 & 45839 & 32 & 1338905 & 2899941 & 8321499 & 64.34 \\
30 & $2^{73}3^{3}4^{3}$ & $2^{31}3^{4}$ & 74 & 22 & 16* & 15 & 12453 & 1308 & 277976 & 640938 & 1792681 & 13.16 \\
apache & $2^{158}3^{8}4^{4}5^{1}6^{1}$ & $2^{3}3^{1}4^{2}5^{1}$ & 22 & 33 & 30* & 29 & 66927 & 3 & 2221926 & 4701044 & 13619419 & 109.16 \\
bugzilla & $2^{49}3^{1}4^{2}$ & $2^{4}3^{1}$ & 11 & 19 & 16* & 15 & 5818 & 4 & 112768 & 247130 & 697953 & 4.82 \\
gcc & $2^{189}3^{10}$ & $2^{37}3^{3}$ & 83 & 23 & 15 & 8 & 82770 & 39 & 1913568 & 5063264 & 13685896 & 112.78 \\
spins & $2^{13}4^{5}$ & $2^{13}$ & 26 & 26 & 19* & 15 & 979 & 13 & 27050 & 64498 & 177169 & 1.23 \\
spinv & $2^{42}3^{2}4^{11}$ & $2^{47}3^{2}$ & 100 & 45 & 33 & 15 & 8741 & 56 & 401069 & 1063265 & 2888090 & 22.53 \\
\midrule

Segall et al. \cite{segall11} &  &  &  &  &  &  &  &  &  &  &  & \\

\midrule
Banking1 & $3^{4}4^{1}$ & $5^{112}$ & 560 & 15 & 13* & 11 & 102 & 0 & 1938 & 5864 & 19573 & 0.11 \\
Banking2 & $2^{14}4^{1}$ & $2^{3}$ & 6 & 11 & 10* & 7 & 473 & 3 & 5591 & 12845 & 34672 & 0.21 \\
CommProtocol & $2^{10}7^{1}$ & \makecell[cl]{$2^{10}3^{10}4^{12}5^{24}$ \\ $6^{30}7^{30}8^{12}$} & 704 & 19 & 16* & 13 & 285 & 35 & 6047 & 15914 & 50363 & 0.29 \\
Concurrency & $2^{5}$ & $2^{4}3^{1}5^{2}$ & 21 & 6 & 5* & 3 & 36 & 4 & 278 & 667 & 1686 & 0.01 \\
Healthcare1 & $2^{6}3^{2}5^{1}6^{1}$ & $2^{3}3^{18}$ & 60 & 30 & 30* & 29 & 361 & 8 & 12090 & 24661 & 70619 & 0.44 \\
Healthcare2 & $2^{5}3^{6}4^{1}$ & $2^{1}3^{6}5^{18}$ & 110 & 16 & 14 & 11 & 466 & 1 & 8212 & 18853 & 52268 & 0.31 \\
Healthcare3 & $2^{16}3^{6}4^{5}5^{1}6^{1}$ & $2^{31}$ & 62 & 38 & 34* & 29 & 3092 & 59 & 121950 & 271023 & 768538 & 5.38 \\
Healthcare4 & $2^{13}3^{12}4^{6}5^{2}6^{1}7^{1}$ & $2^{22}$ & 44 & 49 & 46* & 41 & 5707 & 38 & 287980 & 619634 & 1783211 & 13.14 \\
Insurance & $2^{6}3^{1}5^{1}6^{2}11^{1}13^{1}17^{1}31^{1}$ & - & 0 & 527 & 527* & 526 & 4573 & 0 & 2509047 & 5009492 & 14863678 & 122.95 \\
NetworkMgmt & $2^{2}4^{1}5^{3}10^{2}11^{1}$ & $2^{20}$ & 40 & 112 & 110* & 109 & 1228 & 20 & 148402 & 301059 & 877206 & 6.28 \\
ProcessorComm1 & $2^{3}3^{6}4^{6}$ & $2^{13}$ & 26 & 29 & 21 & 15 & 1058 & 13 & 32957 & 80475 & 219601 & 1.54 \\
ProcessorComm2 & $2^{3}3^{12}4^{8}5^{2}$ & $1^{4}2^{121}$ & 246 & 32 & 25* & 24 & 2525 & 854 & 85287 & 193248 & 541399 & 3.73 \\
Services & $2^{3}3^{4}5^{2}8^{2}10^{2}$ & $3^{386}4^{2}$ & 1166 & 106 & 100* & 99 & 1819 & 16 & 204692 & 460866 & 1346965 & 9.78 \\
Storage1 & $2^{1}3^{1}4^{1}5^{1}$ & $4^{95}$ & 380 & 17 & 17* & 14 & 53 & 18 & 1294 & 4270 & 13468 & 0.07 \\
Storage2 & $3^{4}6^{1}$ & - & 0 & 18 & 18* & 17 & 126 & 0 & 2826 & 5652 & 15552 & 0.09 \\
Storage3 & $2^{9}3^{1}5^{3}6^{1}8^{1}$ & $2^{38}3^{10}$ & 106 & 50 & 50* & 39 & 1020 & 120 & 54810 & 122009 & 344328 & 2.47 \\
Storage4 & $2^{5}3^{7}4^{1}5^{2}6^{2}7^{1}10^{1}13^{1}$ & $2^{24}$ & 48 & 136 & 130* & 129 & 3491 & 24 & 495046 & 1012862 & 2970799 & 22.45 \\
Storage5 & $2^{5}3^{8}5^{3}6^{2}8^{1}9^{1}10^{2}11^{1}$ & $2^{151}$ & 302 & 218 & 215 & 109 & 5342 & 246 & 1206084 & 3020149 & 8366680 & 72.16 \\
SystemMgmt & $2^{5}3^{4}5^{1}$ & $2^{13}3^{4}$ & 38 & 17 & 15* & 14 & 310 & 14 & 5935 & 12813 & 35376 & 0.21 \\
Telecom & $2^{5}3^{1}4^{2}5^{1}6^{1}$ & $2^{11}3^{1}4^{9}$ & 61 & 32 & 30* & 29 & 440 & 11 & 15650 & 32761 & 93262 & 0.59 \\
\midrule

Yu et al. \cite{Yu15} &  &  &  &  &  &  &  &  &  &  &  & \\

\midrule
RL-A & $2^{5}3^{4}4^{7}5^{4}6^{5}7^{4}8^{1}12^{3}$ & $1^{12}2^{491}3^{345}$ & 2029 & 155 & 153 & 143 & 7066 & 7156 & 1142671 & 2491775 & 7220414 & 59.32 \\
RL-B & \makecell[cl]{$2^{8}3^{2}4^{3}5^{3}6^{1}9^{1}$ \\ $10^{1}12^{2}14^{3}20^{1}24^{1}37^{1}$} & \makecell[cl]{$1^{8}2^{1127}3^{277}$ \\ $4^{1755}5^{1064}6^{2048}$} & 27721 & 767 & 727 & 519 & 17018 & 5597 & 13365222 & 35733711 & 109278283 & 1026.60 \\
\midrule %

Yamada et al. \cite{Yamada16} &  &  &  &  &  &  &  &  &  &  &  & \\

\midrule
Company2 & $2^{6}3^{4}8^{4}$ & \makecell[cl]{$1^{2}2^{35}3^{89}4^{54}5^{34}$ \\ $6^{20}7^{34}8^{16}9^{4}$} & 1247 & 81 & 72 & 55 & 1149 & 261 & 100546 & 252543 & 744203 & 5.35 \\
\bottomrule
\end{tabular}

}
\caption{General information of all benchmarks used.}
\label{tab:inst-sizes}
\end{table}

Regarding existing tools for solving Mixed Covering Arrays with Constraints, the main tool we compare with is CALOT \cite{YamadaKACOB15}. Unfortunately, CALOT is not available from the authors but we did our best to reproduce it (see section \ref{sec:calot}), showing our experimental investigation that the results are consistent with those of \cite{YamadaKACOB15}. Our implementation of CALOT and all algorithms presented in this paper will be available for reproducibility, which we think is also a nice contribution  for both the combinatorial testing and satisfiability communities.

Since all the algorithms presented in this paper are built on top of a SAT solver, we compared, when possible, all the algorithms with the same underlying SAT solver. That is not the case in \cite{YamadaKACOB15}, which may lead to flawed conclusions. In our experimental investigation we choose Glucose (version 4.1) \cite{audemard2013improving}, as most of the state-of-the-art MaxSAT solvers are built on top of it.

We also use the ACTS tool \cite{BorazjanyYLKK12} to compute fast and good enough upper bounds of the  Covering Array Number problem, although it is not competitive with SAT-based approaches.

The environment of execution consists of a computer cluster with machines equipped with two Intel Xeon Silver 4110 (octa-core processors at 2.1GHz, 11MB cache memory) and 96GB DDR4 main memory.  Unless otherwise stated, all the experiments were executed with a timeout of 2h and a memory limit of 18GB. To mitigate the impact of randomness we executed all the algorithms using five different seeds for each instance.

The rest of the experimental section is organized as follows.
Regarding the Covering Array Number, in subsection \ref{sec:exp-inc}, we compare the CALOT algorithm with the MaxSAT encodings and SAT-based MaxSAT approaches described in sections \ref{sec:mcacMaxSAT} and \ref{sec:MaxSATbased}. 
Regarding the Tuple Number problem, in subsection \ref{sec:com_incomp_OSCAR}, we evaluate the complete and incomplete MaxSAT algorithms on the encoding described in section \ref{sec:shortening}. Then, in subsection \ref{sec:incrementalOSCAR}, we evaluate the incomplete approach for computing the Tuple Number described in section \ref{sec:incompleteTN}.

\subsection{SAT-based MaxSAT approaches for the Covering Array Number problem}
\label{sec:exp-inc}

In this experiment, we compare the performance of state-of-the-art SAT-based MaxSAT solvers with the CALOT algorithm described in section \ref{sec:calot}. We hypothesise that since these SAT-based MaxSAT algorithms, once executed on the suitable MaxSAT encodings, can simulate the behaviour of the CALOT algorithm (see Propositions \ref{prop:linear-wpm-calot} and \ref{prop:wpm1-wpm-calot}) but the opposite is not true, MaxSAT algorithms may perform similarly or outperform the CALOT algorithm. This hypothesis would contradict the findings in \cite{YamadaKACOB15}, where it was reported that the CALOT algorithm clearly dominates the MaxSAT-based approach in \cite{Ansotegui2013}. If our hypothesis is correct, MaxSAT approaches for solving the Covering Array Number problem would be put back on the agenda. We focus in anytime algorithms that must be able to report suboptimal solutions \footnote{We adapted RC2 MaxSAT solver to report suboptimal solutions when applying the stratified strategy (see section \ref{sec:MaxSATbased})}.

{\bf Solvers:} The \ref{alg:calot} algorithm (described in section \ref{sec:calot}) and the model-guided Linear SAT-based MaxSAT algorithm \ref{alg:linear} (described in section \ref{sec:MaxSATbased}) were implemented on top of a custom python framework for SAT solving. This framework includes python bindings for several state-of-the-art SAT solvers and the python binding to the PBLib \cite{pypblib}.

We additionally tested several complete and incomplete algorithms from the MaxSAT Evaluation 2020 \cite{bacchus2020maxsat}. From complete MaxSAT solvers we tested MaxHS \cite{bacchus2020maxhs}, EvalMaxSAT \cite{avellaneda2020short}, RC2 \cite{ignatiev2020rc2} and maxino \cite{alviano2015maxsat}. We only report results for RC2 and one seed\footnote{Unfortunately \emph{RC2} MaxSAT solver does not allow to specify a seed.}, as this was the complete solver that reported better results. MaxHS obtained the best results for 2 of the tested instances, but we decided to exclude it from the comparison since it cannot report upper bounds for most of the instances and it uses another underlying SAT solver different than Glucose41.

Regarding incomplete MaxSAT algorithms we tested Loandra \cite{berg2020loandra}, tt-open-wbo-inc \cite{nadel2020tt} and SatLike \cite{ijcai2018-187}. We report results for Loandra and tt-open-wbo-inc as SatLike crashed in some of the tested instances.

{\bf MaxSAT encodings:} Respect to the MaxSAT encodings we report results on $PMSat_{CCX}^{N,t,S,lb}$ and the weighted version $WPMSat_{CCX}^{N,t,S,lb}$ using a linear increase for the weights ($w_i = i-(lb+2)+1$, see equation \ref{WSOFTU} in section \ref{sec:mcacMaxSAT}). We found that $WPMSat_{CCX}^{N,t,S,lb}$ with the linear and exponential increase ($w_i = 2^{i-(lb+2)}$) lead to the same performance, but the exponential increase represented a problem for some MaxSAT solvers when $i$ was high enough. 

We further tested the three different alternatives for equation (a) from $CCX$, where two reported good results. The first one is the original (a) equation shown in section \ref{sec:mcacSAT}, $(c^{i}_{\tau} \rightarrow c^{i-1}_{\tau} \vee x_{i,p,v})$, which we will refer to as a.0. The second one is the variation $(c^{i}_{\tau} \rightarrow c^{i-1}_{\tau} \vee x_{i,p,v}) \wedge (c^{i}_{\tau} \leftarrow c^{i-1}_{\tau})$, which we will refer to as a.1. 

{\bf Results:} Table \ref{tab:ca-res} shows the results of our experimentation. For each row and solver column, we give the average size of the minimum MCAC (out of the 5 executions per instance) and the average runtime. Bold values represent the best results. In case there are ties in size, the best time is marked. Sizes that have a star represent that the optimum has been certified in at least one of the five seeds executed for the current benchmark instance.

Table \ref{tab:ca-dom} aggregates the information presented in Table \ref{tab:ca-res} to analyze the dominance relations among approaches, i.e., the number of instances where algorithm A finds a smaller MCAC than algorithm B (size) and the number of instances where A needs less runtime than B (time). If A finds a smaller MCAC than B, then it is also considered that it needs less runtime. In this sense, we will say that an approach outperforms another if it provides a strictly better solution within the given timeout or finds the same best suboptimal solution faster.

\begin{sidewaystable}
\centering
\resizebox{\textwidth}{!}{
\begin{tabular}{l|cr|cr|cr|cr|cr|cr|cr|cr|cr|cr|cr|cr}
\toprule
\multicolumn{1}{c|}{\textbf{}} & \multicolumn{2}{c|}{ACTS} & \multicolumn{2}{c|}{\makecell{CALOT \\CCX a.0}} & \multicolumn{2}{c|}{\makecell{CALOT \\CCX a.1}} & \multicolumn{2}{c|}{\makecell{RC2-B \\CCX a.0}} & \multicolumn{2}{c|}{\makecell{RC2-B \\CCX a.0 wpm}} & \multicolumn{2}{c|}{\makecell{linear \\CCX a.0}} & \multicolumn{2}{c|}{\makecell{loandra \\CCX a.0}} & \multicolumn{2}{c|}{\makecell{loandra \\CCX a.1}} & \multicolumn{2}{c|}{\makecell{tt-open-wbo-inc \\CCX a.0}} & \multicolumn{2}{c|}{\makecell{tt-open-wbo-inc \\CCX a.1}} & \multicolumn{2}{c|}{\makecell{tt-open-wbo-inc \\CCX a.0 wpm}} & \multicolumn{2}{c|}{\makecell{tt-open-wbo-inc \\CCX a.1 wpm}} \\
Instance & size & time & size & time & size & time & size & time & size & time & size & time & size & time & size & time & size & time & size & time & size & time & size & time  \\
\midrule
Cohen et al. \cite{cohen2008constructing} \\
\midrule
1 & 48.00 & 0.83 & 38.00 & 1688.39 & 38.00 & 1470.88 & -1.00 & -1.00 & 37.00 & 7146.52 & 37.00 & 5367.02 & 37.00 & 1718.97 & \textbf{37.00} & \textbf{426.06} & 37.00 & 902.46 & 38.00 & 133.26 & 37.00 & 4724.61 & 38.00 & 1196.04 \\
2 & 32.00 & 1.08 & 30.00* & 2.80 & 30.00* & 3.57 & 30.00* & 9.85 & 30.00* & 10.81 & 30.00* & 2.91 & 30.00* & 7.19 & 30.00* & 7.68 & \textbf{30.00} & \textbf{0.81} & 30.00 & 1.42 & 30.00 & 0.92 & 30.00 & 1.70 \\
3 & 19.00 & 1.21 & 18.00* & 0.09 & 18.00* & 0.12 & 18.00* & 0.59 & 18.00* & 0.73 & 18.00* & 0.13 & 18.00* & 0.36 & 18.00* & 0.47 & 18.00 & 0.06 & 18.00 & 0.08 & \textbf{18.00} & \textbf{0.05} & 18.00 & 0.07 \\
4 & 22.00 & 1.04 & 20.00* & 0.52 & 20.00* & 0.95 & 20.00* & 2.58 & 20.00* & 3.19 & 20.00* & 0.74 & 20.00* & 1.34 & 20.00* & 2.24 & 20.00 & 0.30 & \textbf{20.00} & \textbf{0.29} & 20.00 & 0.31 & 20.00 & 0.45 \\
5 & 54.00 & 1.31 & 47.00 & 1955.92 & 48.00 & 241.15 & -1.00 & -1.00 & 48.00 & 378.46 & 47.00 & 1693.29 & 46.00 & 1629.29 & \textbf{\textbf{45.80}} & \textbf{1795.81} & 46.00 & 1624.87 & 46.00 & 1348.31 & 46.20 & 1803.99 & 46.00 & 4655.60 \\
6 & 25.00 & 1.11 & 24.00* & 1.18 & 24.00* & 1.46 & 24.00* & 4.46 & 24.00* & 5.47 & 24.00* & 1.48 & 24.00* & 2.98 & 24.00* & 3.51 & \textbf{24.00} & \textbf{0.33} & 24.00 & 0.48 & 24.00 & 0.34 & 24.00 & 0.47 \\
7 & 12.00 & 1.13 & 9.00 & 0.06 & 9.00 & 0.08 & -1.00 & -1.00 & 9.00 & 0.61 & 9.00 & 0.13 & 9.00 & 30.98 & 9.00 & 30.57 & 9.00 & 0.06 & 9.00 & 0.06 & \textbf{9.00} & \textbf{0.05} & 9.00 & 0.06 \\
8 & 47.00 & 1.01 & 38.00 & 1403.51 & 37.00 & 4727.41 & -1.00 & -1.00 & 37.00 & 3407.87 & 38.00 & 547.77 & 36.80* & 2849.90 & 37.00 & 3731.63 & 37.00 & 1178.58 & \textbf{\textbf{36.60}} & \textbf{2948.62} & 37.00 & 2333.83 & 37.00 & 5133.82 \\
9 & 22.00 & 1.21 & 20.00* & 0.57 & 20.00* & 0.84 & 20.00* & 2.36 & 20.00* & 2.66 & 20.00* & 0.49 & 20.00* & 1.22 & 20.00* & 2.30 & \textbf{20.00} & \textbf{0.18} & 20.00 & 0.27 & 20.00 & 0.19 & 20.00 & 0.41 \\
10 & 47.00 & 1.28 & 44.00 & 6860.63 & 44.20 & 5250.31 & -1.00 & -1.00 & 44.00 & 4759.95 & 44.00 & 1343.51 & 41.80 & 1235.36 & 41.00 & 1891.73 & 41.00 & 2720.18 & 41.00 & 970.10 & 41.00 & 5823.09 & \textbf{41.00} & \textbf{641.10} \\
11 & 47.00 & 1.26 & 41.00 & 2641.09 & 41.00 & 4167.69 & -1.00 & -1.00 & 42.00 & 6172.69 & 42.00 & 1916.85 & 39.00 & 6308.22 & 40.00 & 4323.47 & 40.00 & 741.96 & 40.00 & 1086.24 & \textbf{39.00} & \textbf{1628.02} & 39.00 & 2892.01 \\
12 & 43.00 & 1.29 & 36.00* & 14.13 & 36.00* & 24.01 & 36.00* & 36.81 & 36.00* & 49.94 & 36.00* & 20.66 & 36.00* & 36.16 & 36.00* & 42.45 & \textbf{36.00} & \textbf{8.18} & 36.00 & 10.46 & 36.00 & 10.75 & 36.00 & 11.37 \\
13 & 40.00 & 1.15 & 36.00* & 6.78 & 36.00* & 9.80 & 36.00* & 24.42 & 36.00* & 29.91 & 36.00* & 6.73 & 36.00* & 14.91 & 36.00* & 25.04 & 36.00 & 3.46 & 36.00 & 3.31 & \textbf{36.00} & \textbf{3.23} & 36.00 & 3.60 \\
14 & 39.00 & 0.83 & 36.00* & 3.76 & 36.00* & 5.25 & 36.00* & 13.81 & 36.00* & 15.56 & 36.00* & 4.80 & 36.00* & 8.84 & 36.00* & 11.51 & 36.00 & 1.89 & 36.00 & 2.48 & \textbf{36.00} & \textbf{1.51} & 36.00 & 4.28 \\
15 & 32.00 & 1.05 & 30.00* & 1.02 & 30.00* & 1.49 & 30.00* & 4.73 & 30.00* & 4.88 & 30.00* & 1.30 & 30.00* & 2.91 & 30.00* & 3.70 & \textbf{30.00} & \textbf{0.47} & 30.00 & 0.66 & 30.00 & 0.90 & 30.00 & 0.77 \\
16 & 25.00 & 0.84 & 24.00* & 1.36 & 24.00* & 1.94 & 24.00* & 5.59 & 24.00* & 6.60 & 24.00* & 1.65 & 24.00* & 4.30 & 24.00* & 4.46 & \textbf{24.00} & \textbf{0.56} & 24.00 & 0.60 & 24.00 & 0.57 & 24.00 & 0.62 \\
17 & 41.00 & 1.17 & 36.00* & 10.16 & 36.00* & 13.60 & 36.00* & 30.57 & 36.00* & 35.75 & 36.00* & 10.98 & 36.00* & 22.72 & 36.00* & 31.18 & \textbf{36.00} & \textbf{5.88} & 36.00 & 7.48 & 36.00 & 7.31 & 36.00 & 6.94 \\
18 & 52.00 & 1.30 & 42.00 & 128.65 & 42.00 & 201.42 & -1.00 & -1.00 & 42.00 & 177.28 & 42.00 & 202.10 & 41.00 & 637.62 & 41.00 & 734.99 & \textbf{41.00} & \textbf{134.63} & 41.00 & 626.83 & 41.00 & 2943.99 & 41.00 & 1871.62 \\
19 & 51.00 & 1.43 & 46.00 & 178.56 & 45.00 & 118.23 & -1.00 & -1.00 & 46.00 & 265.76 & 46.00 & 128.89 & 44.00 & 2056.11 & \textbf{\textbf{43.80}} & \textbf{2638.11} & 44.00 & 755.22 & 44.00 & 756.79 & 44.00 & 2892.85 & 44.00 & 578.78 \\
20 & 60.00 & 1.53 & 57.00 & 129.90 & 57.00 & 224.31 & -1.00 & -1.00 & 57.00 & 236.47 & 56.00 & 1092.85 & 54.80 & 1422.09 & \textbf{\textbf{54.60}} & \textbf{1736.92} & 55.00 & 32.47 & 55.00 & 500.42 & 55.00 & 437.94 & 55.00 & 176.33 \\
21 & 39.00 & 1.33 & 36.00* & 2.59 & 36.00* & 3.86 & 36.00* & 9.99 & 36.00* & 9.95 & 36.00* & 3.40 & 36.00* & 7.08 & 36.00* & 8.88 & \textbf{36.00} & \textbf{1.06} & 36.00 & 1.38 & 36.00 & 1.72 & 36.00 & 1.94 \\
22 & 37.00 & 1.08 & 36.00* & 1.75 & 36.00* & 2.70 & 36.00* & 7.74 & 36.00* & 8.57 & 36.00* & 2.17 & 36.00* & 5.25 & 36.00* & 6.78 & \textbf{36.00} & \textbf{0.65} & 36.00 & 0.88 & 36.00 & 0.72 & 36.00 & 1.04 \\
23 & 14.00 & 0.75 & 12.00* & 0.05 & 12.00* & 0.09 & 12.00* & 0.42 & 12.00* & 0.41 & 12.00* & 0.08 & 12.00* & 0.22 & 12.00* & 0.30 & \textbf{12.00} & \textbf{0.03} & 12.00 & 0.05 & 12.00 & 0.06 & 12.00 & 0.09 \\
24 & 48.00 & 1.40 & 44.40 & 4172.55 & 45.00 & 185.06 & -1.00 & -1.00 & 44.00 & 41.40 & 45.00 & 13.58 & 42.00 & 1693.19 & \textbf{\textbf{41.00}} & \textbf{4304.36} & 42.00 & 1465.17 & 43.00 & 59.40 & 42.00 & 884.17 & 42.00 & 926.51 \\
25 & 52.00 & 1.28 & 52.00 & 100.56 & 52.00 & 117.86 & -1.00 & -1.00 & -1.00 & -1.00 & 52.00 & 423.91 & 50.00 & 2069.85 & \textbf{\textbf{49.80}} & \textbf{1133.70} & 51.00 & 136.11 & 50.00 & 2101.71 & 50.00 & 5892.75 & 50.00 & 263.99 \\
26 & 34.00 & 1.37 & 27.00 & 49.99 & 27.00 & 31.54 & -1.00 & -1.00 & 27.00 & 112.13 & 27.00 & 551.57 & 27.00 & 103.46 & 27.00 & 57.66 & 26.00 & 6740.38 & \textbf{26.00} & \textbf{790.74} & 27.00 & 210.04 & 27.00 & 141.21 \\
27 & 37.00 & 1.32 & 36.00* & 1.14 & 36.00* & 1.68 & 36.00* & 5.35 & 36.00* & 6.70 & 36.00* & 1.33 & 36.00* & 2.96 & 36.00* & 3.56 & \textbf{36.00} & \textbf{0.39} & 36.00 & 0.61 & 36.00 & 0.45 & 36.00 & 0.62 \\
28 & 57.00 & 1.54 & 53.00 & 95.96 & 53.00 & 78.94 & -1.00 & -1.00 & 53.00 & 330.79 & 53.00 & 120.36 & 51.20 & 333.93 & 51.00 & 1236.07 & 51.00 & 3112.51 & 51.00 & 607.78 & 52.00 & 61.99 & \textbf{\textbf{49.00}} & \textbf{3329.73} \\
29 & 29.00 & 1.15 & 25.00* & 5.63 & 25.00* & 8.77 & 25.00* & 20.73 & 25.00* & 26.38 & 25.00* & 7.81 & 25.00* & 12.69 & 25.00* & 17.48 & \textbf{25.00} & \textbf{3.05} & 25.00 & 4.02 & 25.00 & 3.55 & 25.00 & 4.63 \\
30 & 22.00 & 0.81 & 16.00* & 1.38 & 16.00* & 1.75 & 16.00* & 4.53 & 16.00* & 5.98 & 16.00* & 1.56 & 16.00* & 3.57 & 16.00* & 3.94 & \textbf{16.00} & \textbf{0.68} & 16.00 & 0.83 & 16.00 & 0.77 & 16.00 & 1.08 \\
apache & 33.00 & 1.25 & 30.00* & 8.96 & 30.00* & 13.68 & 30.00* & 33.44 & 30.00* & 38.18 & 30.00* & 11.28 & 30.00* & 21.65 & 30.00* & 32.14 & \textbf{30.00} & \textbf{5.17} & 30.00 & 6.70 & 30.00 & 5.82 & 30.00 & 5.57 \\
bugzilla & 19.00 & 1.03 & 16.00* & 0.35 & 16.00* & 0.61 & 16.00* & 1.72 & 16.00* & 1.79 & 16.00* & 0.36 & 16.00* & 0.89 & 16.00* & 1.19 & \textbf{16.00} & \textbf{0.14} & 16.00 & 0.19 & 16.00 & 0.14 & 16.00 & 0.46 \\
gcc & 23.00 & 0.93 & 15.00 & 13.09 & 15.00 & 19.06 & -1.00 & -1.00 & 15.00 & 47.85 & 15.00 & 22.14 & 15.00 & 63.99 & 15.00 & 68.18 & \textbf{15.00} & \textbf{9.44} & 15.00 & 12.76 & 15.00 & 11.82 & 15.00 & 29.62 \\
spins & 26.00 & 1.22 & \textbf{19.00*} & \textbf{0.13} & 19.00* & 0.22 & 19.00* & 0.56 & 19.00* & 0.70 & 19.00* & 0.16 & 19.00* & 0.28 & 19.00* & 0.28 & 19.00 & 0.13 & 19.00 & 0.21 & 19.00 & 0.20 & 19.00 & 0.21 \\
spinv & 45.00 & 0.84 & 33.00 & 5.28 & 32.00 & 188.97 & -1.00 & -1.00 & 32.00 & 55.75 & 33.00 & 21.39 & 32.00 & 61.37 & 32.00 & 97.82 & \textbf{32.00} & \textbf{26.18} & 32.00 & 82.41 & 32.00 & 53.36 & 32.00 & 199.38 \\
\midrule
Segall et al. \cite{segall11} \\
\midrule
Banking1 & 15.00 & 0.85 & \textbf{13.00*} & \textbf{0.01} & 13.00* & 0.01 & 13.00* & 0.20 & 13.00* & 0.29 & 13.00* & 0.08 & 13.00* & 0.05 & 13.00* & 0.05 & 13.00 & 0.09 & 13.00 & 0.08 & 13.00 & 0.06 & 13.00 & 0.06 \\
Banking2 & 11.00 & 0.46 & \textbf{10.00*} & \textbf{0.01} & 10.00* & 0.02 & 10.00* & 25.56 & 10.00* & 22.95 & 10.00* & 0.03 & 10.00* & 19.30 & 10.00* & 10.62 & 10.00 & 0.02 & 10.00 & 0.02 & 10.00 & 0.02 & 10.00 & 0.02 \\
CommProtocol & 19.00 & 1.05 & \textbf{16.00*} & \textbf{0.02} & 16.00* & 0.03 & 16.00* & 0.19 & 16.00* & 0.22 & 16.00* & 0.03 & 16.00* & 0.15 & 16.00* & 0.09 & 16.00 & 0.05 & 16.00 & 0.03 & 16.00 & 0.02 & 16.00 & 0.02 \\
Concurrency & 6.00 & 0.49 & \textbf{5.00*} & \textbf{0.00} & 5.00* & 0.00 & 5.00* & 0.08 & 5.00* & 0.08 & 5.00* & 0.01 & 5.00* & 0.03 & 5.00* & 0.03 & 5.00 & 0.03 & 5.00 & 0.01 & 5.00 & 0.01 & 5.00 & 0.01 \\
Healthcare1 & 30.00 & 0.55 & 30.00* & 0.03 & 30.00* & 0.03 & 30.00* & 0.24 & 30.00* & 0.25 & 30.00* & 0.05 & 30.00* & 0.13 & 30.00* & 0.15 & 30.00 & 0.17 & 30.00 & 0.03 & \textbf{30.00} & \textbf{0.02} & 30.00 & 0.03 \\
Healthcare2 & 16.00 & 0.60 & \textbf{14.00} & \textbf{0.02} & 14.00 & 0.07 & -1.00 & -1.00 & 14.00 & 0.21 & 14.00 & 0.07 & 14.00 & 30.55 & 14.00 & 30.06 & 14.00 & 0.21 & 14.00 & 0.33 & 14.00 & 0.05 & 14.00 & 0.35 \\
Healthcare3 & 38.00 & 0.58 & 34.00* & 0.48 & 34.00* & 0.75 & 34.00* & 2.14 & 34.00* & 2.39 & 34.00* & 0.68 & 34.00* & 1.63 & 34.00* & 1.44 & \textbf{34.00} & \textbf{0.26} & 34.00 & 0.32 & 34.00 & 0.37 & 34.00 & 0.45 \\
Healthcare4 & 49.00 & 0.59 & 46.00* & 1.77 & 46.00* & 2.33 & 46.00* & 5.88 & 46.00* & 7.24 & 46.00* & 2.61 & 46.00* & 3.64 & 46.00* & 3.71 & \textbf{46.00} & \textbf{0.73} & 46.00 & 0.82 & 46.00 & 1.83 & 46.00 & 2.22 \\
Insurance & \textbf{527.00} & \textbf{0.31} & 527.00* & 16.82 & 527.00* & 18.56 & 527.00* & 44.42 & 527.00* & 45.47 & 527.00* & 21.08 & 527.00* & 45.23 & 527.00* & 46.40 & 527.00 & 5.29 & 527.00 & 10.47 & 527.00 & 5.08 & 527.00 & 11.25 \\
NetworkMgmt & 112.00 & 0.58 & 110.00* & 214.18 & 110.00* & 511.52 & 110.00* & 180.82 & 110.00* & 233.46 & 110.00* & 1100.79 & 110.00* & 45.65 & \textbf{110.00*} & \textbf{34.19} & 110.00 & 344.68 & 110.00 & 302.44 & 110.00 & 408.88 & 110.00 & 276.77 \\
ProcessorComm1 & 29.00 & 0.52 & 21.00 & 67.53 & 22.00 & 0.47 & -1.00 & -1.00 & 22.00 & 1.14 & \textbf{21.00} & \textbf{28.65} & 21.00 & 468.17 & 21.00 & 107.68 & 21.00 & 610.20 & 21.00 & 1690.90 & 22.00 & 0.22 & 21.00 & 1015.43 \\
ProcessorComm2 & 32.00 & 0.77 & 25.00* & 0.42 & 25.00* & 0.55 & 25.00* & 1.42 & 25.00* & 1.70 & 25.00* & 0.42 & 25.00* & 0.78 & 25.00* & 1.22 & \textbf{25.00} & \textbf{0.23} & 25.00 & 0.39 & 25.00 & 0.26 & 25.00 & 0.41 \\
Services & 106.00 & 1.20 & 100.00* & 28.65 & 100.00* & 29.89 & 100.00* & 40.55 & 100.00* & 37.37 & 100.00* & 26.69 & 100.00* & 3.94 & 100.00* & 3.97 & 100.00 & 2.43 & 100.00 & 2.44 & 100.00 & 2.42 & \textbf{100.00} & \textbf{1.46} \\
Storage1 & 17.00 & 0.75 & \textbf{17.00*} & \textbf{0.00} & 17.00* & 0.01 & 17.00* & 0.11 & 17.00* & 0.11 & 17.00* & 0.01 & -1.00 & -1.00 & -1.00 & -1.00 & 17.00 & 0.01 & 17.00 & 0.01 & 17.00 & 0.01 & 17.00 & 0.01 \\
Storage2 & 18.00 & 0.25 & \textbf{18.00*} & \textbf{0.01} & 18.00* & 0.01 & 18.00* & 0.12 & 18.00* & 0.12 & 18.00* & 0.02 & 18.00* & 0.05 & 18.00* & 0.05 & 18.00 & 0.02 & 18.00 & 0.01 & 18.00 & 0.01 & 18.00 & 0.01 \\
Storage3 & 50.00 & 0.66 & 50.00* & 0.14 & 50.00* & 0.28 & 50.00* & 11.02 & 50.00* & 1.86 & 50.00* & 0.26 & -1.00 & -1.00 & -1.00 & -1.00 & 50.00 & 0.09 & 50.00 & 0.11 & \textbf{50.00} & \textbf{0.08} & 50.00 & 0.12 \\
Storage4 & 136.00 & 0.60 & 130.00* & 2.49 & 130.00* & 3.24 & 130.00* & 7.24 & 130.00* & 9.28 & 130.00* & 1.69 & 130.00* & 5.16 & 130.00* & 6.65 & \textbf{130.00} & \textbf{0.55} & 130.00 & 0.84 & 130.00 & 0.58 & 130.00 & 0.83 \\
Storage5 & 218.00 & 0.87 & 215.00 & 68.48 & 215.00 & 389.33 & -1.00 & -1.00 & 215.00 & 140.68 & 215.00 & 811.43 & 215.00 & 85.07 & 215.00 & 90.98 & 215.00 & 17.01 & \textbf{215.00} & \textbf{16.83} & 215.00 & 45.27 & 215.00 & 46.54 \\
SystemMgmt & 17.00 & 0.57 & \textbf{15.00*} & \textbf{0.01} & 15.00* & 0.03 & 15.00* & 0.16 & 15.00* & 0.18 & 15.00* & 0.02 & 15.00* & 0.08 & 15.00* & 0.09 & 15.00 & 0.02 & 15.00 & 0.02 & 15.00 & 0.02 & 15.00 & 0.02 \\
Telecom & 32.00 & 0.58 & 30.00* & 0.04 & 30.00* & 0.05 & 30.00* & 0.31 & 30.00* & 0.37 & 30.00* & 0.06 & 30.00* & 0.18 & 30.00* & 0.22 & 30.00 & 0.03 & 30.00 & 0.04 & \textbf{30.00} & \textbf{0.03} & 30.00 & 0.04 \\
\midrule
Yu et al. \cite{Yu15} \\
\midrule
RL-A & 155.00 & 2.88 & 153.00 & 14.02 & 153.00 & 25.95 & -1.00 & -1.00 & 153.00 & 28.52 & 153.00 & 18.89 & 153.00 & 58.49 & 153.00 & 63.11 & \textbf{153.00} & \textbf{3.25} & 153.00 & 3.77 & 153.00 & 12.04 & 153.00 & 15.40 \\
RL-B & 767.00 & 170.61 & 760.00 & 4471.65 & 763.00 & 5862.61 & -1.00 & -1.00 & -1.00 & -1.00 & 764.00 & 5633.81 & -1.00 & -1.00 & -1.00 & -1.00 & -1.00 & -1.00 & \textbf{727.00} & \textbf{966.65} & -1.00 & -1.00 & 727.00 & 5809.65 \\
\midrule
Yamada et al. \cite{Yamada16} \\
\midrule
Company2 & 81.00 & 9.37 & 72.00 & 1.80 & 72.00 & 1.88 & -1.00 & -1.00 & 72.00 & 3.33 & 72.00 & 2.48 & 72.00 & 32.15 & 72.00 & 32.96 & \textbf{72.00} & \textbf{0.68} & 72.00 & 0.75 & 72.00 & 0.99 & 72.00 & 1.29 \\
\midrule
\bottomrule
\end{tabular}
}
\caption{Comparison of SAT-based MaxSAT approaches versus CALOT for the Covering Array Number problem.
Bold values represent the best results. In case of ties in size, the best time is marked. For sizes with a star the optimum has been certified in at least one of the five seeds executed.
}
\label{tab:ca-res}
\end{sidewaystable}

\begin{table}[h!t]
\centering
\resizebox{\textwidth}{!}{
\begin{tabular}{l|c|c|c|c|c|c|c|c|c|c|c|c|c|c|c|c|c|c|c|c|c|c}
\toprule
\multirow{2}{*}{} & \multicolumn{2}{c|}{\makecell{CALOT \\CCX a.0}} & \multicolumn{2}{c|}{\makecell{CALOT \\CCX a.1}} & \multicolumn{2}{c|}{\makecell{RC2-B \\CCX a.0}} & \multicolumn{2}{c|}{\makecell{RC2-B \\CCX a.0 wpm}} & \multicolumn{2}{c|}{\makecell{linear \\CCX a.0}} & \multicolumn{2}{c|}{\makecell{loandra \\CCX a.0}} & \multicolumn{2}{c|}{\makecell{loandra \\CCX a.1}} & \multicolumn{2}{c|}{\makecell{tt-open-wbo-inc \\CCX a.0}} & \multicolumn{2}{c|}{\makecell{tt-open-wbo-inc \\CCX a.1}} & \multicolumn{2}{c|}{\makecell{tt-open-wbo-inc \\CCX a.0 wpm}} & \multicolumn{2}{c|}{\makecell{tt-open-wbo-inc \\CCX a.1 wpm}} \\
 & size & time & size & time & size & time & size & time & size & time & size & time & size & time & size & time & size & time & size & time & size & time  \\
\midrule
ACTS & 0 \textbf{52} & 2 \textbf{56} & 0 \textbf{52} & 2 \textbf{56} & 21 \textbf{32} & 23 \textbf{35} & 2 \textbf{51} & 4 \textbf{54} & 0 \textbf{52} & 2 \textbf{56} & 3 \textbf{52} & 4 \textbf{54} & 3 \textbf{52} & 4 \textbf{54} & 1 \textbf{52} & 2 \textbf{56} & 0 \textbf{53} & 1 \textbf{57} & 1 \textbf{52} & 2 \textbf{56} & 0 \textbf{53} & 1 \textbf{57} \\
CALOT CCX a.0 & - - & - - & \textbf{5} 3 & \textbf{52} 6 & \textbf{21} 0 & \textbf{57} 1 & \textbf{5} 4 & \textbf{53} 5 & \textbf{3} 2 & \textbf{47} 11 & 3 \textbf{12} & \textbf{44} 14 & 3 \textbf{12} & \textbf{44} 14 & 1 \textbf{13} & 13 \textbf{45} & 0 \textbf{13} & 12 \textbf{46} & 2 \textbf{12} & 15 \textbf{43} & 0 \textbf{12} & 19 \textbf{39} \\
CALOT CCX a.1 & - - & - - & - - & - - & \textbf{21} 0 & \textbf{57} 1 & \textbf{4} 3 & \textbf{50} 8 & 5 \textbf{5} & 27 \textbf{31} & 3 \textbf{12} & \textbf{42} 16 & 3 \textbf{11} & \textbf{42} 16 & 1 \textbf{12} & 9 \textbf{49} & 0 \textbf{13} & 6 \textbf{52} & 1 \textbf{10} & 7 \textbf{51} & 0 \textbf{11} & 11 \textbf{47} \\
RC2-B CCX a.0 & - - & - - & - - & - - & - - & - - & 0 \textbf{19} & \textbf{31} 25 & 0 \textbf{21} & 1 \textbf{57} & 2 \textbf{20} & 3 \textbf{54} & 2 \textbf{20} & 6 \textbf{51} & 0 \textbf{20} & 1 \textbf{56} & 0 \textbf{21} & 1 \textbf{57} & 0 \textbf{20} & 1 \textbf{56} & 0 \textbf{21} & 1 \textbf{57} \\
RC2-B CCX a.0 wpm & - - & - - & - - & - - & - - & - - & - - & - - & 3 \textbf{5} & 7 \textbf{51} & 2 \textbf{11} & 8 \textbf{49} & 2 \textbf{10} & 10 \textbf{47} & 0 \textbf{11} & 1 \textbf{56} & 1 \textbf{13} & 4 \textbf{54} & 0 \textbf{9} & 2 \textbf{55} & 1 \textbf{11} & 6 \textbf{52} \\
linear CCX a.0 & - - & - - & - - & - - & - - & - - & - - & - - & - - & - - & 3 \textbf{11} & \textbf{41} 17 & 3 \textbf{11} & \textbf{41} 17 & 1 \textbf{12} & 7 \textbf{51} & 1 \textbf{13} & 4 \textbf{54} & 2 \textbf{11} & 3 \textbf{55} & 1 \textbf{12} & 7 \textbf{51} \\
loandra CCX a.0 & - - & - - & - - & - - & - - & - - & - - & - - & - - & - - & - - & - - & 2 \textbf{7} & \textbf{38} 17 & 4 \textbf{5} & 9 \textbf{48} & 4 \textbf{7} & 9 \textbf{49} & \textbf{5} 3 & 12 \textbf{45} & 3 \textbf{5} & 10 \textbf{48} \\
loandra CCX a.1 & - - & - - & - - & - - & - - & - - & - - & - - & - - & - - & - - & - - & - - & - - & \textbf{5} 3 & 13 \textbf{44} & \textbf{6} 5 & 9 \textbf{49} & \textbf{7} 3 & 13 \textbf{44} & \textbf{6} 5 & 13 \textbf{45} \\
tt-open-wbo-inc CCX a.0 & - - & - - & - - & - - & - - & - - & - - & - - & - - & - - & - - & - - & - - & - - & - - & - - & 2 \textbf{3} & \textbf{39} 19 & \textbf{4} 2 & \textbf{38} 19 & 2 \textbf{4} & \textbf{42} 16 \\
tt-open-wbo-inc CCX a.1 & - - & - - & - - & - - & - - & - - & - - & - - & - - & - - & - - & - - & - - & - - & - - & - - & - - & - - & \textbf{6} 3 & 21 \textbf{37} & 2 \textbf{3} & \textbf{39} 19 \\
tt-open-wbo-inc CCX a.0 wpm & - - & - - & - - & - - & - - & - - & - - & - - & - - & - - & - - & - - & - - & - - & - - & - - & - - & - - & - - & - - & 1 \textbf{4} & \textbf{43} 15 \\
\bottomrule
\end{tabular}
}
\caption{Dominance relations for CALOT and SAT-based MaxSAT approaches for the Covering Array Number problem.
Bold values highlight winning algorithm per size or runtime.
}
\label{tab:ca-dom}
\end{table}

We observe how both \emph{tt-open-wbo-inc} and \emph{loandra} outperform the results obtained by \emph{CALOT}, improving the sizes in more than 10 of the 58 available instances and, in the case of \emph{tt-open-wbo-inc}, we also improve runtimes in more than 40 instances. This confirms our hypothesis that MaxSAT approaches can simulate and even improve the results obtained by the \emph{CALOT} algorithm.

Regarding the different variations of the $CCX$ encoding, we notice that for \emph{tt-open-wbo-inc} and \emph{loandra}, variation a.1 slightly improves results obtained by the original variation a.0. In particular, we observe that \emph{tt-open-wbo-inc} with this specific encoding obtains the best size in instance \emph{RL-B} (727), while algorithm \emph{CALOT} reports a size of 760. However, this behaviour of the encoding a.1 is not observed in algorithm \emph{CALOT}, as in this case the best variation of equation (a) seems to be a.0. These results suggest that in case we use a new MaxSAT solver we should not discard at front any encoding variation.

For \emph{RC2} and \emph{linear} approaches we can observe clear differences among them when applying the $PMSat_{CCX}^{N,t,S,lb}$ encoding, as \emph{linear} obtains better sizes and times in 21 and 57 instances respectively, showing that for the Covering Array Number problem is more effective to perform a search that incrementally refines the upper bound as the linear approach does (see section \ref{sec:MaxSATbased}). However, we observe a substantial improvement when using the $WPMSat_{CCX}^{N,t,S,lb}$ with the \emph{RC2} MaxSAT solver, improving the sizes obtained by its unweighted counterpart in 19 of the 58 instances, which produces similar results than \emph{CALOT} and $PMSat_{CCX}^{N,t,S,lb}$ \emph{linear} approaches. This is expected since the weighted version forces RC2 to perform a top-down search as discussed in section \ref{sec:MaxSATbased}.

We also tested the $WPMSat_{CCX}^{N,t,S,lb}$ encoding over the \emph{tt-open-wbo-inc}, which is not core-guided MaxSAT solver. We observe that results are similar or slightly worse than with the $PMSat_{CCX}^{N,t,S,lb}$. We believe the $WPMSat_{CCX}^{N,t,S,lb}$ encoding could be more useful for core-guided MaxSAT solvers as it modifies their refinement strategy (i.e. improve the upper bound instead of the lower bound). We also observed that refining the lower bound for the Covering Array Number problem is more challenging than refining the upper bound, as there are some instances where encoding $PMSat_{CCX}^{N,t,S,lb}$ with \emph{RC2} (which would refine the lower bound) is not able to report any results, usually on instances where the CAN is not found. 

\subsection{Weighted Partial MaxSAT approaches for the Tuple Number problem}\label{sec:com_incomp_OSCAR}

Encouraged by the good results of the proposed MaxSAT approaches for the Covering Array Number problem, we now evaluate the MaxSAT approach described in section \ref{sec:shortening} on SAT-based MaxSAT approaches for solving the Tuple Number problem. Notice that the CALOT algorithm only works for solving the Covering Array Number problem. In this sense, this is a pioneering work on applying SAT technology to solve the Tuple Number problem.

\textbf{Solvers:} We choose the \emph{tt-open-wbo-inc} MaxSAT solver to perform these experiments, as this has been the approach that achieved better results in section \ref{sec:exp-inc}.

\textbf{MaxSAT encodings:} We recall there are also some variations of the $TPMSat_{CCX}^{N,t,S,lb}$ encoding, due to the way constraint $CCX$ is formulated, i.e. the relation among $c^{i}_{\tau}$ vars and $x_{i,p,v}$ vars (see remark \ref{rmrk:ccx-alternative} in section \ref{sec:mcacSAT}). According some preliminary experimentation we observed that variation $(c^{i}_{\tau} \leftrightarrow c^{i-1}_{\tau} \vee x_{i,p,v})$, to which we refer as a.2, reported also good results, while variation a.1 did not and was excluded. 

We additionally noticed that, when computing the tuple number, the cost of the solution returned by the MaxSAT solver when using the original encoding of equation (a) in $CCX$, $(c^{i}_{\tau} \rightarrow c^{i-1}_{\tau} \vee x_{i,p,v})$, can indeed overestimate the real cost of the solution induced by the value of the $x_{i,p,v}$ vars, i.e., the assignments that represent the actual tests used in the solution. This can happen since it is possible to set to False a $c^{i}_{\tau}$ even if the right-hand side of the implication is True. Enforcing the other side of the implication corrects this issue. For these reasons we will use the $(c^{i}_{\tau} \leftrightarrow c^{i-1}_{\tau} \vee x_{i,p,v})$ variation of $CCX$.

\textbf{Results:} We would like to study the evolution of the number of covered tuples as a function of the number of tests, as we hypothesise that adding a new test close to the Covering Array Number (that guarantees all tuples can be covered) will allow to add very few additional tuples. In that sense, if these tests are expensive enough, they will not pay off in terms of the available budget and the additional percentage of coverage we can achieve.

\begin{figure}[h!t]
    \caption{Number of tests required to reach a certain coverage percentage for the \emph{tt-open-wbo-inc} approach.}
    \label{fig:shortening-tt-open-wbo-inc}
    \centering
    \includegraphics[width=0.41\textwidth]{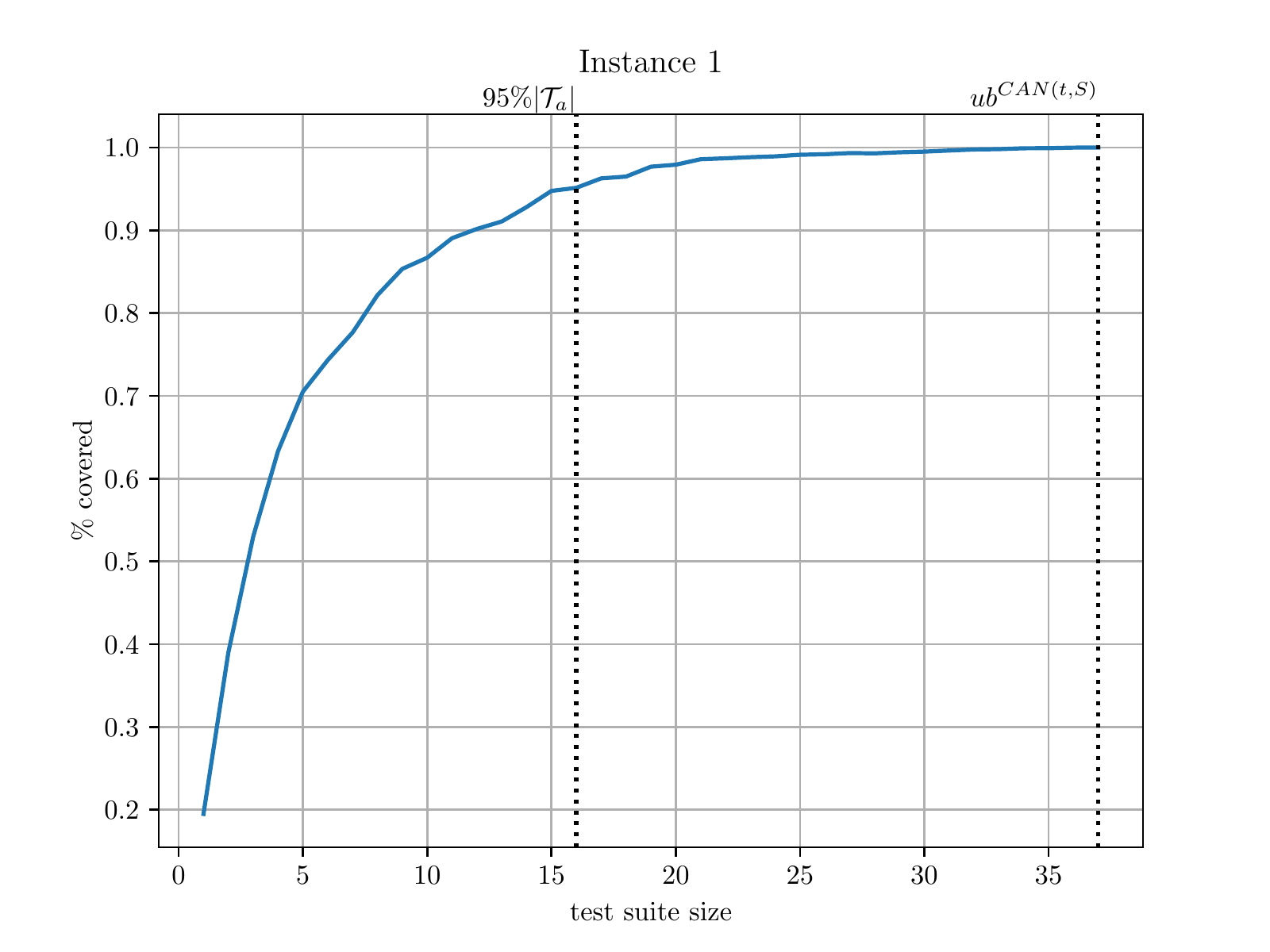}
    \includegraphics[width=0.41\textwidth]{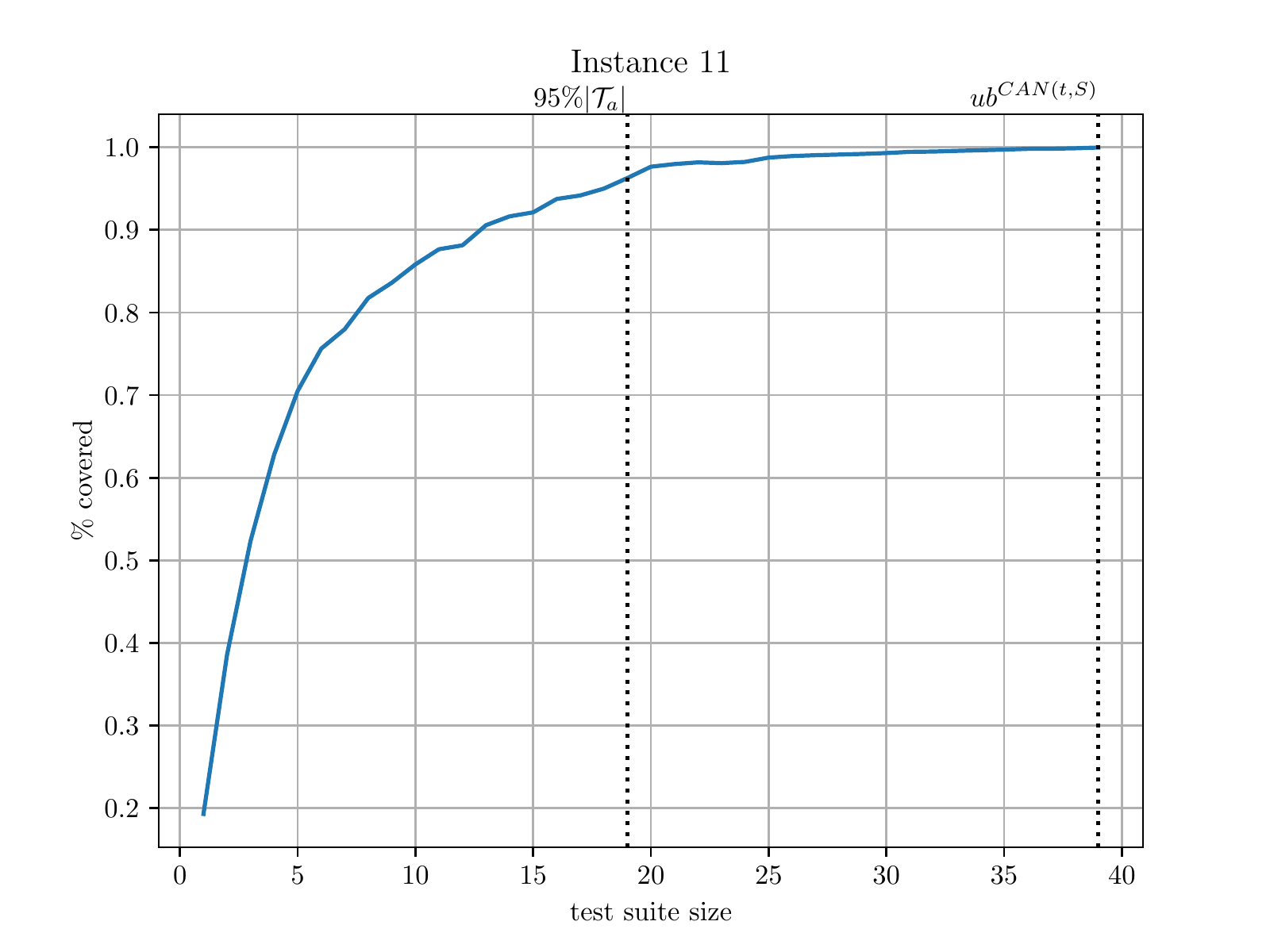} \\
    \includegraphics[width=0.41\textwidth]{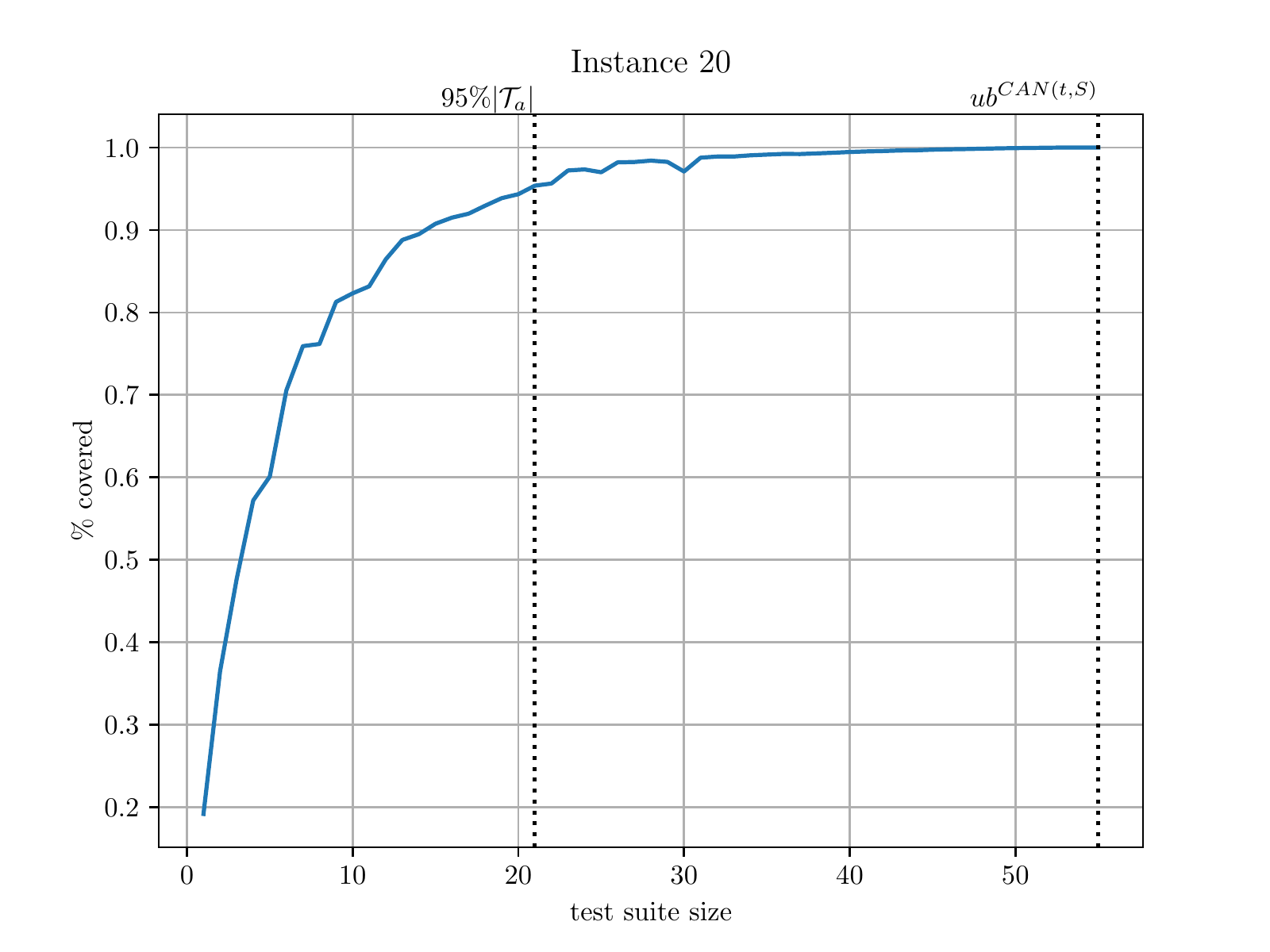}
    \includegraphics[width=0.41\textwidth]{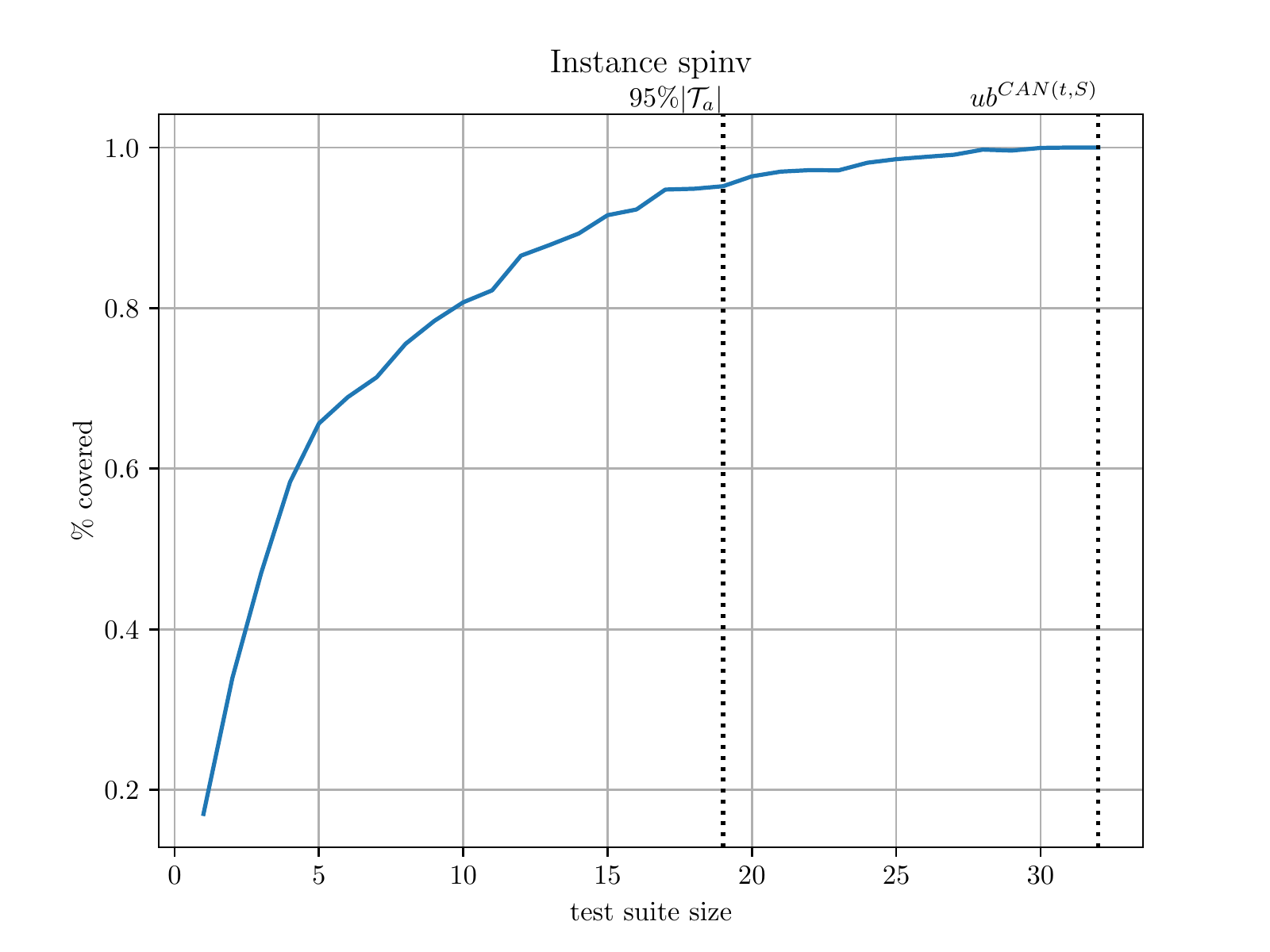} \\
    \includegraphics[width=0.41\textwidth]{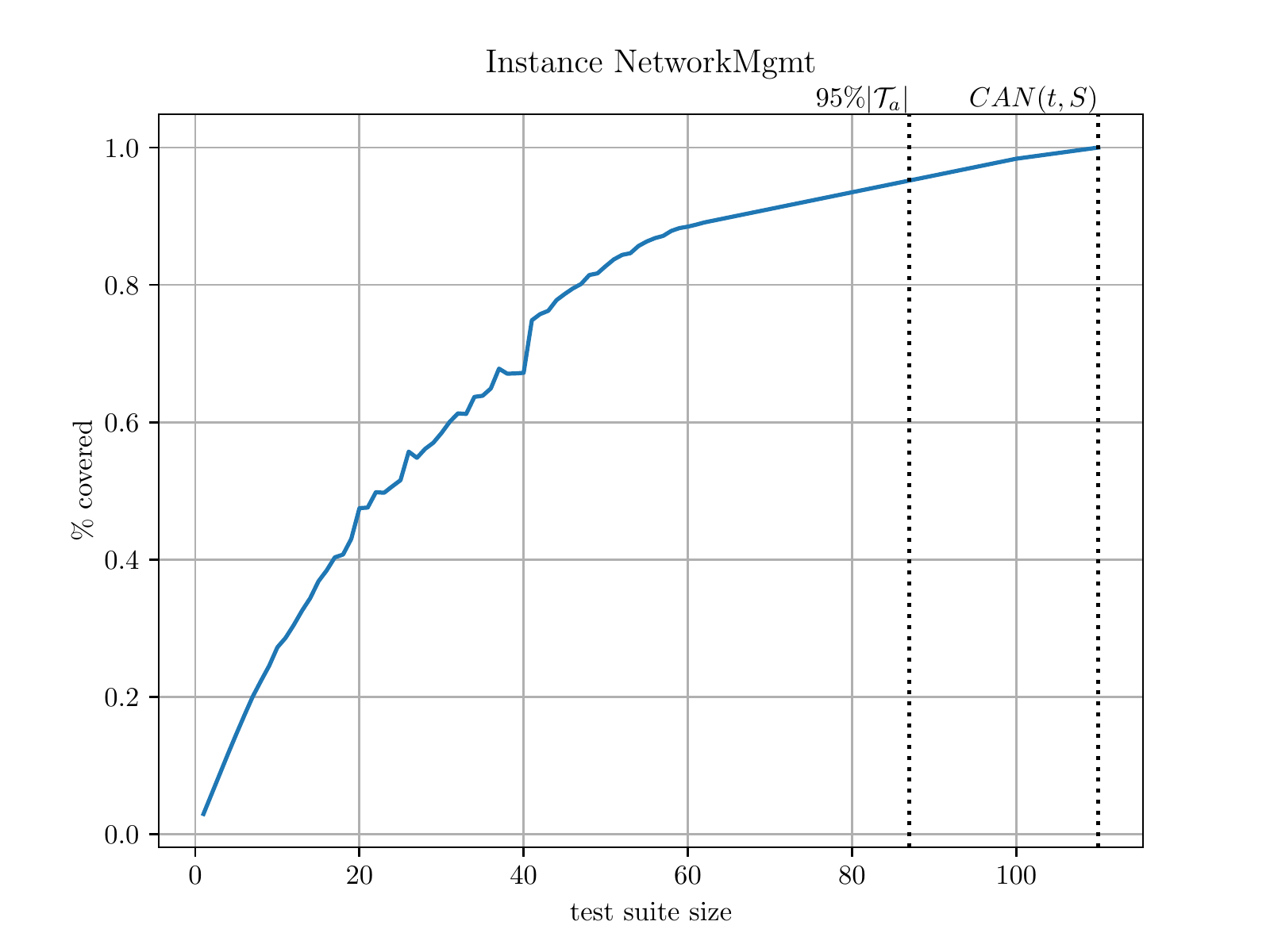}
    \includegraphics[width=0.41\textwidth]{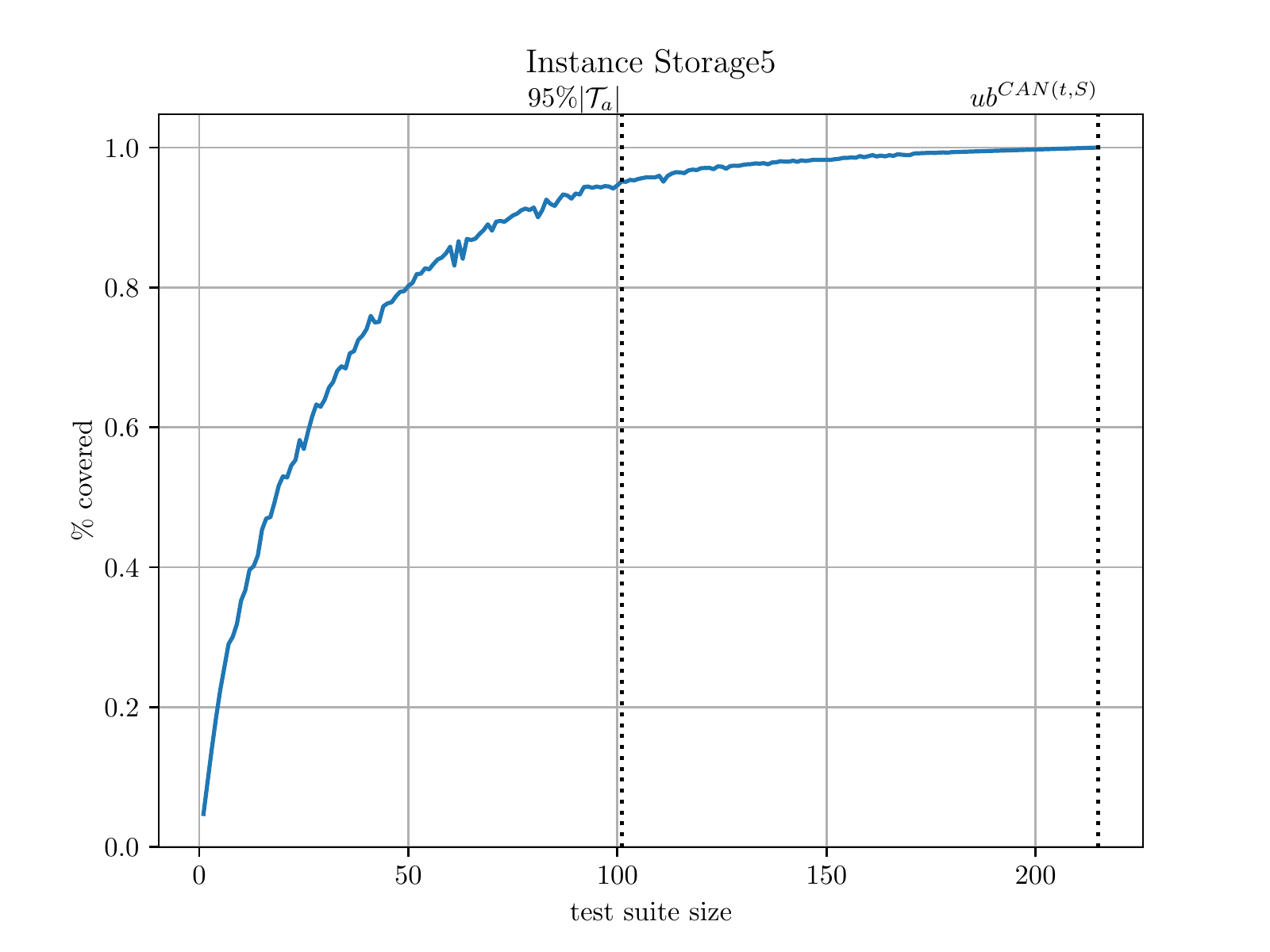} \\
    \includegraphics[width=0.41\textwidth]{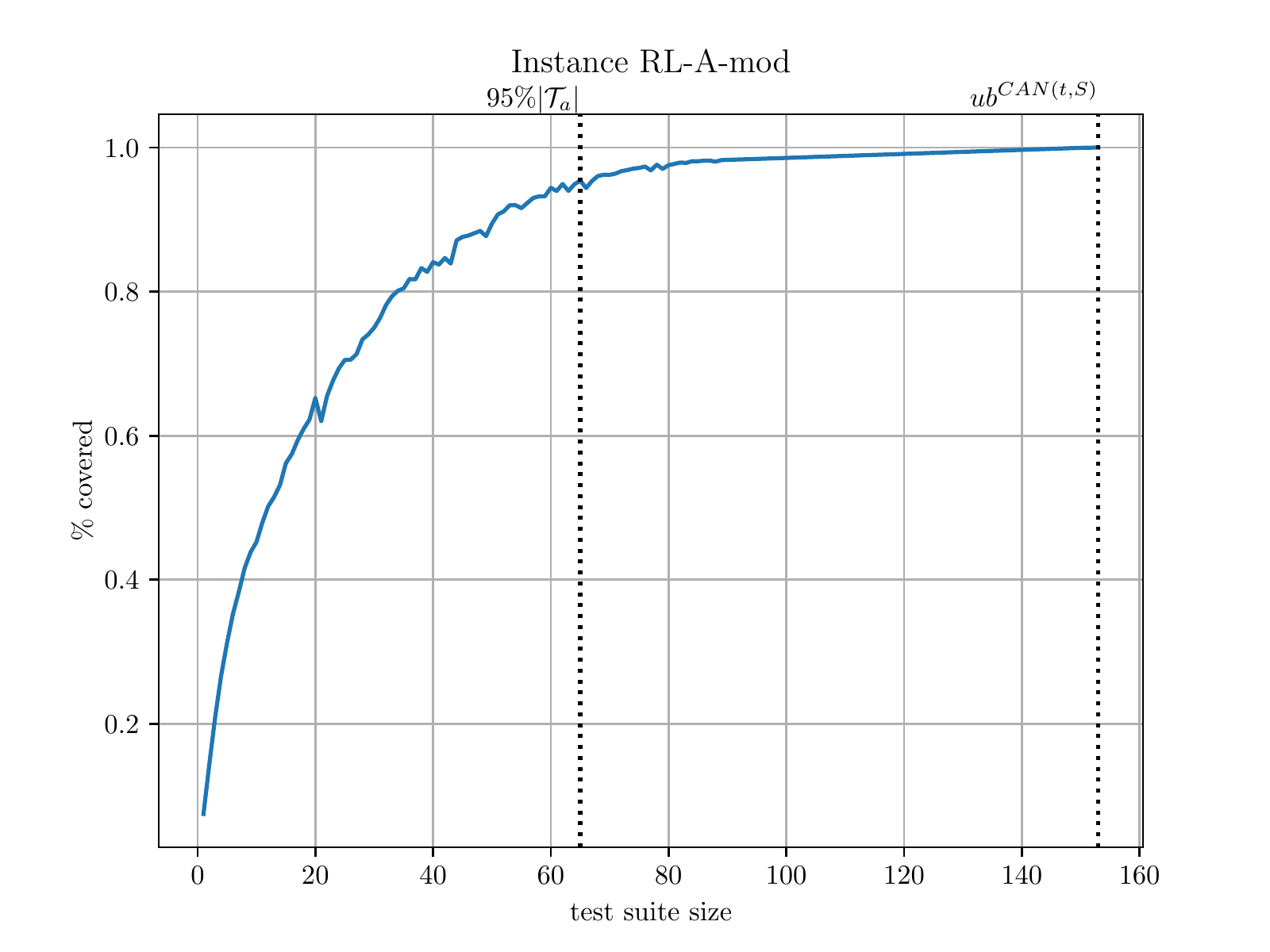}
    \includegraphics[width=0.41\textwidth]{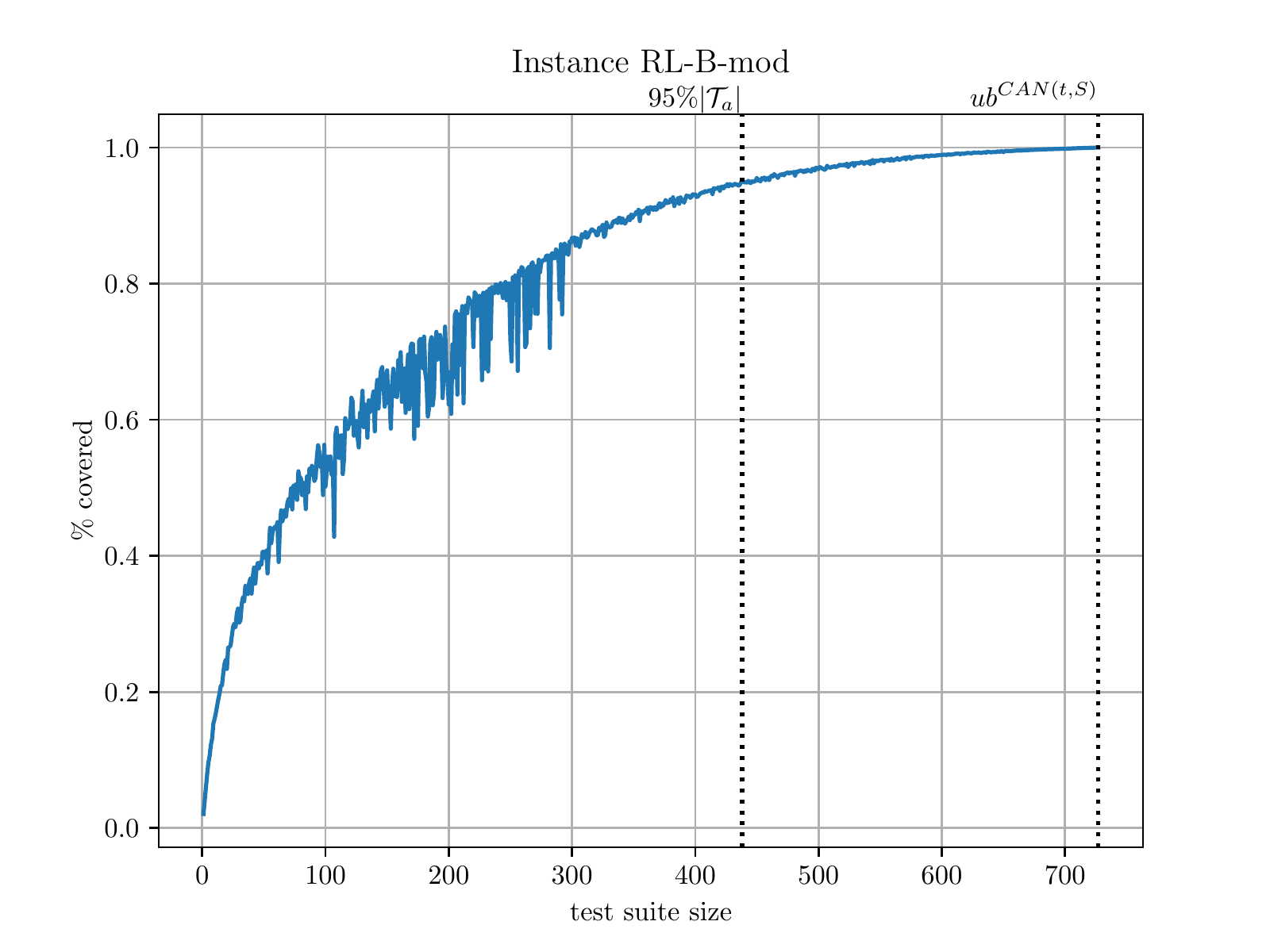}

\end{figure}

In Figure \ref{fig:shortening-tt-open-wbo-inc}, we show the number of tests required to reach a certain percentage of the tuples to cover for the \emph{tt-open-wbo-inc} approach. Notice that \emph{tt-open-wbo-inc} is an incomplete MaxSAT solver and we are therefore reporting a lower bound on the possible percentage by a particular number of tests. For lack of space, we only show the most representative instances of all the benchmark families.

We observe, for all the tested instances, that most of the tuples are covered using a relatively small number of tests and the remaining tuples require a relatively large additional number of tests. In our experiments, with only 52\% of tests required for the Covering Array Number or for the best suboptimal solution from Table \ref{tab:ca-res} in section \ref{sec:exp-inc}, we are able to reach a 95\% coverage, whereas the remaining 5\% of tuples need the remaining 48\% of tests.

We also notice that the Tuple Number problem is more challenging than the Covering Array Number problem. According some experimentation that we performed using complete MaxSAT solvers, none of the tested approaches has been able to certify any optimum for $N > 1$, even for the instances that were easy to solve for the Covering Array Number problem. 

Another interesting observation is the erratic behavior on the \emph{RL-B} instance \cite{Yu15} (Figure \ref{fig:shortening-tt-open-wbo-inc}, bottom right). \emph{RL-B} is the biggest instance in the available benchmarks, having 27 parameters with domains up to 37, and with a suboptimal solution for the Covering Array Number (for $t=2$) of 727 tests. After 100 tests, the results for the Tuple Number problem become quite unstable in contrast to the behaviour on the rest of instances. This phenomenon might point out that the approach analyzed in this section has some limitations when instances are large enough. For a fixed set of parameters, instances become bigger when we increase the strength $t$ or the number of tests as in this case.

To conclude this section, we have confirmed that MaxSAT could be a good approach to solve the Tuple Number problem with constraints. We have also observed that with a relatively small number of tests we can cover most of the tuples, and that this approach can be useful for medium-sized instances that do not need a large number of tests to reach a reasonable coverage percentage.

In the next section, we explore the Incremental Test Suite Construction for the Tuple Number problem described in section \ref{sec:hybrid}. It allows us to tackle more efficiently those Tuple Number problems involving a relatively large number of tests.

\subsection{MaxSAT based Incremental Test Suite Construction for $T(N;t,S)$}
\label{sec:incrementalOSCAR}

In section \ref{sec:com_incomp_OSCAR}, we have analyzed an approach that can be used to maximise the number of tuples covered by a number of tests inferior to $CAN(t,S)$. However, we have seen that this might not be the most efficient solution if we require to compute the Tuple Number problem for a large enough number of tests.

{\bf Solving approaches:} Here we propose three incomplete alternatives for solving the Tuple Number problem, with the aim of improving the results obtained in section \ref{sec:com_incomp_OSCAR}. Our hypothesis is that the application of incomplete approaches can be more suitable when solving bigger instances. 

The first approach is the greedy algorithm presented in \cite{Yamada16}, referred  as $maxh-its$. This algorithm incrementally adds a test at a time. The test is constructed through a heuristic \cite{Czerwonka06} that tries to increase the number of covered tuples so far, by selecting at each step the parameter tuple with most value tuples yet to be covered.

The second approach is the Incremental Test Suite Construction from section~\ref{sec:hybrid} (referred here as $maxsat-its$), which also adds a test at a time \footnote{The algorithm allows to add more than one test at a time, but this experiment is out of reach in this paper.}, but this test is built by solving the Tuple Number problem through an incomplete MaxSAT solver instead of using a heuristic as in the previous approach.  

In the third approach, instead of a MaxSAT query, as in the second approach, we apply a SAT query to return a test that covers at least one more tuple (referred as $sat-its$) than the incremental test suite built so far. 

We also evaluate the approach described in section \ref{sec:pctg}. The idea is to relax the Covering Array Number problem by allowing to cover only a 95\% of the allowed tuples ($\tau_a$). We refer to this approach as 
$mints-95\%|\tau_a|$. As for the Covering Array Number problem, we use the upper bound returned by the ACTS tool (see section \ref{sec:preproc}) for the initial number of tests.

{\bf Results:} We present the relative performance of the previous four approaches respect to the best incomplete MaxSAT approach (\emph{tt-open-wbo-inc}) for solving the Tuple Number problem from section \ref{sec:com_incomp_OSCAR}, referred as $\simeq T(N;t,S)$ (we use the symbol $\simeq$ to indicate that the values reported for $\simeq T(N;t,S)$ correspond to suboptimal solutions). All the approaches shown in this section also use the incomplete SAT-based MaxSAT solver \emph{tt-open-wbo-inc}, except $sat-its$ which uses the Glucose41 SAT solver. For the encoding of equation (a) of $CCX$ we use variation a.2 $(c^{i}_{\tau} \leftrightarrow c^{i-1}_{\tau} \vee x_{i,p,v})$ as in section \ref{sec:com_incomp_OSCAR}.

To perform a fair comparison we tried to execute all the algorithms within the same runtime conditions. We use the runtime $maxsat-its$ needs to cover all the allowed tuples as a reference. In more detail, we set a timeout of 100s to each iteration of the $maxsat-its$ approach \footnote{We assume that $maxsat-its$ is able to cover at least one more tuple in 100 seconds}. Therefore, the total runtime in seconds consumed by $maxsat-its$ is the number of test it reaches multiplied by 100. For $maxh-its$ and $sat-its$, the timeout is the total runtime consumed by $maxsat-its$. For $mints-95\%|\tau_a|$, we use as timeout the runtime consumed by  $\simeq T(N;t,S)$ to reach 95\% of coverage. Finally, for $\simeq T(N;t,S)$, we use a timeout of $N \cdot 100$ seconds for each $N$. Notice that in this last case we are ensuring that for a given $N$, both $\simeq T(N;t,S)$ and $maxsat-its$ approaches will have the same execution time limits.

\begin{figure}[h!t]
    \caption{Comparison of the required number of tests for different methods with regards to the number of test used by $\simeq T(N,t,S)$ (as base) to cover each number of tuples. %
    }
    \label{fig:hybrid-tests}
    \centering
    \includegraphics[width=0.4\textwidth]{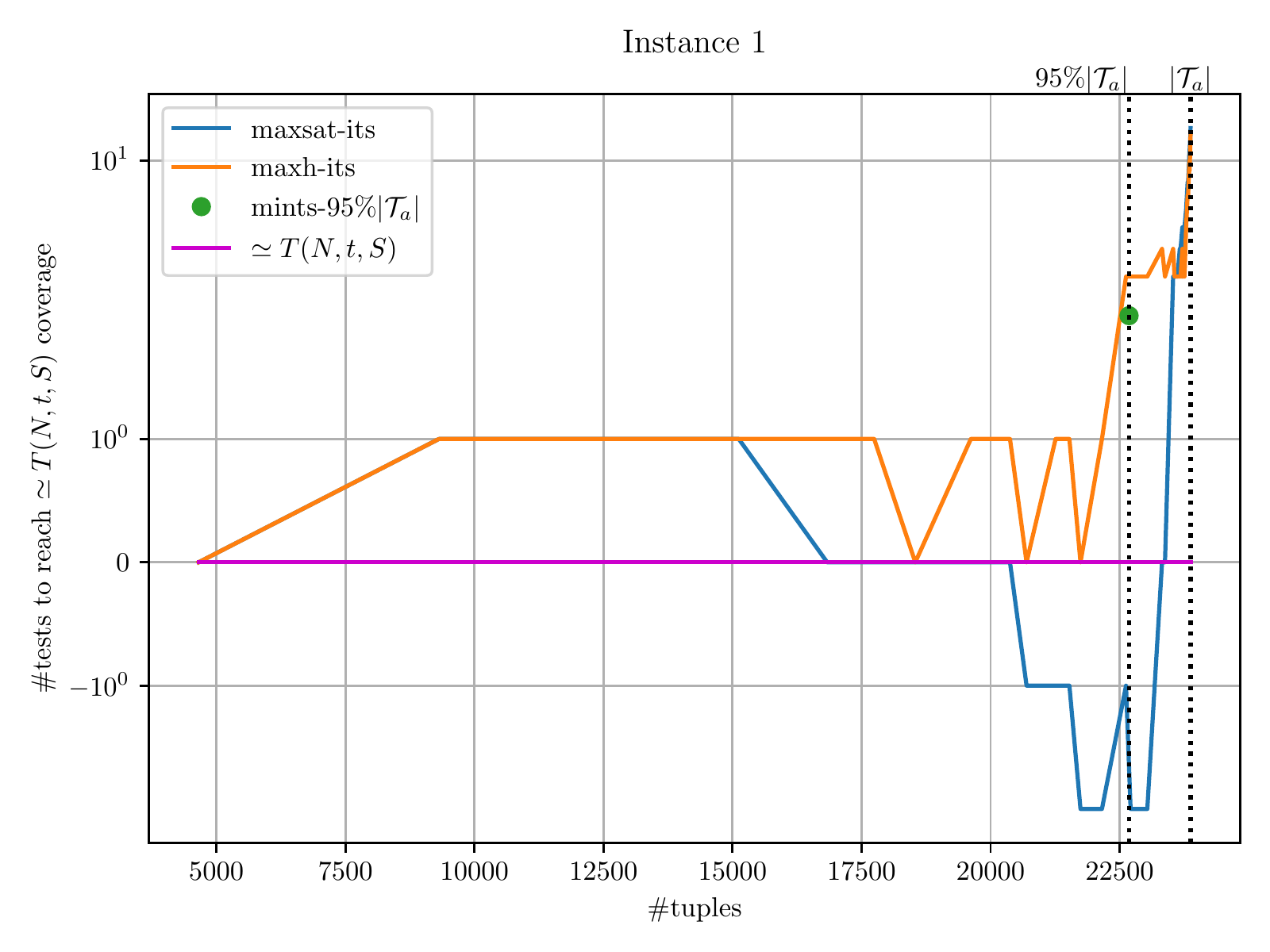}
    \includegraphics[width=0.4\textwidth]{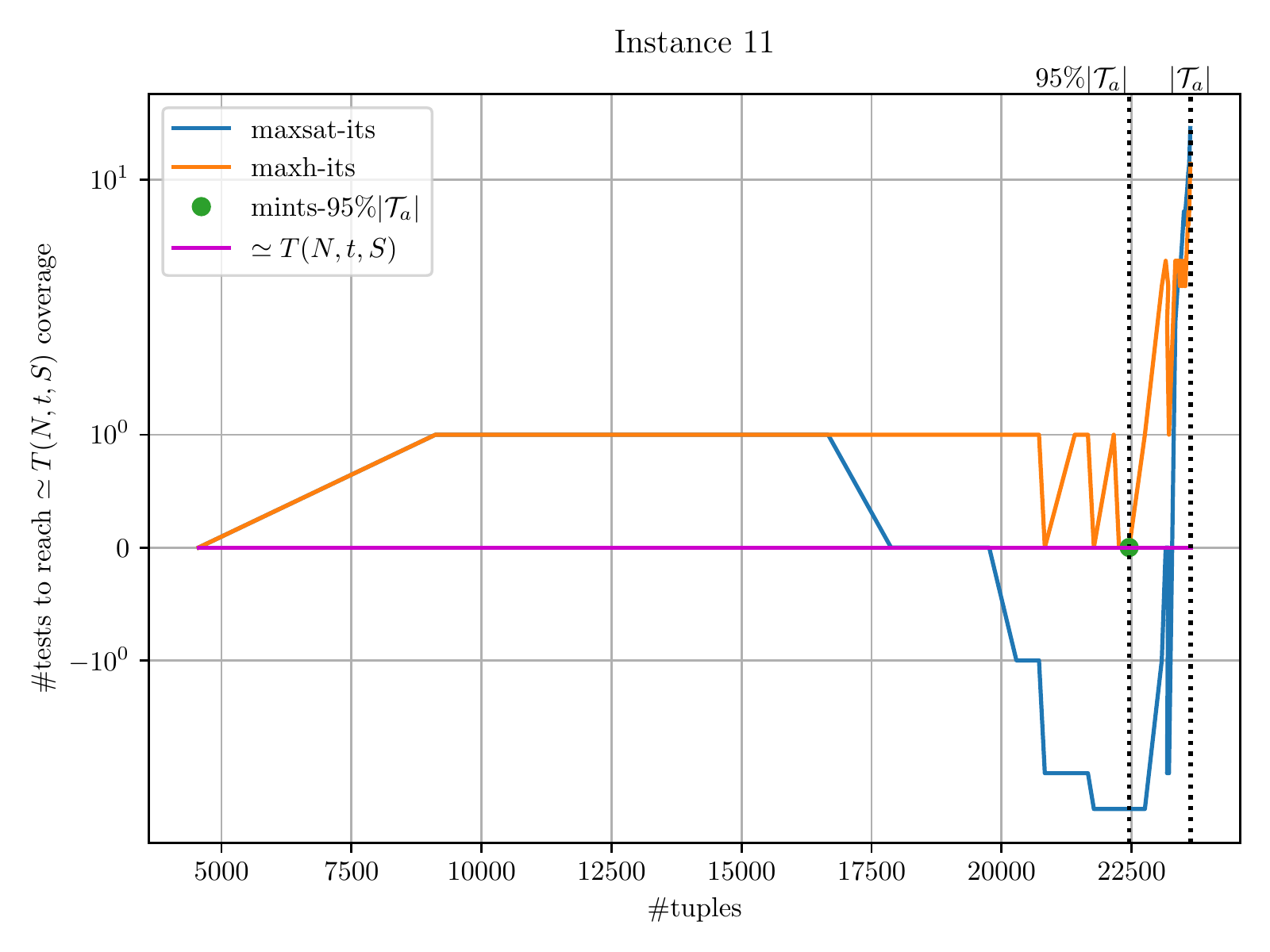} \\
    \includegraphics[width=0.4\textwidth]{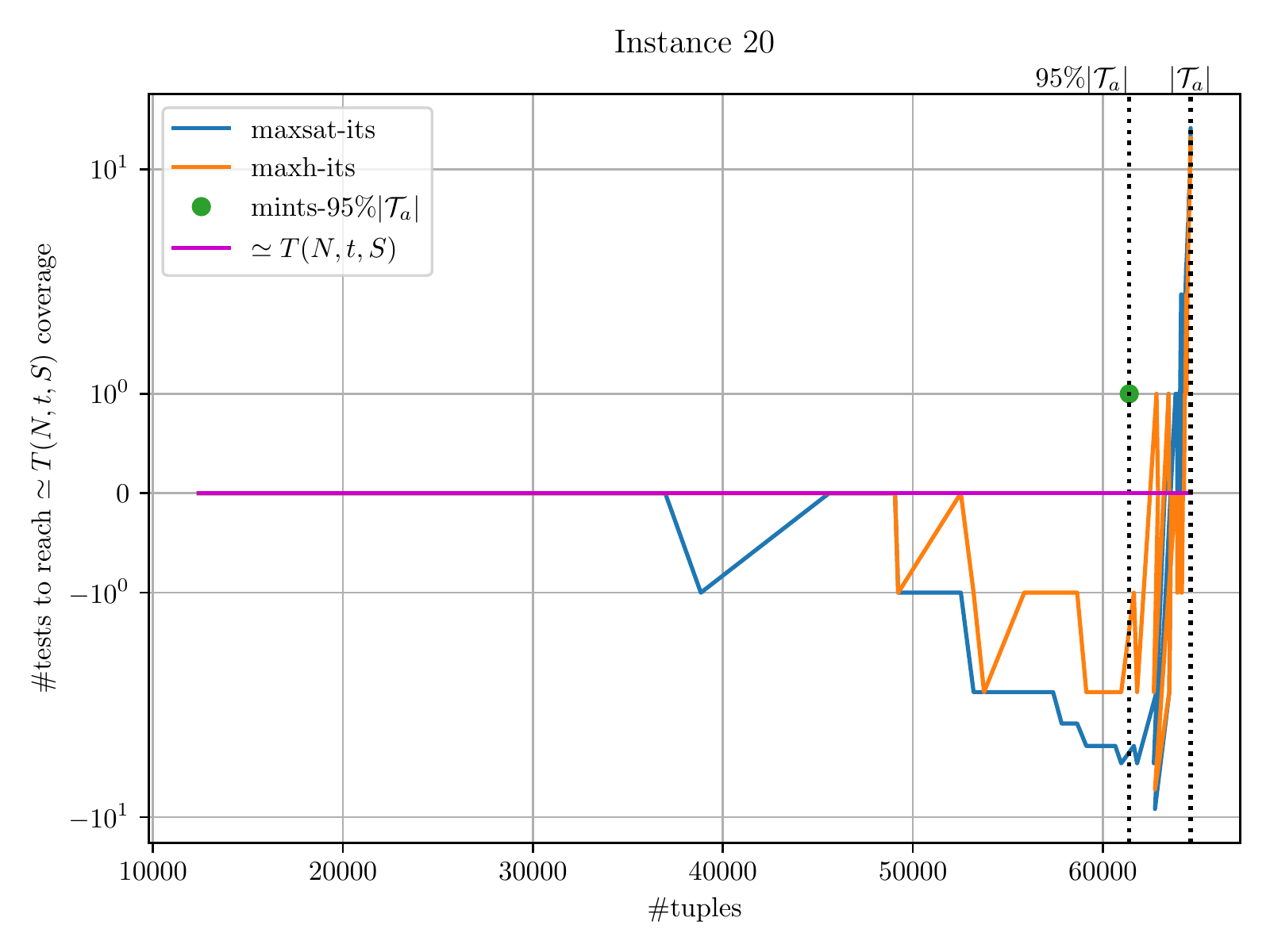}
    \includegraphics[width=0.4\textwidth]{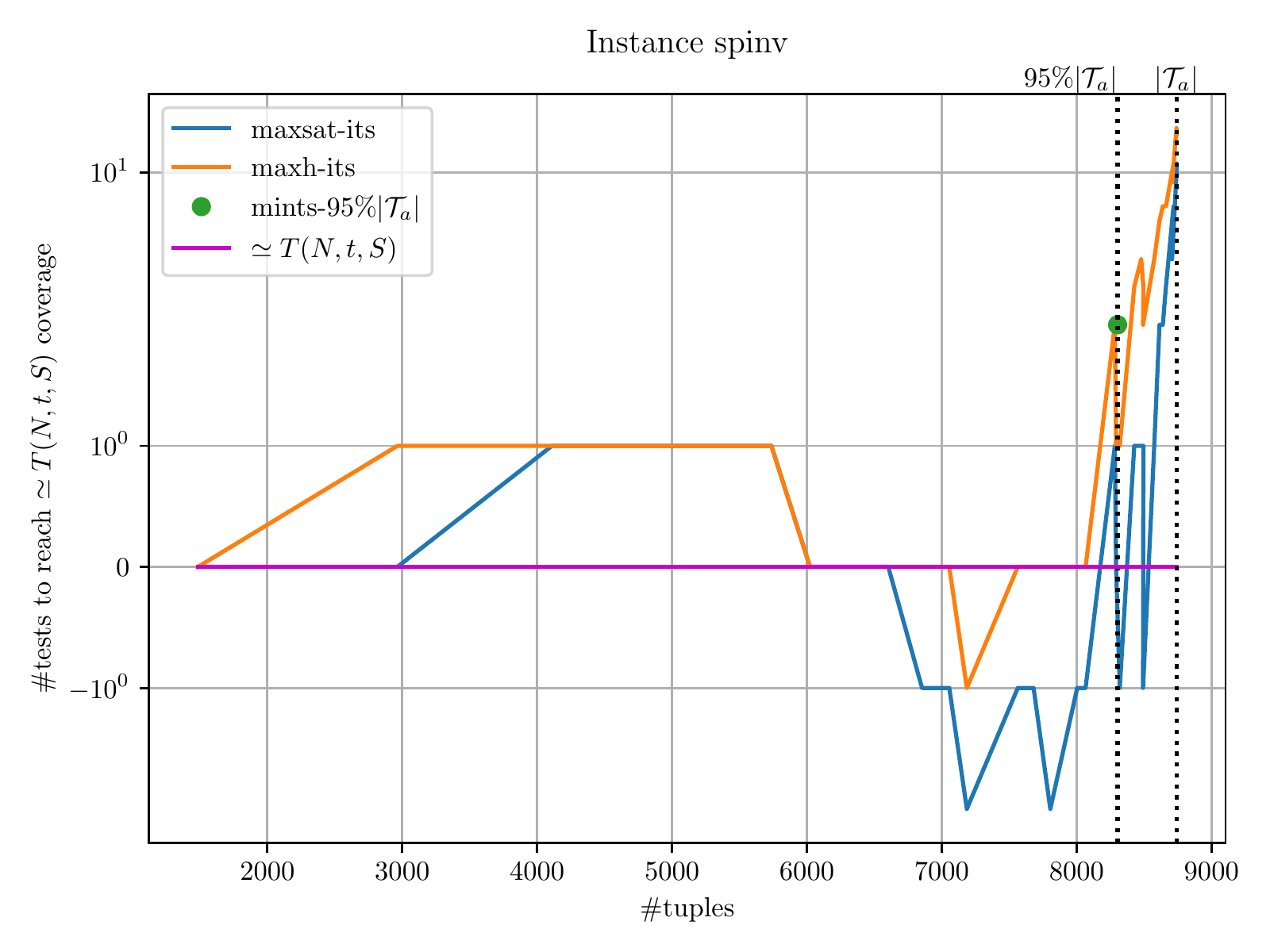} \\
    \includegraphics[width=0.4\textwidth]{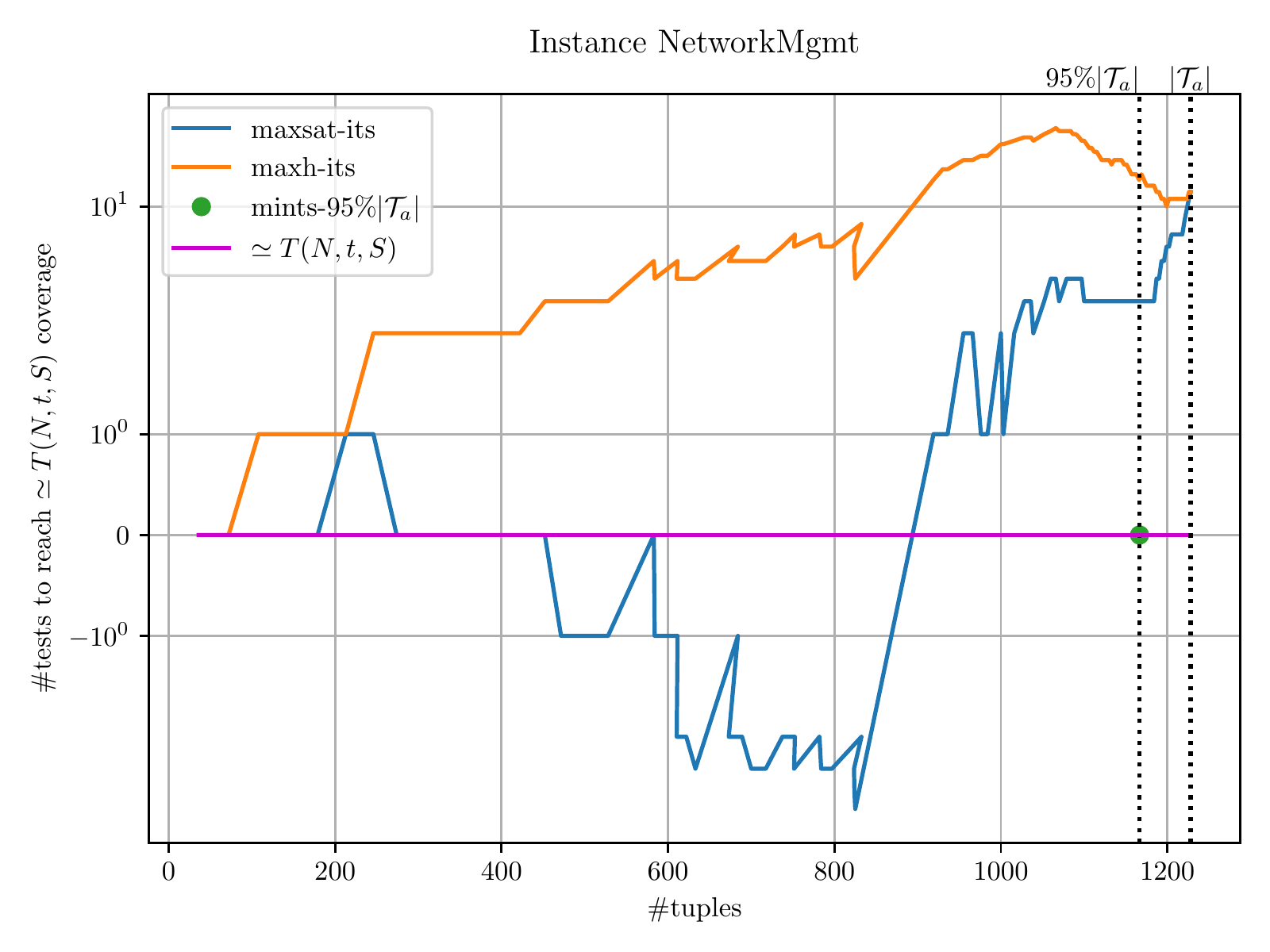}
    \includegraphics[width=0.4\textwidth]{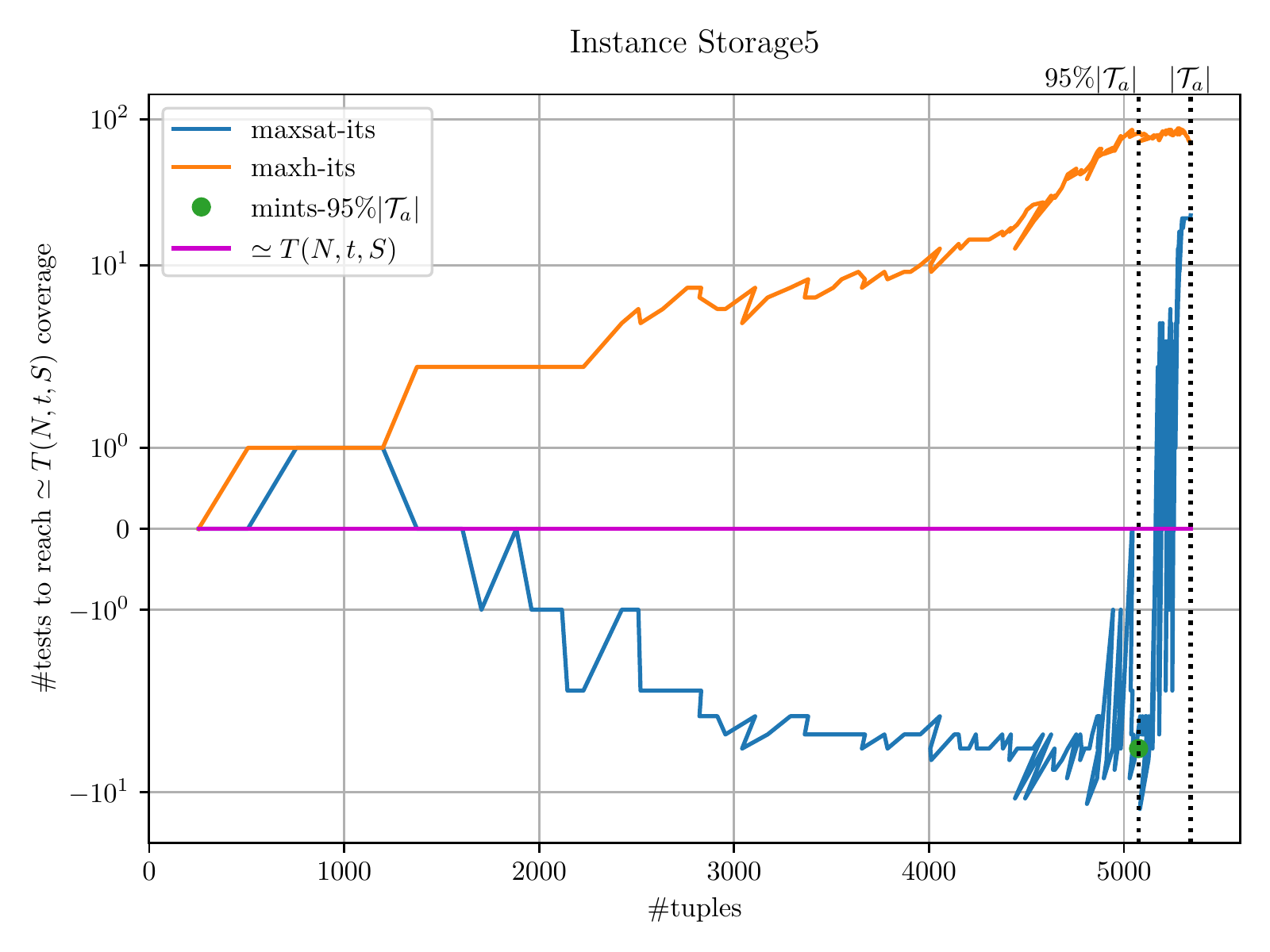} \\
    \includegraphics[width=0.4\textwidth]{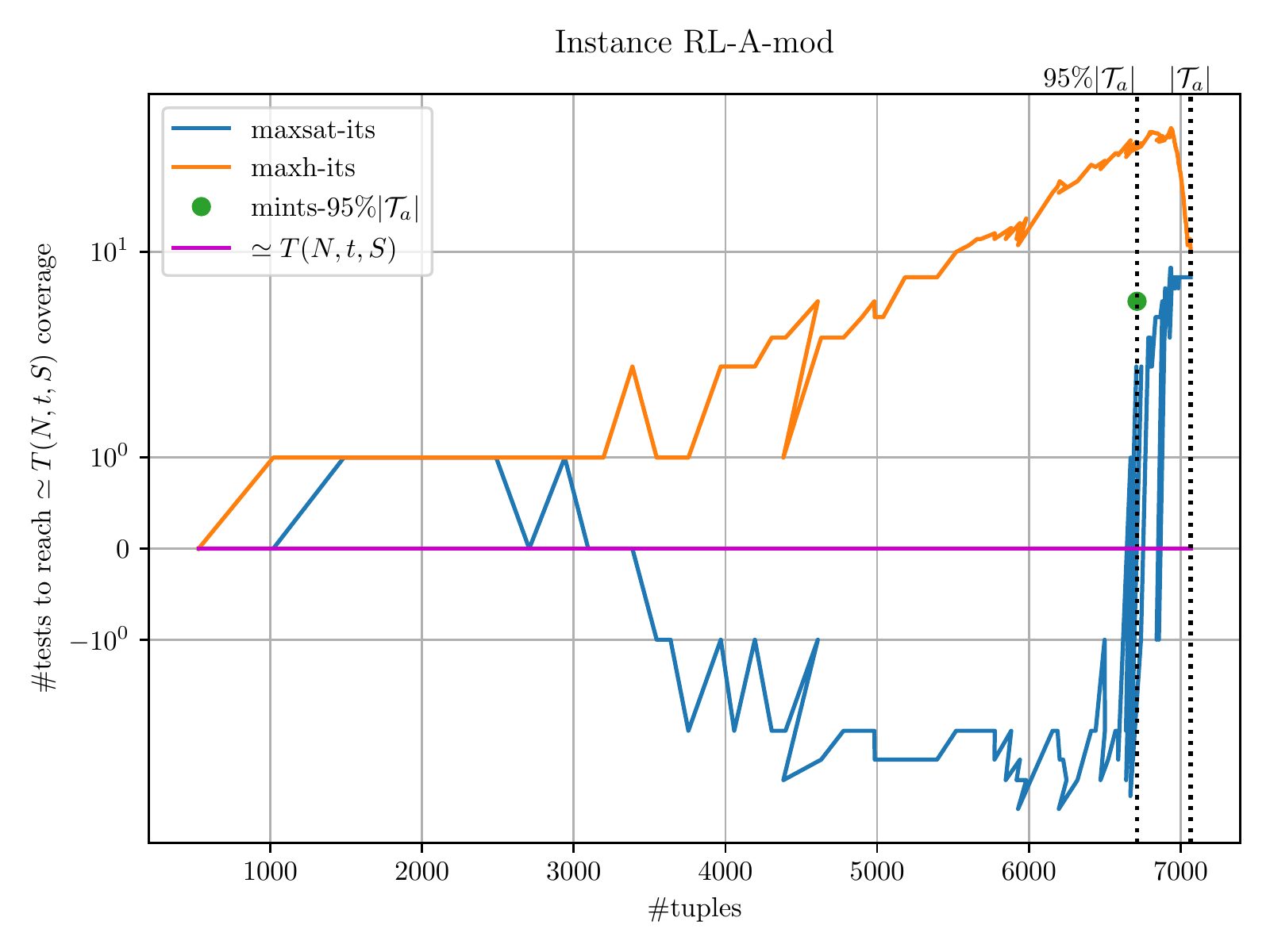}
    \includegraphics[width=0.4\textwidth]{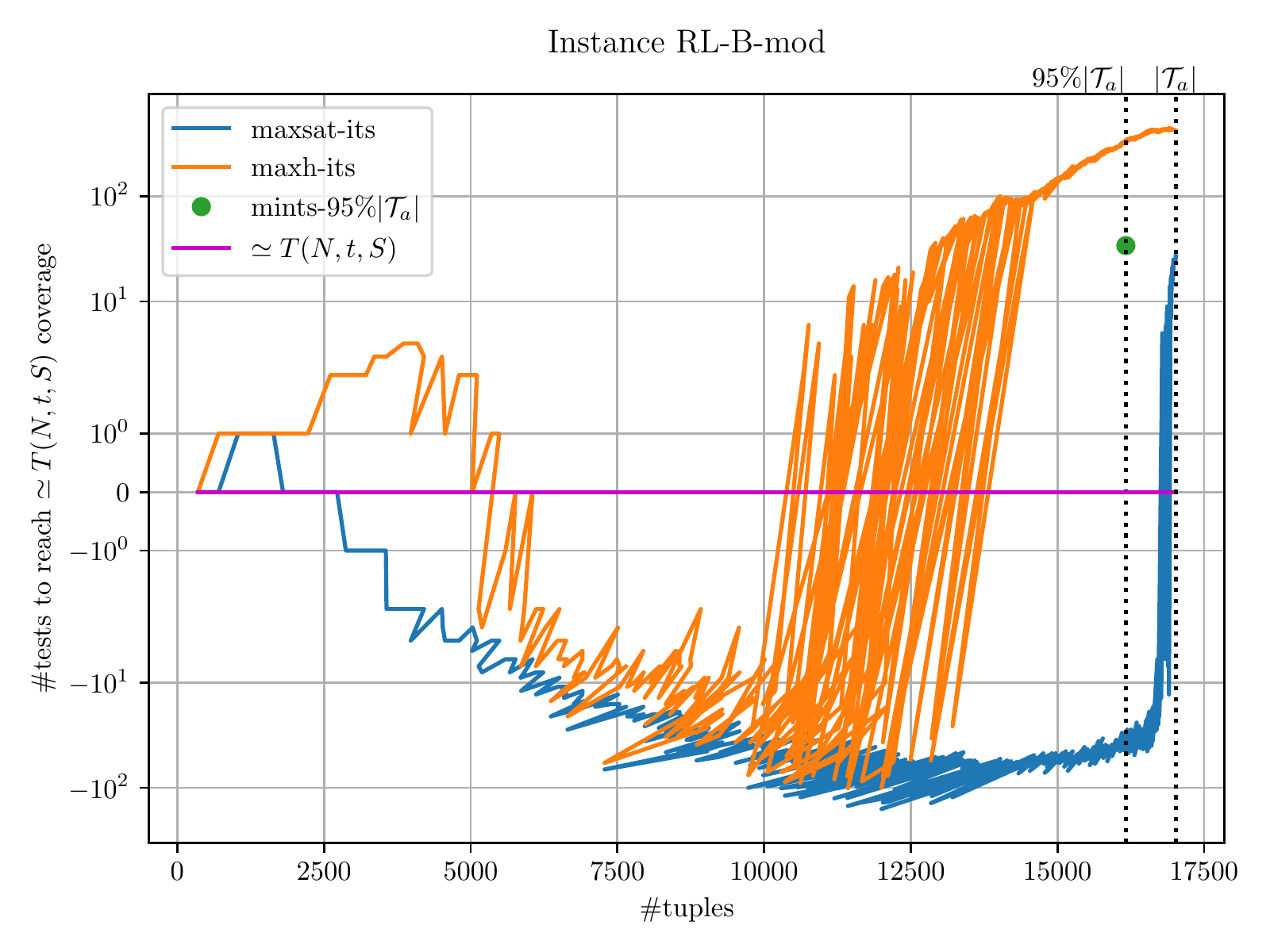}
\end{figure}

\begin{figure}[h!t]
    \caption{Comparison of the required number of tests for different methods to cover as much tuples at each test from $\simeq T(N,t,S)$ (as base). %
    }
    \label{fig:hybrid-test-diff}
    \centering
    \includegraphics[width=0.4\textwidth]{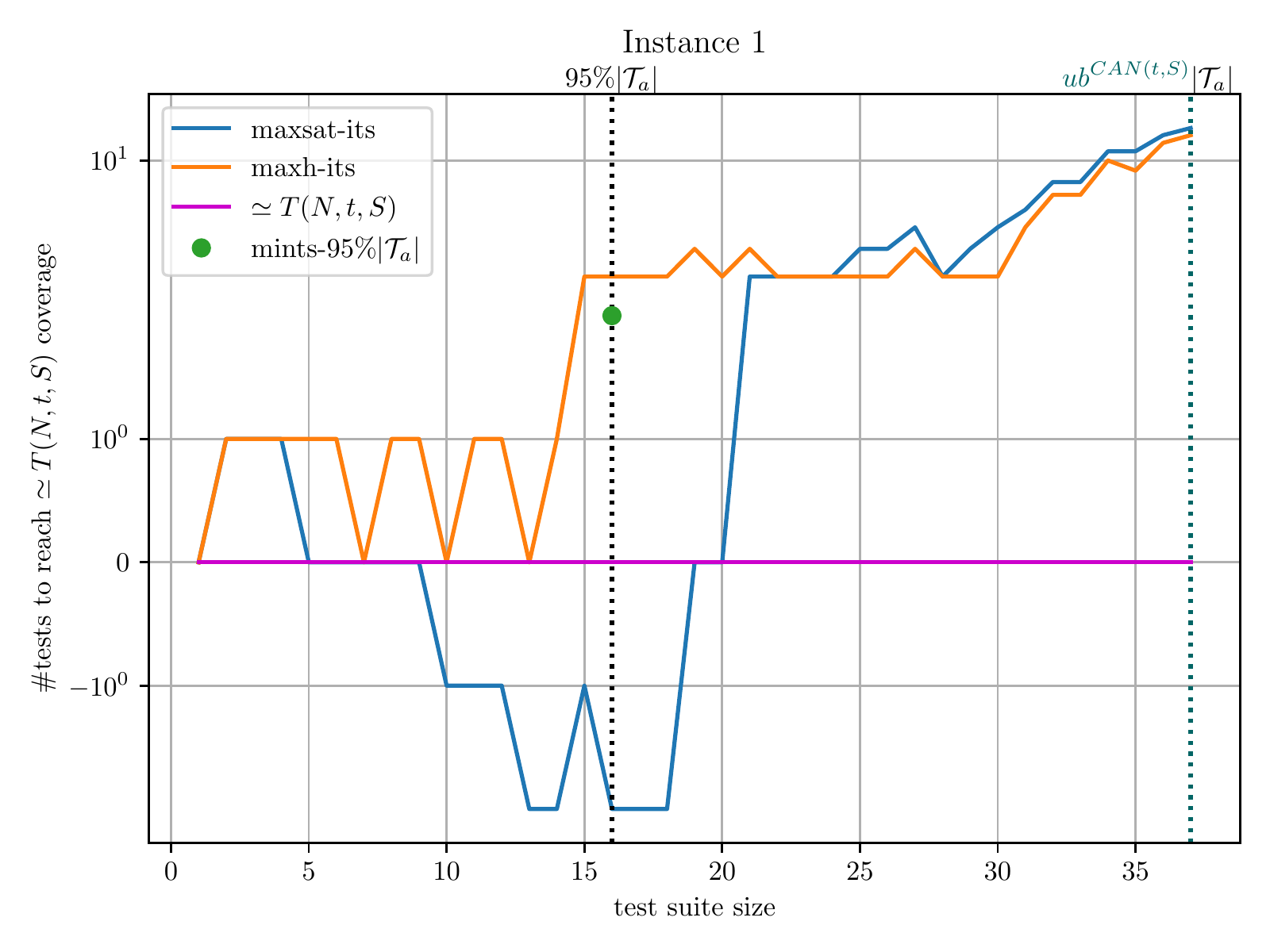}
    \includegraphics[width=0.4\textwidth]{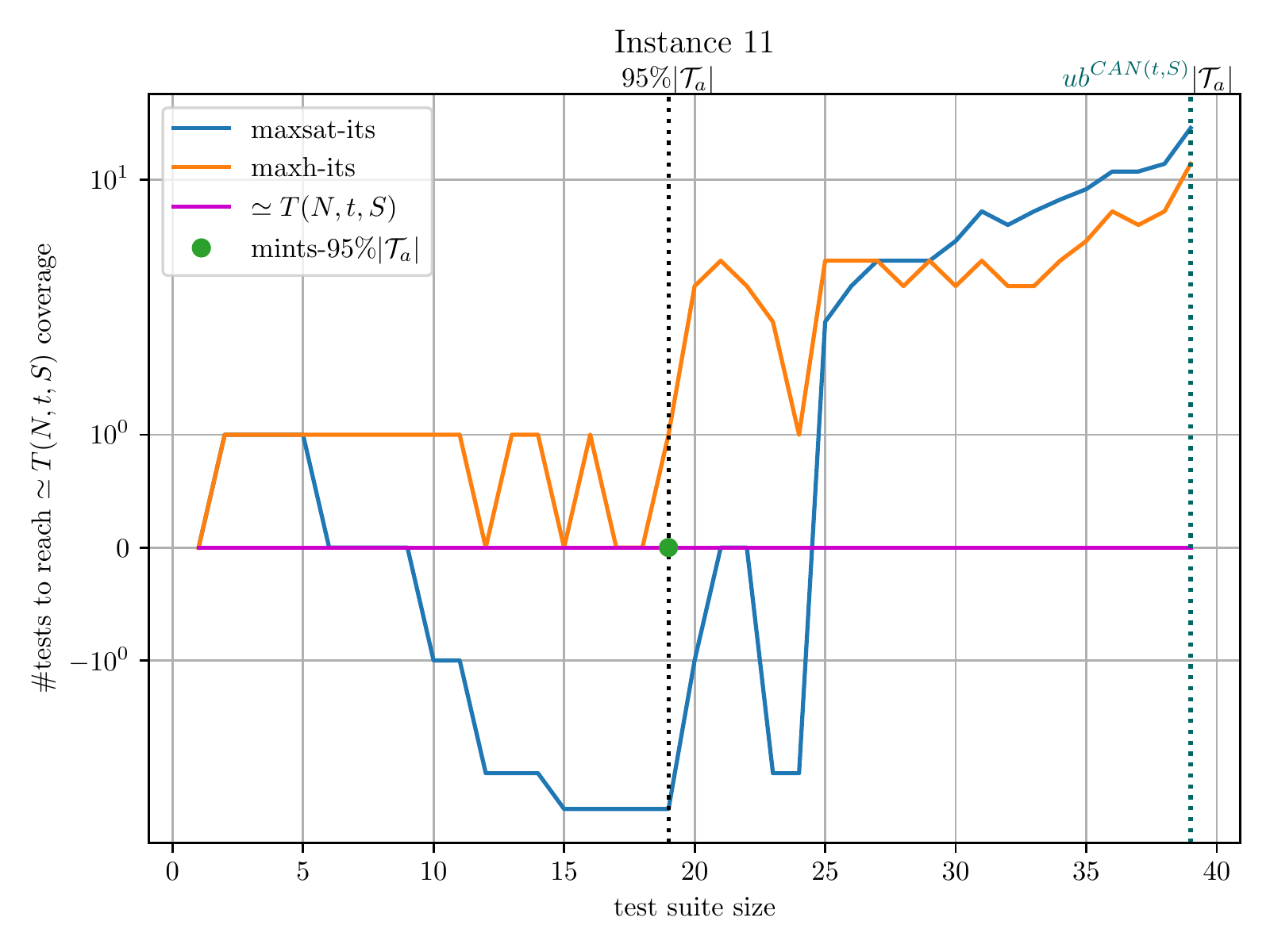} \\
    \includegraphics[width=0.4\textwidth]{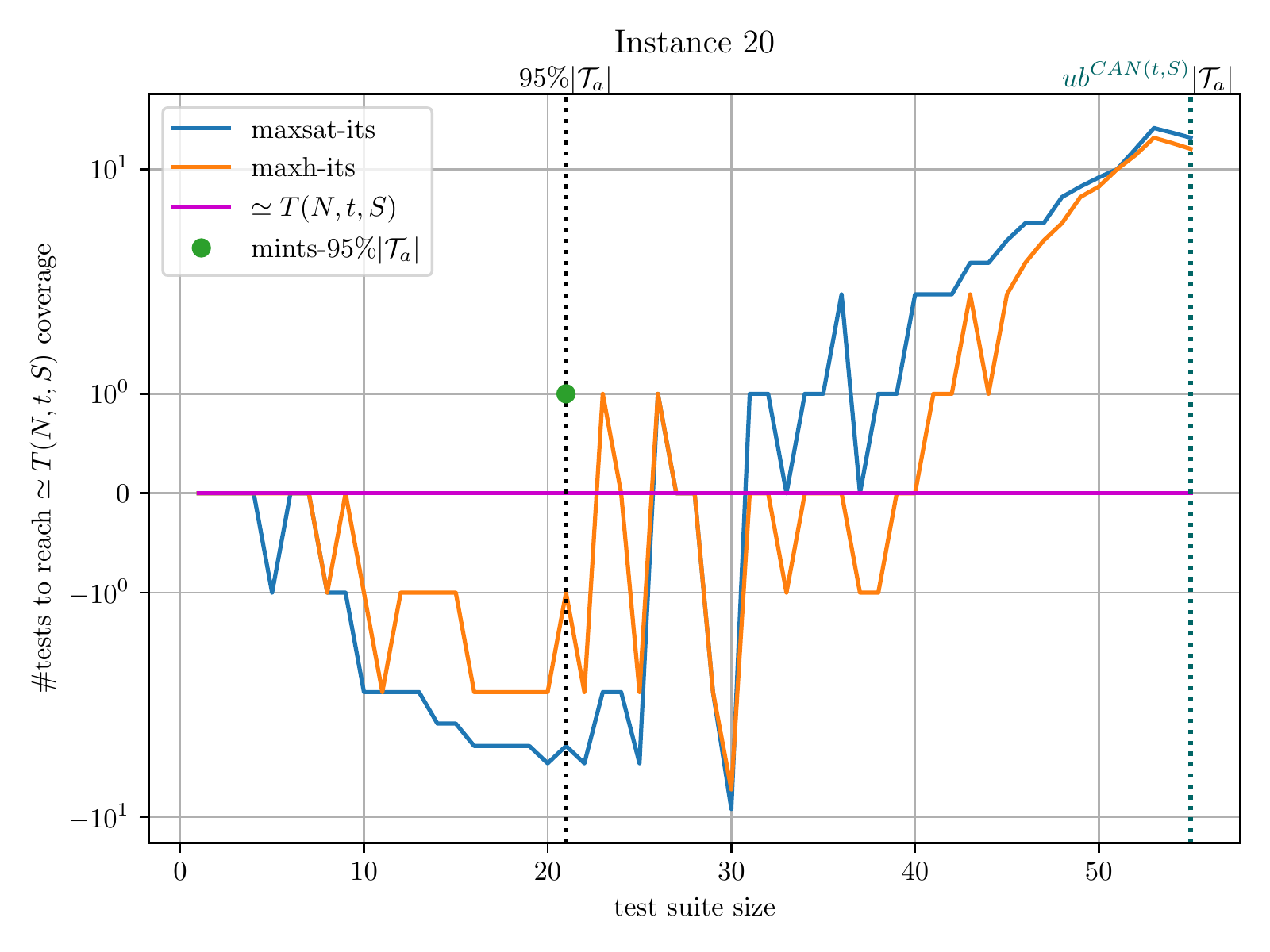}
    \includegraphics[width=0.4\textwidth]{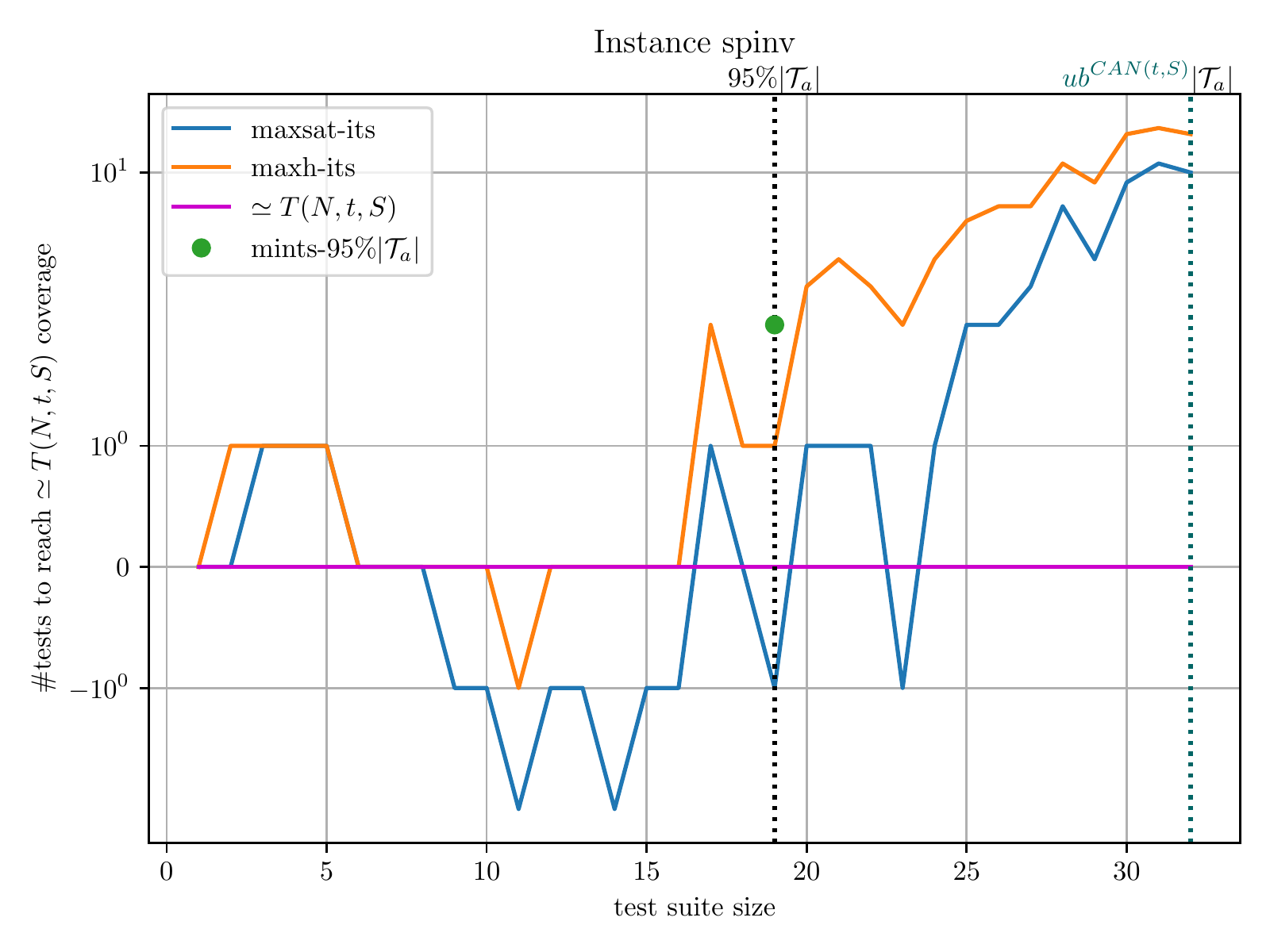} \\
    \includegraphics[width=0.4\textwidth]{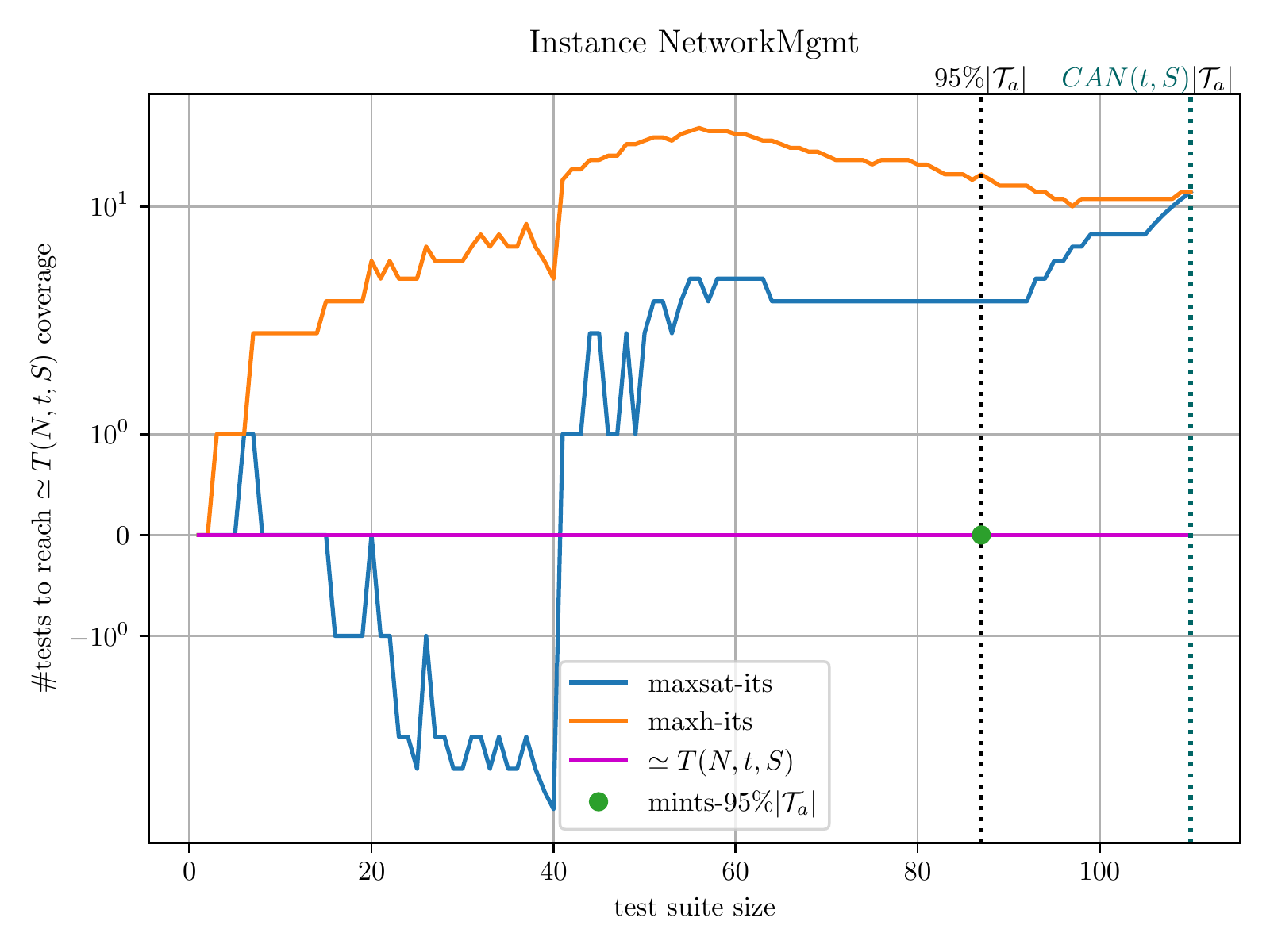}
    \includegraphics[width=0.4\textwidth]{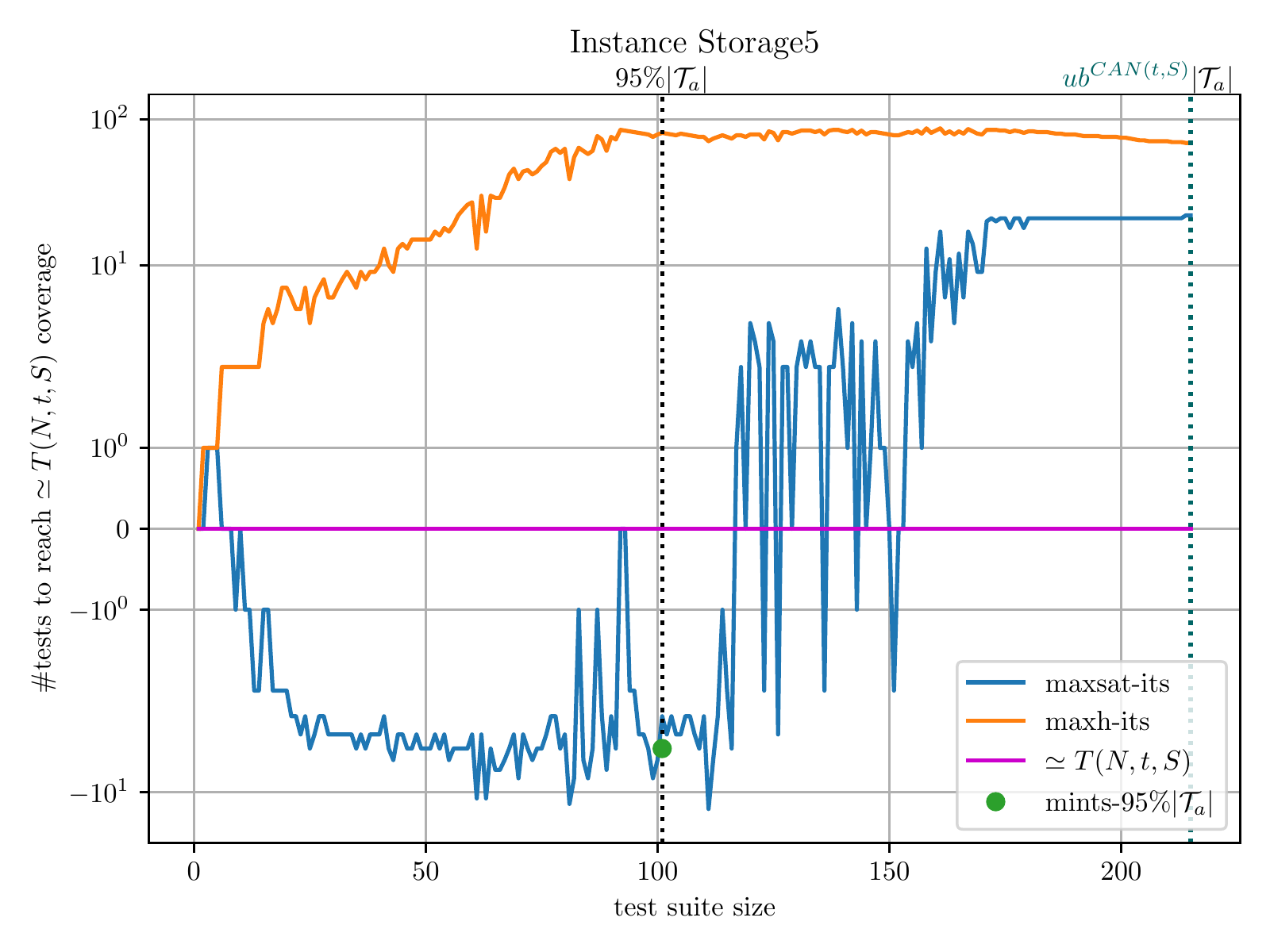} \\
    \includegraphics[width=0.4\textwidth]{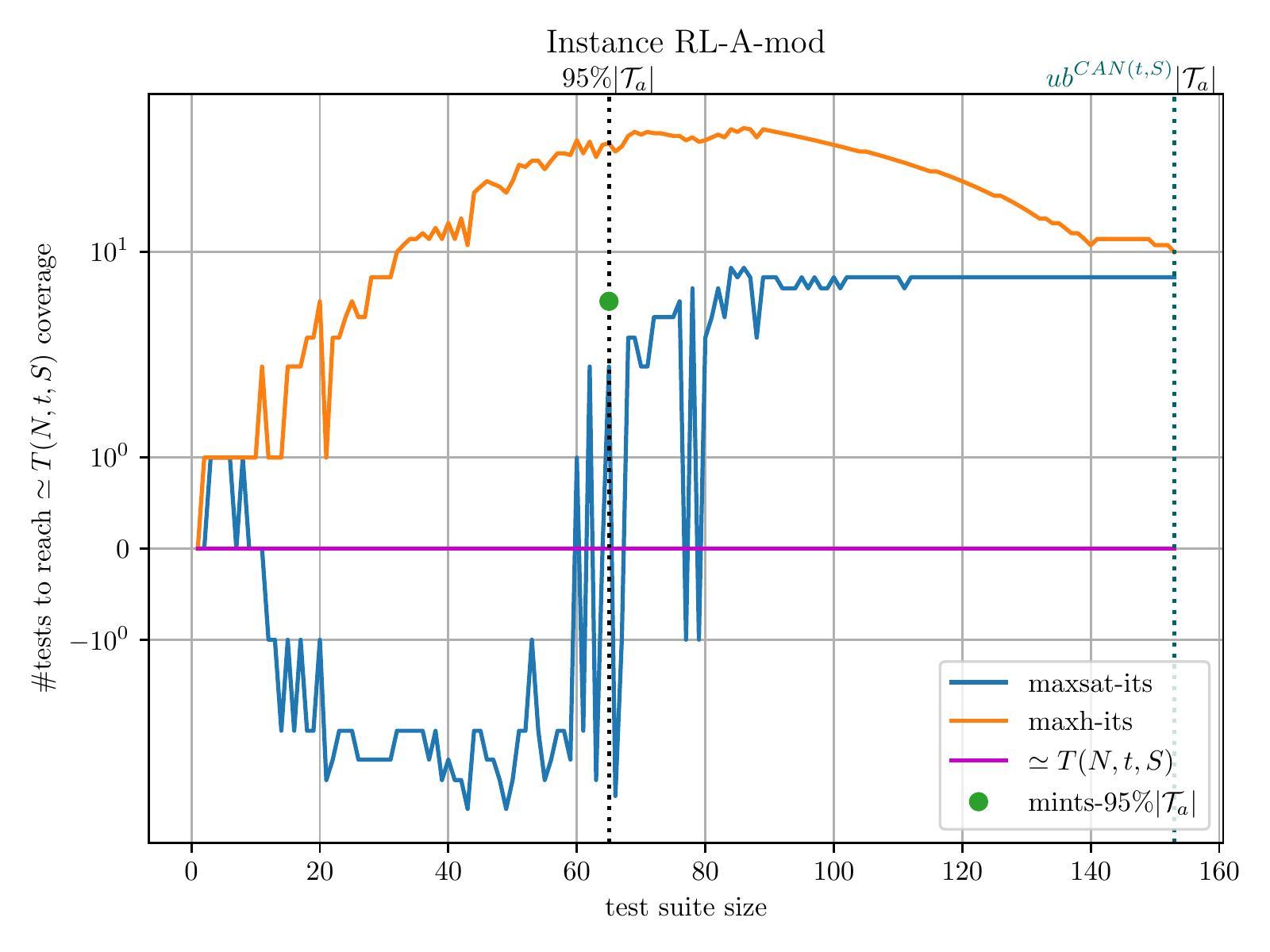}
    \includegraphics[width=0.4\textwidth]{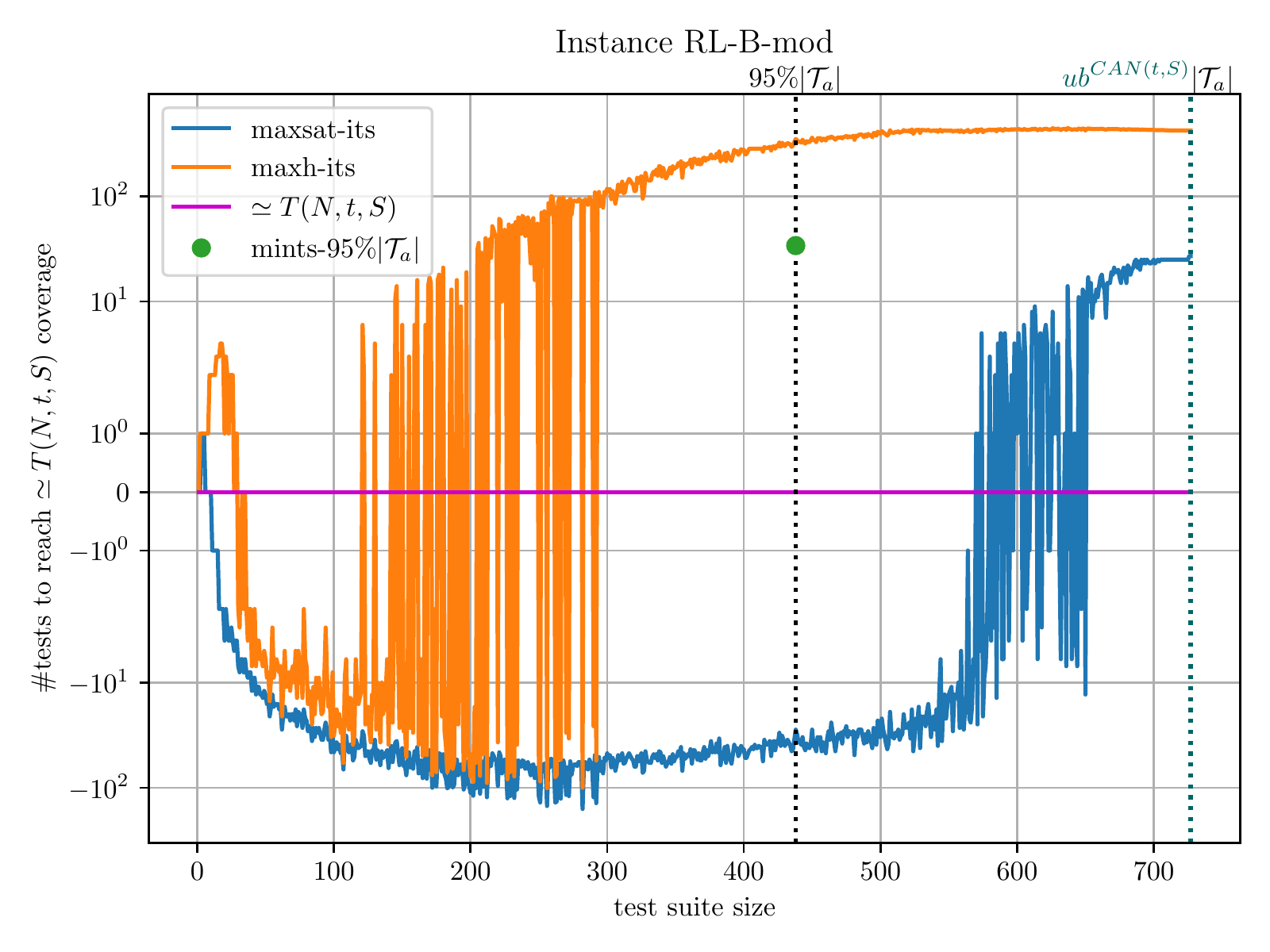}
\end{figure}

All approaches have been executed with 3 seeds and the mean is reported. The experimental results are presented in Figures \ref{fig:hybrid-tests} and \ref{fig:hybrid-test-diff}. As in section \ref{sec:com_incomp_OSCAR}, we only plot the most representative instances.

Figure \ref{fig:hybrid-tests} shows the increment (or decrement) of the number of tests required by $maxsat-its$, $maxh-its$ and $mints-95\%|\tau_a|$ to cover the same number of tuples as $\simeq T(N;t,S)$. On the other hand, Figure \ref{fig:hybrid-test-diff} shows the increment (or decrement) of tests required to reach the same coverage ratio as $\simeq T(N;t,S)$. For $sat-its$ approach we found that in most cases it is able to cover only one tuple per test, so we decided to exclude these results in the figures as they were clearly outperformed by the rest of the presented approaches.

In both figures, we plot a vertical line to show the points where $\simeq T(N;t,S)$ reaches 95\% and 100\% of tuples covered.

In general, $maxsat-its$ clearly outperforms $maxh-its$. This can be expected since the nature of the incremental approach is to do the best at each possible iteration, and $maxsat-its$ tackles exactly this goal by solving the Tuple Number problem, while $maxh-its$ do not.

We also observe that $maxsat-its$ outperforms the tuple coverage that $\simeq T(N;t,S)$ can achieve on the first tests. Particularly, $maxsat-its$ is able to improve the number of tests required to cover 95\% of the allowed tuples in 7 of the 8 instances we show in Figures \ref{fig:hybrid-tests} and \ref{fig:hybrid-test-diff}. On the other hand, above 95\%, $\simeq T(N;t,S)$ seems to be the best approach in terms of using fewer tests for the same coverage. This makes sense since the incomplete nature of $maxsat-its$ make it less efficient when approaching the complete coverage, what may not be need it for several applications.

\begin{figure}[h!t]
    \centering
    \includegraphics[width=0.5\textwidth]{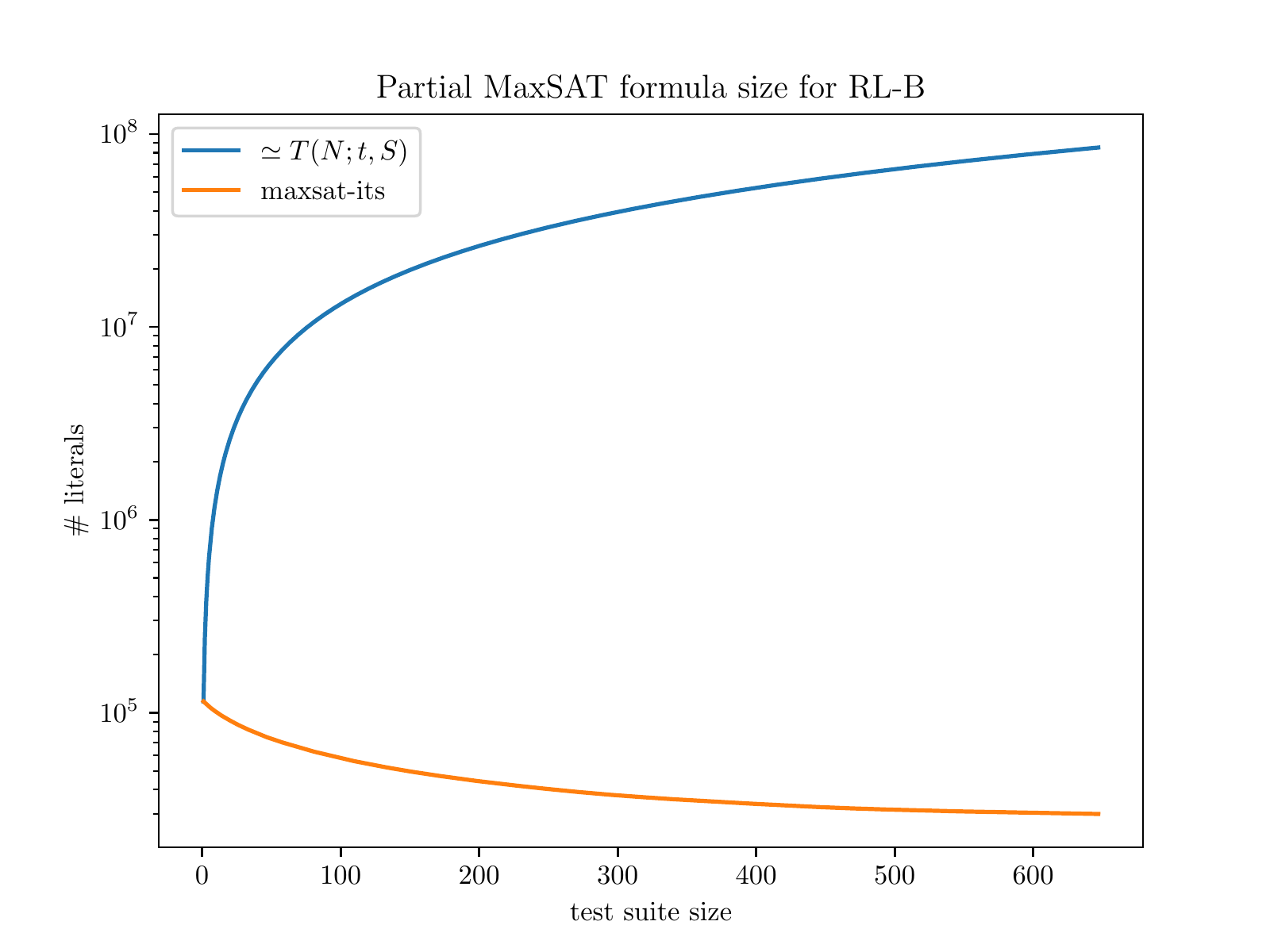}
    \caption{Partial MaxSAT formula size for RL-B in literals as a function of test suite size.}
    \label{fig:rl-b-sizes}
\end{figure}

In figure \ref{fig:hybrid-tests} we observe an erratic behaviour of instance \emph{RL-B}, which is the larges instance that we had available. These results are in line with the ones in figure \ref{fig:shortening-tt-open-wbo-inc} of section \ref{sec:com_incomp_OSCAR}, and might show the possible issues that $\simeq T(N;t,S)$ can suffer when dealing with large instances. In particular, figure \ref{fig:rl-b-sizes} shows the number of literals of the MaxSAT instance solved by $\simeq T(N;t,S)$ and $maxsat-its$ as the size of the test suite increases for the \emph{RL-B} benchmark. We observe that $\simeq T(N;t,S)$ has to deal with an increasing size of the Partial MaxSAT instance proportional to the number of tests in the test suite.
In contrast, for $maxsat-its$, the size of the instance decreases since only encodes one test and the number of tuples to cover decreases along with the size of the test suite built so far.  This is an interesting insight since RL-B instance comes from an industrial application and it may reflect what we can face in harder real-world scenarios. Therefore, $maxsat-its$ may seem more well suited for these harder real-world domains and may extend the reach of Combinatorial Testing for more complex SUTs.

Finally, we also observe that the $mints-95\%|\tau_a|$ approach might not be the best option to obtain a good suboptimal test suite that covers 95\% of the total tuples. However, for some instances, it obtains better results than any other tested method (\emph{NetworkMgmt} and \emph{Storage5}). We also have to note that this is the only tested method that can certify the optimality of the obtained test suite when combined with a complete MaxSAT solver.

\section{Conclusions}\label{sec:conclusions}

We have shown that MaxSAT technology is well-suited for solving the  Covering Array Number problem for Mixed Covering Arrays with Constraints through SAT technology. In particular, we discussed efficient encodings and how MaxSAT algorithms perform on them.

We also presented MaxSAT encodings for the Tuple Number problem. To our best knowledge, this is the first time that this problem is studied with SUT Constraints. Additionally, we presented a new incomplete algorithm which can be applied efficiently to solve those instances where the Tuple Number problem encoding into MaxSAT is too large. In particular, we proved we can build good enough solutions by incrementally adding a new test  synthesized through a MaxSAT query that aims to maximize the coverage of additional allowed tuples, respect to the test suite under construction.

Another interesting result that we obtained is that if we do not aim to cover all $t$-tuples but a \emph{statistically significant} fraction, we can save a great amount of tests. We experimentally showed that,  to cover a 95\% percentage, we just need, on average, a 52\% percentage of the best suboptimal solution reported so far. This is of high practical importance for applications where test cases are expensive according to the budget.

From the point of view of Combinatorial Testing, it is reasonable to say that the practical and theoretical interest application of our findings and approaches will grow proportionally to the hardness or complexity of the SUT constraints. This will certainly extend the reach of Combinatorial Testing to more challenging SUTs.

From the point of view of Constraint programming, the lessons learnt on how to design efficient encodings for MaxSAT solvers can be exported to solve similar problems. These problems are roughly characterized by having an objective function whose size is proportional to the best known upper bound. 

SAT and MaxSAT communities will also benefit from new challenging benchmarks to test the new advances in the field. Moreover, any future advance in MaxSAT technology can be applied to solve more efficiently the Covering Array Number and Tuple Number problems with no additional cost.

\section*{Acknowledgements}
We would like to thank specially Akihisa Yamada for the access to several benchmarks for our experiments and solving some questions about his previous work on Combinatorial Testing with Constraints. 

This work was partially supported by  the MINECO-FEDER project  TASSAT3  (TIN2016-76573-C2-2-P), the  MICINNs  project  PROOFS (PID2019-109137GB-C21) and ISINC (PID2019-111544GB-C21),  and the MICNN FPU fellowship (FPU18/02929).

\bibliographystyle{abbrv}
\bibliography{main}

\end{document}